\documentclass{article}

\PassOptionsToPackage{numbers, sort}{natbib}
 \usepackage[preprint]{neurips_2026}

\usepackage[utf8]{inputenc} 
\usepackage[T1]{fontenc}    
\usepackage{hyperref}       
\usepackage{url}            
\usepackage{booktabs}       
\usepackage{amsfonts}       
\usepackage{nicefrac}       
\usepackage{microtype}      
\usepackage{xcolor}         

\usepackage{amsmath}
\usepackage{graphicx}
\usepackage{subcaption}
\usepackage{multirow}
\usepackage[most]{tcolorbox}

\usepackage{listings}
\lstdefinestyle{dialogue}{
    language={},
    basicstyle=\ttfamily\scriptsize,
    framesep=3pt,           
    aboveskip=3pt,          
    belowskip=3pt,
    lineskip=0pt,           
    breaklines=true,
    breakatwhitespace=true,
    breakautoindent=false,
    breakindent=20pt,
    columns=flexible,
    keepspaces=true,
    tabsize=2,
    showstringspaces=false,
    backgroundcolor=\color{gray!2},      
    frame=tb,
    rulecolor=\color{black!40},          
    stringstyle=\color{black!90},        
    morestring=[b]",
    classoffset=1, morekeywords={System}, keywordstyle=\bfseries\color{black},
    classoffset=2, morekeywords={User}, keywordstyle=\bfseries\color{blue!45!black},
    classoffset=3, morekeywords={Assistant}, keywordstyle=\bfseries\color{teal!70!black},
    classoffset=0
}
\title{SemanticOpt: Towards LLM-Based Semantic Black-Box Optimization}

\author{
\textbf{Jamison Meindl}$^{1}$\thanks{Correspondence to: Jamison Meindl <\texttt{jmeindl@mit.edu}>.}
\quad
\textbf{Yunsheng Tian}$^{1}$
\quad
\textbf{Tony Cui}$^{1}$
\quad
\textbf{Veronika Thost}$^{2}$ \\
\textbf{Zhang-Wei Hong}$^{2}$
\quad
\textbf{Jie Chen}$^{2}$
\quad
\textbf{Wojciech Matusik}$^{1}$
\quad
\textbf{Mina Konaković Luković}$^{1}$ \\
$^{1}$MIT
\quad
$^{2}$MIT-IBM Watson AI Lab
}

\begin{document}

\maketitle

\begin{abstract}
 Optimizing an experimental system can be extremely challenging when each experiment is expensive, time-consuming, or difficult to perform. Existing optimizers for expensive black-box problems, such as Bayesian optimization, are typically limited to numerical or categorical observations. They do not make use of broader domain knowledge, such as expert heuristics, relevant scientific papers, or similar previous experiments. Large language models (LLMs) can interpret this semantic information; however, even state-of-the-art LLMs struggle to reliably solve black-box optimization problems. We introduce SemanticOpt, a framework for semantic black-box optimization that equips LLMs with optimization capabilities by fine-tuning them on structured Bayesian optimization trajectories augmented with natural-language context. SemanticOpt jointly uses numerical and semantic evidence when proposing new experiments, while producing interpretable predictions aligned with Bayesian surrogate models. We construct a range of real-world optimization problems paired with semantic information to create a diverse benchmark for evaluating semantic black-box optimization. Across these domains, SemanticOpt outperforms both classical optimizers and existing LLM-based approaches on average when given relevant semantic information.
\end{abstract}

\section{Introduction}

In many scientific and engineering workflows, researchers and scientists must optimize the performance of a system while only being able to run a small number of costly evaluations. By testing a different set of parameters that define the system, the user iteratively builds an understanding of the possible performance and decides which experiment to perform next. Each evaluation may involve training a large model, running an expensive simulation, or performing a physical experiment, and the relationship between the design parameters and the final outcome is typically unknown. 

To illustrate this type of problem, consider optimizing a chemical reaction process to produce a desired compound. A scientist may want to maximize product yield by adjusting variables such as temperature, pH, and reaction time. Each trial may require running an experiment for several days before measuring the final concentration, so a limited number of experiments can be performed. A standard black-box optimizer, such as Bayesian optimization (BO), only observes the numerical configurations and the measured yield. It does not know that high temperatures may accelerate reaction rates but increase byproducts. Even if a numerical optimizer observes one high-temperature trial with low yield, it cannot determine whether high temperature is generally harmful or helpful only at a different pH or reaction time. An effective semantic optimizer would use both numerical history and semantic context to select better reaction conditions with a limited evaluation budget.

\begin{figure*}[t]
    \centering
    \begin{minipage}[b]{0.59\textwidth}
        \centering
        \includegraphics[width=\linewidth]{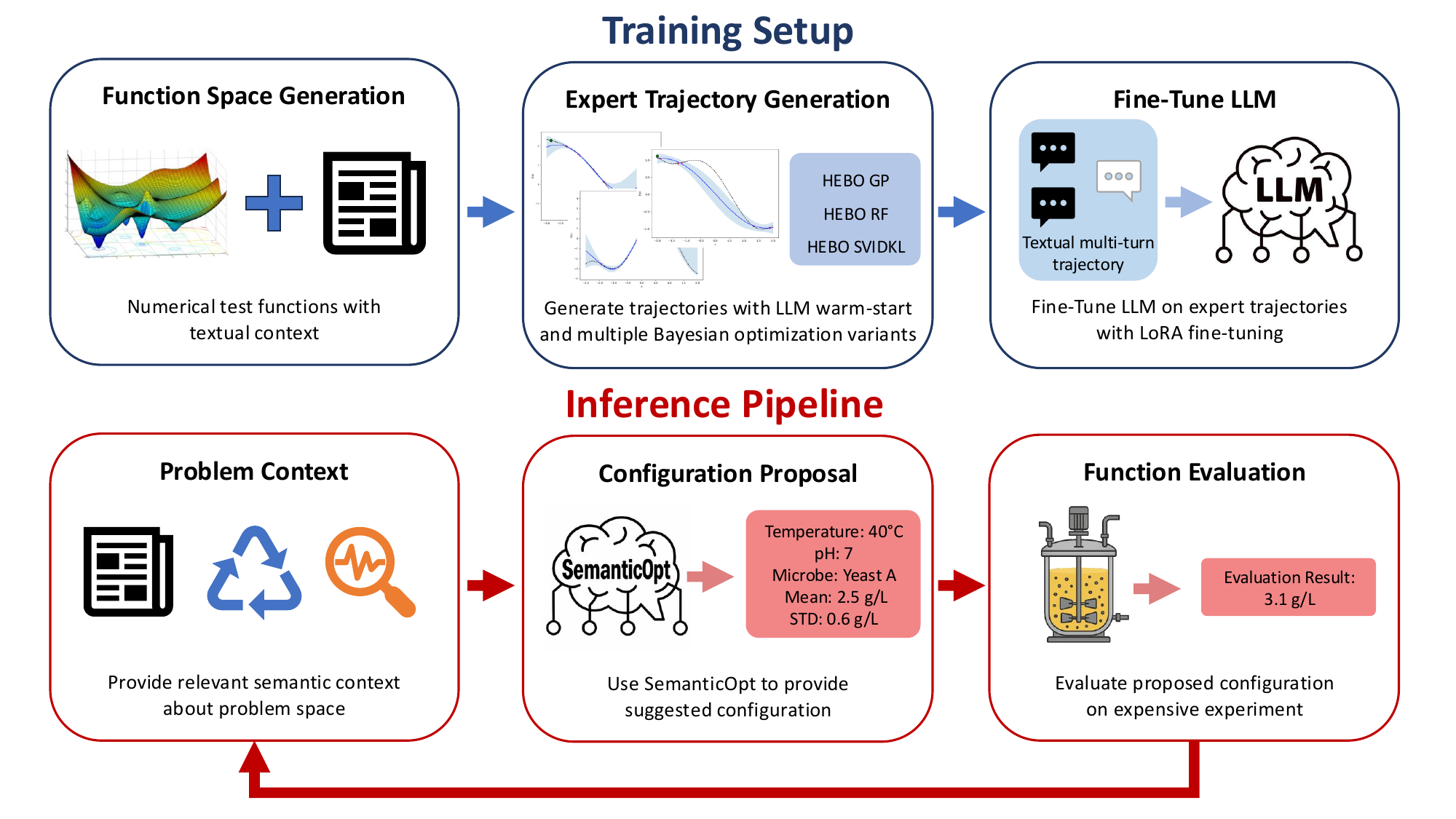}
        \captionof{figure}{Overview of SemanticOpt training and inference.}
        \label{fig:overall}
    \end{minipage}
    \hfill
    \begin{minipage}[b]{0.38\textwidth}
        \centering
        \includegraphics[width=\linewidth]{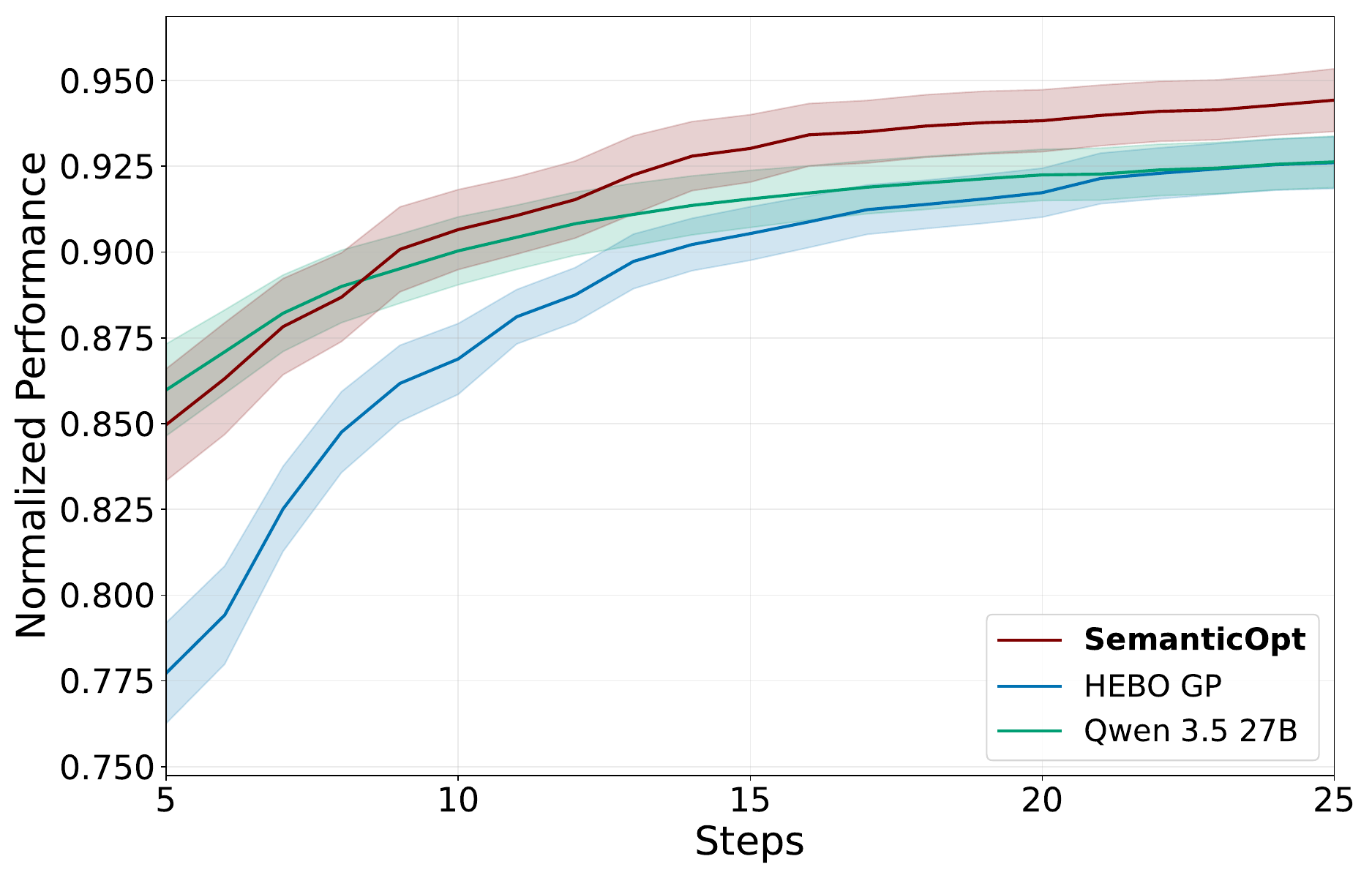}
        \captionof{figure}{Our SemanticOpt model compared to base LLM (Qwen 3.5 27B) and Bayesian optimization (HEBO GP).}
        \label{fig:llm_initialization}
    \end{minipage}

\end{figure*}

Large language models (LLMs) offer a powerful mechanism for handling contextual information but are not designed for reliable iterative optimization. Taking advantage of their knowledge base, LLMs are often capable of successful warm-starting (proposing initial experiments). However, performance often plateaus over longer trajectories, while our method, SemanticOpt, is able to iteratively improve, as shown in Figure~\ref{fig:llm_initialization}. Recent studies support these results, as they have shown that LLMs can be effective during the initialization stage of search, but alternative methods or a combination of LLM guidance with BO methods are needed for LLMs to be effective iterative optimizers~\cite{liu2024large,yuanunleashing,huang2024exploring, ferreira2026can}. This suggests that while LLMs contain useful priors for proposing strong initial candidates, they lack the exploration-exploitation balance and uncertainty-aware adaptation that makes BO effective over extended optimization horizons. Our goal is to bridge this gap by training LLMs specifically for black-box optimization, enabling them to retain their semantic reasoning advantages while improving long-horizon decision making through learned optimization policies.

We introduce SemanticOpt, an LLM-based optimizer for expensive black-box functions that enhances the power of numerical optimizers by incorporating semantic data into the optimization process. We focus on training an LLM as an end-to-end sequential optimizer that combines semantic reasoning with uncertainty-aware model-based search. Rather than relying only on prompting or using an LLM inside an external BO loop, we fine-tune the model on structured BO trajectories augmented with surrogate predictions of the black-box function and relevant semantic context. This teaches the model to use both observed evaluations and semantic information when selecting the next configuration.

Our key contributions are as follows:
\begin{enumerate}
    \item We construct a large-scale training dataset of more than 100,000 optimization trajectories over diverse Gaussian process-generated and LLM-generated function spaces, using BO methods to provide high-quality optimization data.
    \item We introduce a structured fine-tuning approach that fine-tunes LLMs on BO data to output both candidate configurations and surrogate-style predictions, creating SemanticOpt.
    \item We evaluate SemanticOpt across real-world surrogate benchmarks with varying levels of semantic context, showing average performance surpassing BO methods and LLM baselines.
\end{enumerate}

\section{Related Work}

\paragraph{Bayesian Optimization (BO):} BO is a widely used global optimization framework for expensive black-box functions because it can achieve strong sample efficiency by balancing exploration and exploitation through acquisition functions such as Log Expected Improvement~\cite{ament2023unexpected}. Additionally, BO can utilize different surrogate models, such as Gaussian processes (GPs), which may perform better in different types of landscapes and parameter spaces. BO has become broadly accessible through open-source tools including Spearmint~\cite{snoek2012practical}, BoTorch~\cite{balandat2020botorch}, AutoOED~\cite{tian2021autooed}, SMAC3~\cite{JMLR:v23:21-0888}, HEBO~\cite{cowen2022hebo}, Openbox~\cite{jiang2024openbox}, and Vizier~\cite{golovin2017google, song2024vizier}. In particular, HEBO~\cite{cowen2022hebo} has emerged as a high-quality and competitive optimizer, demonstrating strong empirical performance across benchmark suites \cite{turner2021bayesian}.

\paragraph{Learning to Optimize:} Learned black-box optimization has progressed from early RNN-based optimizers~\cite{li2016learning,andrychowicz2016learning} to transformer-based approaches enabled by larger datasets and increased compute. Early transformer methods explored online RL for task-specific solvers, including MELBA~\cite{chaybouti2022meta} and Neural Acquisition Process~\cite{maraval2024end}, while POM~\cite{li2024pretrained} targeted high-dimensional evolutionary search. More recent work emphasizes broader generalization: PFNs4BO~\cite{muller2023pfns4bo} learns surrogate-model behavior, OptFormer~\cite{chen2022towards} trains text-based transformers on proprietary optimization data, RIBBO~\cite{song2024reinforced} uses offline reinforcement learning with conditional rewards, and ZeroShotOpt~\cite{meindl2025zso} shows that transformers trained on large synthetic function sets can imitate BO trajectories. These results suggest that large-scale training can produce optimization policies with cross-task generalization. 

\paragraph{LLM-based Optimization:} Large language models have also been used directly for black-box optimization. LLAMBO~\cite{liu2024large}, Reasoning BO~\cite{yang2025reasoning}, and EvoLLM~\cite{lange2024large} show that prompting strategies, iterative interaction, and contextual information can produce competitive optimization trajectories in scenarios where LLM knowledge is strong, such as machine learning hyperparameter optimization. Additionally, LGBO~\cite{yuanunleashing} and LLINBO~\cite{chang2025llinbo} provide frameworks for integrating LLM outputs with BO acquisition functions, representing another path toward incorporating semantic information into model-based decision making. These methods highlight the promise of LLMs for using semantic information in optimization, but use the LLM as an auxiliary component in a BO pipeline and require fitting a GP in addition to LLM outputs. In contrast, our work fine-tunes an LLM directly on structured BO trajectories with semantic context, training it to produce both candidate configurations and calibrated surrogate-style predictions within a single end-to-end optimizer.

Overall, prior work suggests three complementary strengths: BO provides reliable sample-efficient search, learned optimizers can emulate desired optimization behavior, and LLMs can utilize semantic information that standard numerical optimizers ignore. SemanticOpt combines these strengths by fine-tuning an LLM on large-scale BO trajectories paired with natural-language problem context and surrogate-model predictions, creating an end-to-end semantic optimizer.

\section{Methodology}

\paragraph{Problem Framing:}
We cast optimization as a sequential decision-making problem in which an LLM policy \(\pi_\theta\) interacts with an unknown objective function \(f\) through multi-turn prompting. Each evaluation queries a configuration \(\mathbf{x}_i \in \mathcal{X}\), where \(\mathcal{X}\) may contain continuous, integer, or categorical parameters, and returns a scalar objective value \(y_i = f(\mathbf{x}_i)\). Each prompt also includes semantic context \(C\), such as problem descriptions, historical experiments, or evaluation-time feedback. At step \(t\), the LLM generates the next configuration given the observation history and semantic context:
\[
\mathcal{D}_t = \{(\mathbf{x}_i, y_i)\}_{i=1}^t,
\qquad
\mathbf{x}_{t+1} \sim \pi_\theta(\cdot \mid C, \mathcal{D}_t).
\]
The generated configuration \(\mathbf{x}_{t+1}\) is then evaluated by the objective function and appended to the history. The goal is to achieve the best configuration within the evaluation budget.

\subsection{Function Space Generation}

Because widely available real-world benchmark functions are limited, we construct a large-scale training dataset using realistic and diverse synthetic objectives. We generate two classes of evaluation functions for this purpose: Gaussian process (GP)-based functions and LLM-generated function spaces. Together, these sources provide both scalable synthetic diversity and semantically grounded problem structures. GPs serve as flexible function generators that represent a wide variety of complex functions, enabling large-scale data collection while producing rich optimization landscapes. The motivation is that GP-based Bayesian optimization methods are most widely used as they perform well across many function classes \cite{shahriari2015taking, wang2023recent}. Therefore, by training on GP-generated landscapes, our objective is to learn a policy that can be similarly generalized in diverse problem spaces. To further diversify these tasks, we vary both input and output scales to reflect realistic regimes. For example, objectives may lie in ranges such as $[0,100]$ or $[1,10]$, while parameters may span $[10^{-4}, 10^{-1}]$. We augment sampled GP functions with nonlinear coordinate warping, discontinuities, and constraints, transforming smooth samples into landscapes that better resemble real-world systems. We also add failure regions into the landscape, so the model learns to account for experiment failures.

We also use LLM-generated function spaces to introduce real-world domains into the dataset. To do this, we prompt Qwen 3.5 35B-A3B to produce semantic context of real-world black-box optimization problems, along with 100 evaluated points in the function space. We show two examples of this generated context in Figure~\ref{fig:semantic_context_examples}. We choose this model for its combination of speed and capabilities. By prompting the model to produce example problems with various types of semantic data included, we aim to create training data that matches the diversity of semantic information available in real-world problems. We define the ground truth function space by fitting a surrogate model to these proposed points. By combining GP-generated and LLM-generated objectives, we create a broader and more representative training distribution for model fine-tuning.

\paragraph{Trajectory Generation:}

Using our generated function spaces, we generate sequences of proposed configurations and evaluated points, which we call trajectories. We use HEBO~\cite{cowen2022hebo}, a state-of-the-art BO framework, to generate training trajectories because we need a high-quality optimizer that includes reliable surrogate modeling capabilities. It has also demonstrated strong empirical performance, including first place in the NeurIPS 2020 Black-Box Optimization Challenge~\cite{turner2021bayesian}. We generate our training dataset by running HEBO on our generated function spaces with different surrogate models. We begin by initializing our process with 5 initial steps. We use Sobol sampling~\cite{sobol1967distribution} for the GP-based functions and use LLM-generated initializations for the functions with semantic information. We then proceed with using 3 different surrogate models implemented in HEBO, including Gaussian process (GP), random forest (RF), and Stochastic Variational Inference Deep Kernel Learning (SVIDKL). To compile the dataset, we select the best performing surrogate model after 10, 25, 50, and 100 total steps. Therefore, the model is trained to act like the best BO surrogate in each scenario, with the goal of performing better than any individual method.

\begin{figure*}[t]
\centering
\small

\begin{minipage}[t]{0.49\textwidth}
\begin{tcolorbox}[
    enhanced,
    equal height group=semanticexamples,
    colback=gray!5,
    colframe=black!25,
    boxrule=0.6pt,
    arc=2mm,
    left=1.5mm,
    right=1.5mm,
    top=1.2mm,
    bottom=1.2mm,
    title=\textbf{Vertical Farm Spectrum Optimization}
]

\textbf{Task:} Design an LED lighting recipe for indoor crop growth.

\textbf{Inputs:} Spectrum ratios, PPFD target, photoperiod length, light cycle type, and far-red pulse use.

\textbf{Extra Evaluation Context:} Each trial also returns stem height index, canopy penetration efficiency, and total energy cost.

\textbf{Objective:} Maximize biomass yield efficiency.
\end{tcolorbox}
\end{minipage}
\hfill
\begin{minipage}[t]{0.49\textwidth}
\begin{tcolorbox}[
    enhanced,
    equal height group=semanticexamples,
    colback=gray!5,
    colframe=black!25,
    boxrule=0.6pt,
    arc=2mm,
    left=1.5mm,
    right=1.5mm,
    top=1.2mm,
    bottom=1.2mm,
    title=\textbf{Dynamic Shipping Time Calibration}
]

\textbf{Task:} Tune a shipping policy to reduce average delivery time.

\textbf{Inputs:} Free-shipping limit, express cost, and max wait time before switching carriers.

\textbf{Historical Context:} Other markets found that a medium express cost, high free-shipping limit and long switch-wait lowered delivery time by 18\%.

\textbf{Objective:} Minimize average shipping time.
\end{tcolorbox}
\end{minipage}

\caption{Example problems with extra evaluation metrics and historical context from the generated training dataset. This context is used to begin trajectories with LLM-generated initializations before switching to HEBO methods to generate high-quality examples for LLM fine-tuning.}
\label{fig:semantic_context_examples}
\end{figure*}

\paragraph{Prompt Structure:} An example of the interaction loop is shown in Figure~\ref{fig:opt_prompt_full}. The prompt begins with a system instruction defining the agent's role as a black-box optimization assistant and enforcing a strict JSON output format containing the action and predicted mean and standard deviation. We include a failure probability field after the optimizer has observed at least one failed evaluation. The user then provides the problem description and relevant context. We provide all numbers rounded to 4 significant figures and convert the BO steps into a multi-turn user and assistant conversation. The interaction follows in a turn-based manner where the assistant proposes a new configuration, which is the action proposed by BO and the surrogate model's predicted mean and standard deviation, and the user returns the evaluated objective value, repeating until the evaluation budget is exhausted. Further examples and details of dataset generation are available in Appendix~\ref{app:dataset}.

\begin{figure*}[h]
\centering
\begin{lstlisting}[
    style=dialogue
]
User: Problem description: Financial Trend Prediction Model Hyperparameter Tuning. Optimize the hyperparameters of a LSTM network designed to predict short-term stock market trends. 
Tunable search space:
- lstm_units: integer (log search space) in range [16, 256] | description: The number of hidden units in the LSTM layer, controlling model capacity. | unit: units
- learning_rate: float (log search space) in range [0.0001, 0.05] | description: The step size for the optimizer during model training. | unit: dimensionless
Objective:
- Direction: maximize
- Name: risk_adjusted_return
- Unit: ratio
- Description: The Sharpe Ratio calculated from simulated trades based on the model's trend predictions.
Budget: - Total evaluations available (including 5 initial evaluations): 50
Provide 5 initial configurations that explore promising regions in the search space.
Assistant: {"initial_configuration_1": {"lstm_units": 256, "learning_rate": 0.05}, ..., "initial_configuration_5": {"lstm_units": 16, "learning_rate": 0.0001}}
User: Initial evaluations:
Evaluation 1: Configuration: {"lstm_units": 256, "learning_rate": 0.05}
Objective (risk_adjusted_return): 0.964 ratio
<Evaluations 2 - 4 omitted>
Evaluation 5: Configuration: {"lstm_units": 16, "learning_rate": 0.0001}
Objective (risk_adjusted_return): 0.45 ratio
Output the next configuration.
Assistant: {"next_configuration": {"lstm_units": 116, "learning_rate": 0.00516}, "mean": 1.242, "std": 0.0854}
User: Evaluation 6:
Objective (risk_adjusted_return): 1.302 ratio
\end{lstlisting}

\caption{An example problem with semantic context from our training data. We use these text-based sequences to fine-tune the model and use this structure at inference time to sample new configurations.}
\label{fig:opt_prompt_full}

\end{figure*}

\subsection{Training}

Our experiments utilize the Qwen 3.5 9B and 27B models \cite{team2026qwen3}. We chose these models for their balance of instruction-following ability and computational accessibility for fine-tuning. Fine-tuning is performed via Low-Rank Adaptation (LoRA)~\cite{hu2022lora} leveraging Unsloth~\cite{unsloth}. We use LoRA fine-tuning to allow the model to maintain its knowledge base while adding the ability to model BO trajectories. We fine-tune Qwen 3.5 27B on over 100,000 trajectories, with half of the data containing LLM-generated context. This takes around 2.5 days for the 27B model on 4 Nvidia H200 GPUs.

\subsection{Inference}

At inference time, we use our fine-tuned model with the same context structure as training time. We begin our initialization prompt with the relevant context and ask the model to produce five initialization configurations. We then provide the evaluations for these initial points and iteratively propose and evaluate configurations until the evaluation budget is reached. For each proposal, we sample two possible configurations from the LLM and select the configuration with the higher expected improvement with failure conditioning from the model's predicted distribution. More details about the training and inference process are available in Appendix~\ref{app:train}.

\section{Results}

\subsection{Baselines and Benchmarks}

We compare SemanticOpt against a comprehensive set of baselines, including Bayesian Optimization (BO) approaches and transformer and LLM-based methods. We utilize implementations from Optuna~\cite{optuna_2019} and OptunaHub~\cite{ozaki2025optunahub} for the following primary baselines:
\begin{itemize}
    \item \textbf{BO:} 
    HEBO~\cite{cowen2022hebo} (with GP, RF, and SVIDKL surrogate models)
\item \textbf{Transformer or LLM-based:} LLAMBO~\cite{liu2024large}, PFNs4BO~\cite{muller2023pfns4bo}, GPT-5.4~\cite{singh2025openai}.
\end{itemize}
To isolate the impact of our fine-tuning, we also evaluate the Qwen 3.5 27B model~\cite{team2026qwen3} using the same prompts as SemanticOpt but without fine-tuning the model. These methods represent the current state-of-the-art in Bayesian optimization and LLM-based optimization \cite{turner2021bayesian, singh2025openai}. More information about baselines is available in Appendix~\ref{app:baselines}. We report mean normalized performance,
\begin{equation*}
    P = 1 - |f_{\text{best}} - f^*| / |f_m - f^*|.
\end{equation*}
Here, $f_{\text{best}}$ is the best sampled value, $f^*$ is the global optimum, and $f_m$ is the median random-sampling value. This normalizes results by the random-sampling optimality gap.

\paragraph{Benchmarks:}
Due to the lack of real-world black-box optimization benchmarks with included semantic data, we compile a benchmark of diverse problems on which to evaluate our model. We evaluate on out-of-distribution black-box optimization benchmarks and surrogate benchmarks constructed from real-world experimental data. For the surrogate benchmarks, we fit fast evaluation models offline from tabular datasets of previously evaluated configurations obtained through Sobol sampling of the function space or using publicly available tabular datasets. For each target objective, we select the best surrogate from a fixed candidate set including XGBoost, CatBoost, Gaussian processes, and tree-based models, using repeated randomized train-validation splits. The selected surrogate is then used during evaluation as a fast proxy for the real-world task. To support semantic optimization, each benchmark is augmented with contextual information. At minimum, this includes problem, parameter, and objective descriptions. We also optionally include long-form context such as papers or documentation, history examples from related tasks, and extra evaluation metrics beyond the main objective. We show the included benchmark domains and context in Table~\ref{tab:benchmarks}. More information about benchmark creation is available in Appendix~\ref{app:benchmark_descriptions}.

\begin{table*}[h]
\centering
\small
\caption{Summary of benchmark tasks and semantic data used in the evaluation.}
\label{tab:benchmarks}
\resizebox{\textwidth}{!}{%
\begin{tabular}{p{0.20\textwidth} p{0.27\textwidth} p{0.09\textwidth} p{0.12\textwidth} p{0.31\textwidth}}
\toprule
\textbf{Benchmark} & \textbf{Task description} & \textbf{\# Params} & \textbf{Param Type} & \textbf{Semantic data available} \\
\midrule
Airfoil CFD
& Shape optimization for OpenFoam~\cite{jasak2009openfoam} CFD simulation
& 10
& Continuous, Categorical, Integer
& Geometric parameter definitions, simulation descriptions, and drag/lift objective information \\

Convex Decomposition
& Optimization of parameters for decomposing shapes into convex components
& 8
& Continuous, Categorical, Integer
& Parameter meanings and objective descriptions \\

Cookie Recipe~\cite{solnik2017bayesian}
& Recipe optimization over ingredients and preparation 
& 10
& Continuous, Categorical
& Ingredient descriptions and historical recipe examples \\

Iron Mind~\cite{macknight2025pre}
& Reaction-optimization based on experimental data
& Varies
& Categorical
& Reaction descriptions and related paper context \\

MuJoCo Control~\cite{todorov2012mujoco}
& Hyperparameter or controller-parameter optimization for continuous-control tasks
& Varies
& Continuous, Categorical, Integer
& Environment descriptions, parameter names, and evaluation metrics \\

Nasbench201~\cite{dong2020bench}
& Neural architecture search over architectures
& Varies
& Categorical
& Parameter descriptions and test performance objectives \\

NeqSim~\cite{neqsim2026software}
& Optimization of process-simulation configurations
& Varies
& Continuous, Integer
& Process-simulation context and objective definitions \\

Paint Mix
& Optimization of paint-mixture ratios to match target color
& 5
& Continuous
& Color description and target-paint information \\

PD1~\cite{JMLR:v25:23-0269}
& Hyperparameter optimization for neural network training tasks
& Varies
& Continuous, Categorical, Integer
& Parameter names, training descriptions, and auxiliary training metrics \\

\bottomrule
\end{tabular}
}
\end{table*}

\begin{figure}[h]
    \centering
    \includegraphics[width=0.95\linewidth]{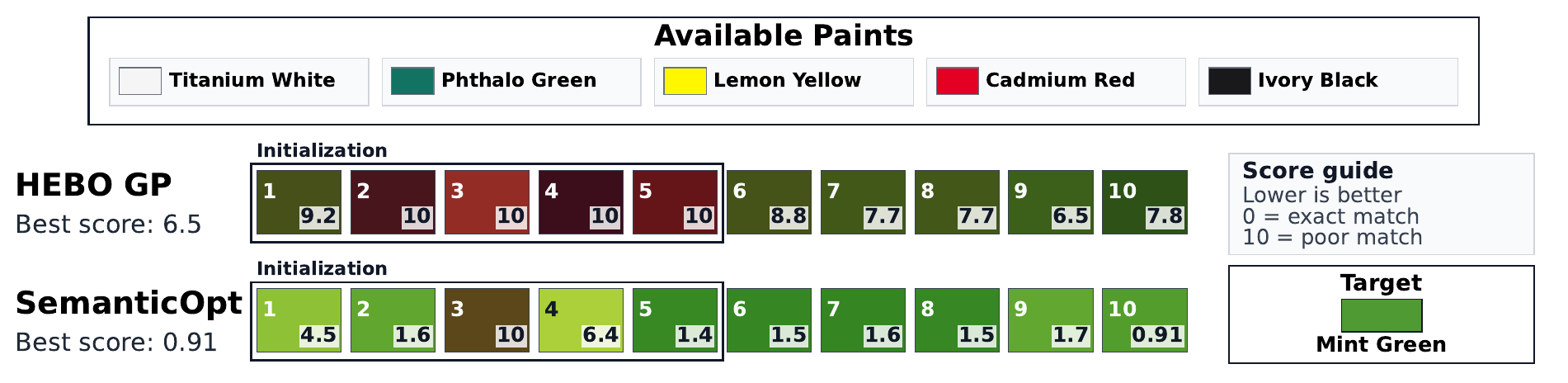}
    \caption{Demonstration of finding the best mixing ratio to match the target paint. SemanticOpt shows better initialization and performance given the semantic context than HEBO GP.}
    \label{fig:paint_mix_examples}
\end{figure}

\paragraph{Semantic Example:} To demonstrate the usefulness and applicability of semantic data in black-box optimization, we show a simple paint-mixing problem. This problem involves the task of mixing a set of 5 possible paints together to achieve a target color. The idea here is that we do not know the formula for each paint and how they mix together and dry to form the eventual color. However, we can provide semantic information about the available paints and the target paint to our LLM-based approaches. For example, we can provide the color: "Phthalo Green" and a description: "strong vivid green paint" within the prompt to the model. This context allows the model to have more information than just numerical data, allowing the LLM-based approaches to outperform BO methods. We show an example of 10 evaluations of this paint mixing comparing HEBO with a GP surrogate model against our SemanticOpt in Figure~\ref{fig:paint_mix_examples}, demonstrating that the semantic context we can provide to SemanticOpt results in a much more direct and accurate search than BO can achieve. 

\subsection{Performance Comparison}

\begin{figure*}[h]
    \centering
    
    \begin{subfigure}[b]{0.45\textwidth}
        \centering
        \includegraphics[width=\linewidth]{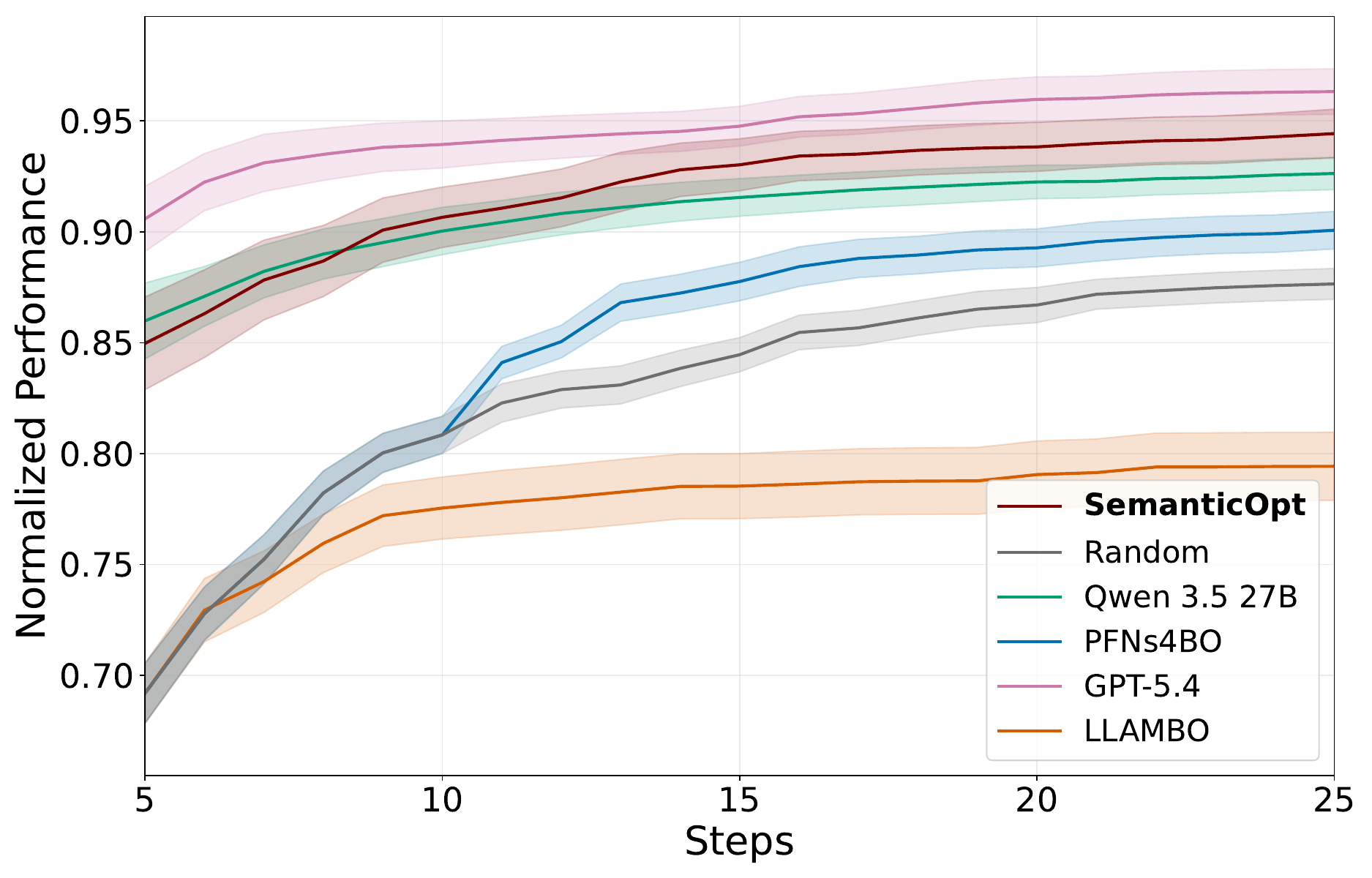}
        \caption{PD1 Results}
        \label{fig:pd1}
    \end{subfigure}
    \hfill
    \begin{subfigure}[b]{0.45\textwidth}
        \centering
        \includegraphics[width=\linewidth]{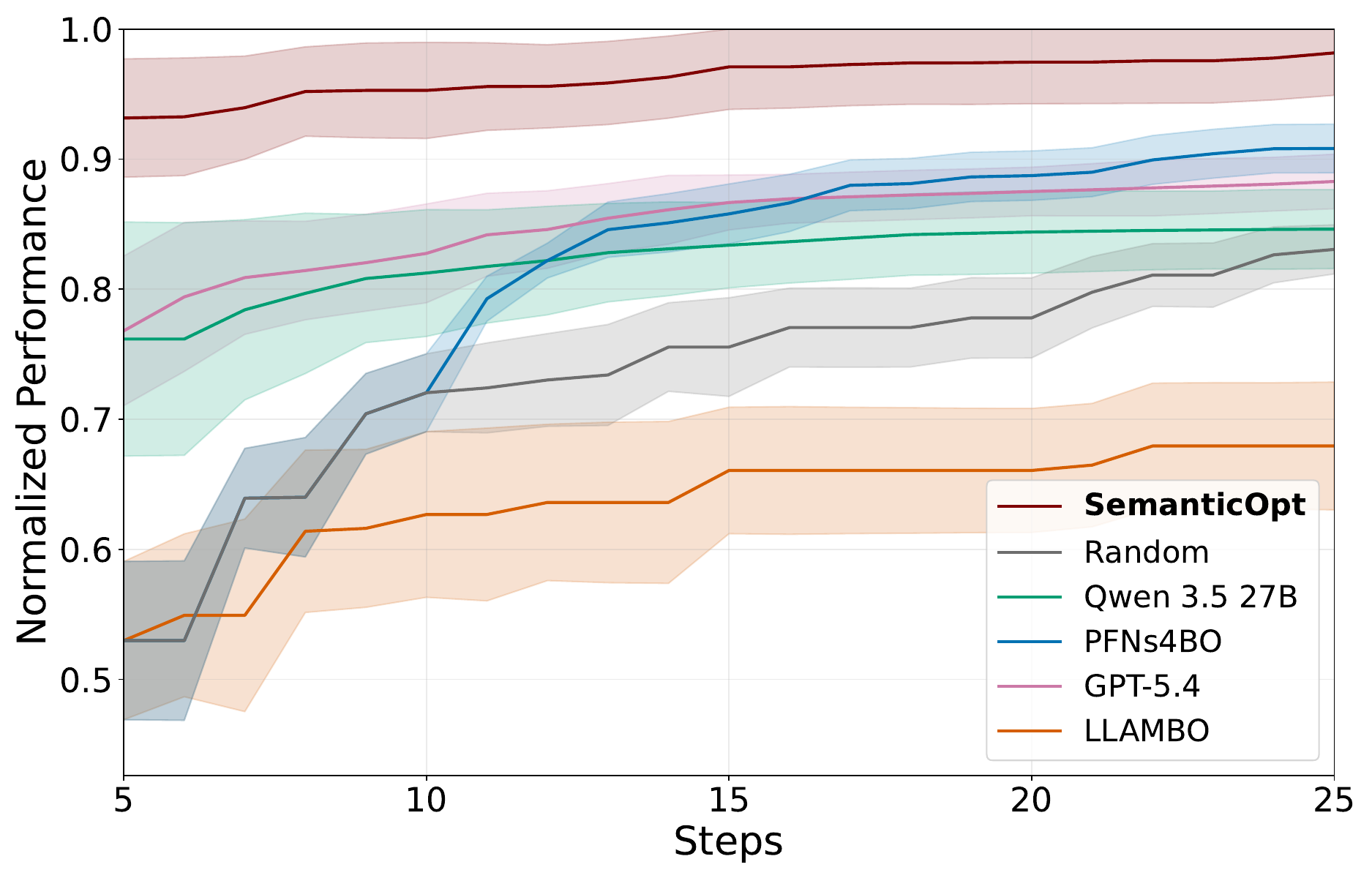}
        \caption{MuJoCo Control Results}
        \label{fig:mujoco}
    \end{subfigure}

    \begin{subfigure}[b]{0.45\textwidth}
        \centering
        \includegraphics[width=\linewidth]{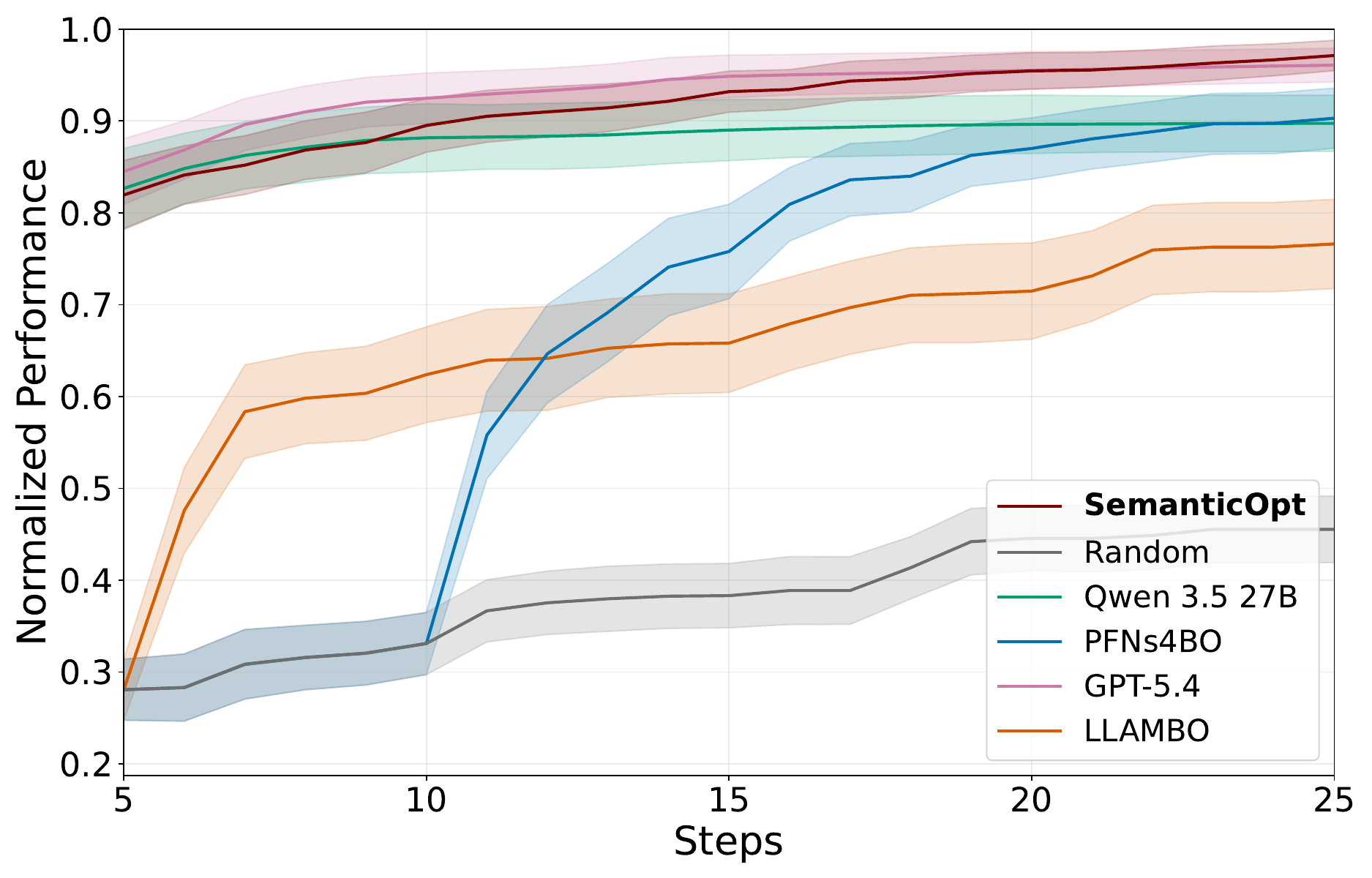}
        \caption{Paint Mixing Results}
        \label{fig:paint_mix}
    \end{subfigure}
    \hfill
    \begin{subfigure}[b]{0.45\textwidth}
        \centering
        \includegraphics[width=\linewidth]{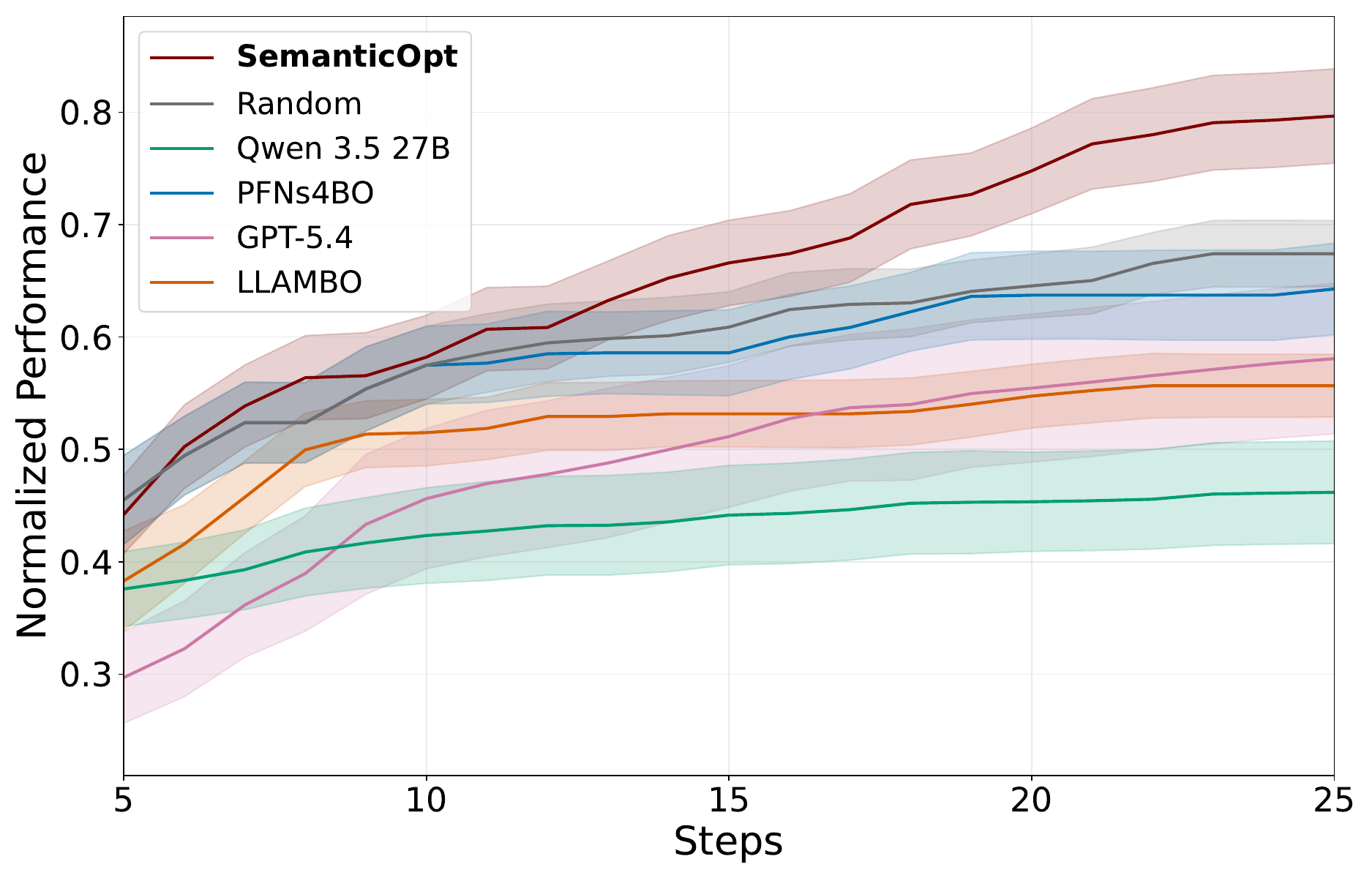}
        \caption{Airfoil Results}
        \label{fig:airfoil}
    \end{subfigure}

    \caption{Mean normalized performance over steps with paired standard error of the mean (SEM) of SemanticOpt compared to other LLM and transformer-based methods. SemanticOpt shows a large increase in iterative optimization capabilities relative to the base Qwen 3.5 27B model.}
    \label{fig:step_results}
\end{figure*}

We show the performance of SemanticOpt over steps against other LLM-based methods on representative benchmarks in Figure~\ref{fig:step_results}. These results are gathered over 100 random problem instances for PD1, 10 for MuJoCo Control, and 25 for Paint Mixing and Airfoil CFD results. We show the mean normalized performance with paired standard error of the mean (SEM) over these instances. We see that LLM-based methods provide better initializations than random sampling, but the base LLMs struggle to continue to improve over further steps, while SemanticOpt performs better as an iterative optimizer. In particular, SemanticOpt shows improved iterative performance after initialization when compared to Qwen 3.5 27B, which is the base model we fine-tuned. We show results with more step curves in Appendix~\ref{app:step}, win-rates across methods in Appendix~\ref{app:win_rate}, and longer trajectories in Appendix~\ref{app:length}. Lastly, we show SemanticOpt's surrogate model performance in Appendix~\ref{app:surrogate}.

\subsection{Ablations}

\paragraph{Comparison with Frontier Models and Data Leakage:}

To further evaluate SemanticOpt, we provide comparisons with OpenAI GPT-5.4~\cite{singh2025openai}. However, providing comparisons with a large frontier model brings concern about data leakage and result memorization. Table~\ref{tab:further_results_final_scores} reports final normalized performance after 25 evaluations across additional benchmarks. The values presented are mean normalized performance with paired SEM over 100 benchmark problems for PD1, 10 for MuJoCo Control, and 25 for all other benchmarks. SemanticOpt achieves the best average performance across benchmarks, outperforming all HEBO variants and GPT-5.4 on mean score.

\begin{table*}[h]
\centering
\caption{Final normalized performance after 25 steps. Values reported are mean $\pm$ SEM across benchmark instances. The best performance is bolded. SemanticOpt achieves the best mean performance over benchmarks and is the best performing method over three datasets generated by us (Airfoil CFD, MuJoCo Control, Paint Mix). We see that GPT-5.4 achieves the best performance on three of the publicly available datasets (Cookie Recipe, Nasbench201, PD1), suggesting possible memorization.}
\label{tab:further_results_final_scores}
\resizebox{\textwidth}{!}{%
\begin{tabular}{lcccccccc}
\toprule
\textbf{Benchmark}
& \textbf{HEBO-GP}
& \textbf{HEBO-RF}
& \textbf{HEBO-SVIDKL}
& \textbf{PFNs4BO}
& \textbf{LLAMBO}
& \textbf{Qwen 3.5 27B}
& \textbf{GPT-5.4}
& \textbf{SemanticOpt} \\
\midrule
Airfoil CFD
& $0.738 \pm 0.040$
& $0.697 \pm 0.032$
& $0.702 \pm 0.034$
& $0.643 \pm 0.037$
& $0.557 \pm 0.024$
& $0.462 \pm 0.045$
& $0.581 \pm 0.072$
& $\mathbf{0.797 \pm 0.041}$ \\

Convex Decomposition
& $0.861 \pm 0.027$
& $0.822 \pm 0.021$
& $0.760 \pm 0.022$
& $\mathbf{0.864 \pm 0.018}$
& $0.718 \pm 0.024$
& $0.808 \pm 0.038$
& $0.858 \pm 0.033$
& $0.848 \pm 0.020$ \\

Cookie Recipe
& $0.837 \pm 0.024$
& $0.525 \pm 0.033$
& $0.701 \pm 0.024$
& $0.438 \pm 0.030$
& $0.644 \pm 0.013$
& $0.861 \pm 0.011$
& $\mathbf{0.872 \pm 0.011}$
& $0.860 \pm 0.019$ \\

Iron Mind
& $0.941 \pm 0.017$
& $0.942 \pm 0.013$
& $0.910 \pm 0.016$
& $0.840 \pm 0.028$
& $0.901 \pm 0.015$
& $\mathbf{0.974 \pm 0.017}$
& $0.958 \pm 0.020$
& $0.939 \pm 0.015$ \\

MuJoCo Control
& $0.933 \pm 0.017$
& $0.897 \pm 0.023$
& $0.946 \pm 0.022$
& $0.908 \pm 0.017$
& $0.680 \pm 0.056$
& $0.846 \pm 0.027$
& $0.883 \pm 0.017$
& $\mathbf{0.982 \pm 0.027}$ \\

Nasbench201
& $0.803 \pm 0.023$
& $0.852 \pm 0.019$
& $0.859 \pm 0.015$
& $0.817 \pm 0.031$
& $0.643 \pm 0.044$
& $0.956 \pm 0.016$
& $\mathbf{0.999 \pm 0.019}$
& $0.931 \pm 0.015$ \\

NeqSim
& $0.934 \pm 0.020$
& $0.909 \pm 0.025$
& $\mathbf{0.976 \pm 0.017}$
& $0.968 \pm 0.020$
& $0.606 \pm 0.037$
& $0.665 \pm 0.047$
& $0.922 \pm 0.018$
& $0.866 \pm 0.022$ \\

Paint Mix
& $0.871 \pm 0.042$
& $0.800 \pm 0.046$
& $0.894 \pm 0.017$
& $0.903 \pm 0.036$
& $0.766 \pm 0.051$
& $0.897 \pm 0.027$
& $0.961 \pm 0.016$
& $\mathbf{0.971 \pm 0.014}$ \\

PD1
& $0.926 \pm 0.007$
& $0.915 \pm 0.005$
& $0.907 \pm 0.006$
& $0.901 \pm 0.009$
& $0.794 \pm 0.016$
& $0.926 \pm 0.007$
& $\mathbf{0.963 \pm 0.010}$
& $0.944 \pm 0.011$ \\
\midrule
\textbf{Mean Score}
& $0.872$
& $0.818$
& $0.851$
& $0.809$
& $0.701$
& $0.822$
& $0.889$
& $\mathbf{0.904}$ \\
\bottomrule
\end{tabular}
}
\end{table*}

GPT-5.4 has very strong initializations on benchmarks composed of publicly available datasets (ex: PD1, Nasbench201, Cookie Recipe, NeqSim), while the model has consistently poor performance on other benchmarks (ex: Airfoil CFD, MuJoCo Control). The PD1~\cite{JMLR:v25:23-0269} and Nasbench201~\cite{dong2020bench} datasets are generated from publicly available tabular datasets on machine learning hyperparameter optimization. The Cookie Recipe benchmark~\cite{solnik2017bayesian} was created with a surrogate fit to the exact best cookie found in the paper. Lastly, the NeqSim benchmark was created using open-source simulation software that includes examples of similar simulations online. However, in datasets generated through simulations whose data points are not seen online, such as Airfoil CFD, MuJoCo Control, and paint mixing, SemanticOpt performs the best of any method, including GPT-5.4. We created these benchmarks by identifying and implementing problems in these domains, evaluating a large number of randomly selected samples, and fitting a surrogate model to the evaluations. These examples have far fewer online examples than the previous set and are less likely to contain memorization.

\begin{figure*}[h]
    \centering
    
    \begin{subfigure}[b]{0.32\textwidth}
        \centering
        \includegraphics[width=\linewidth]{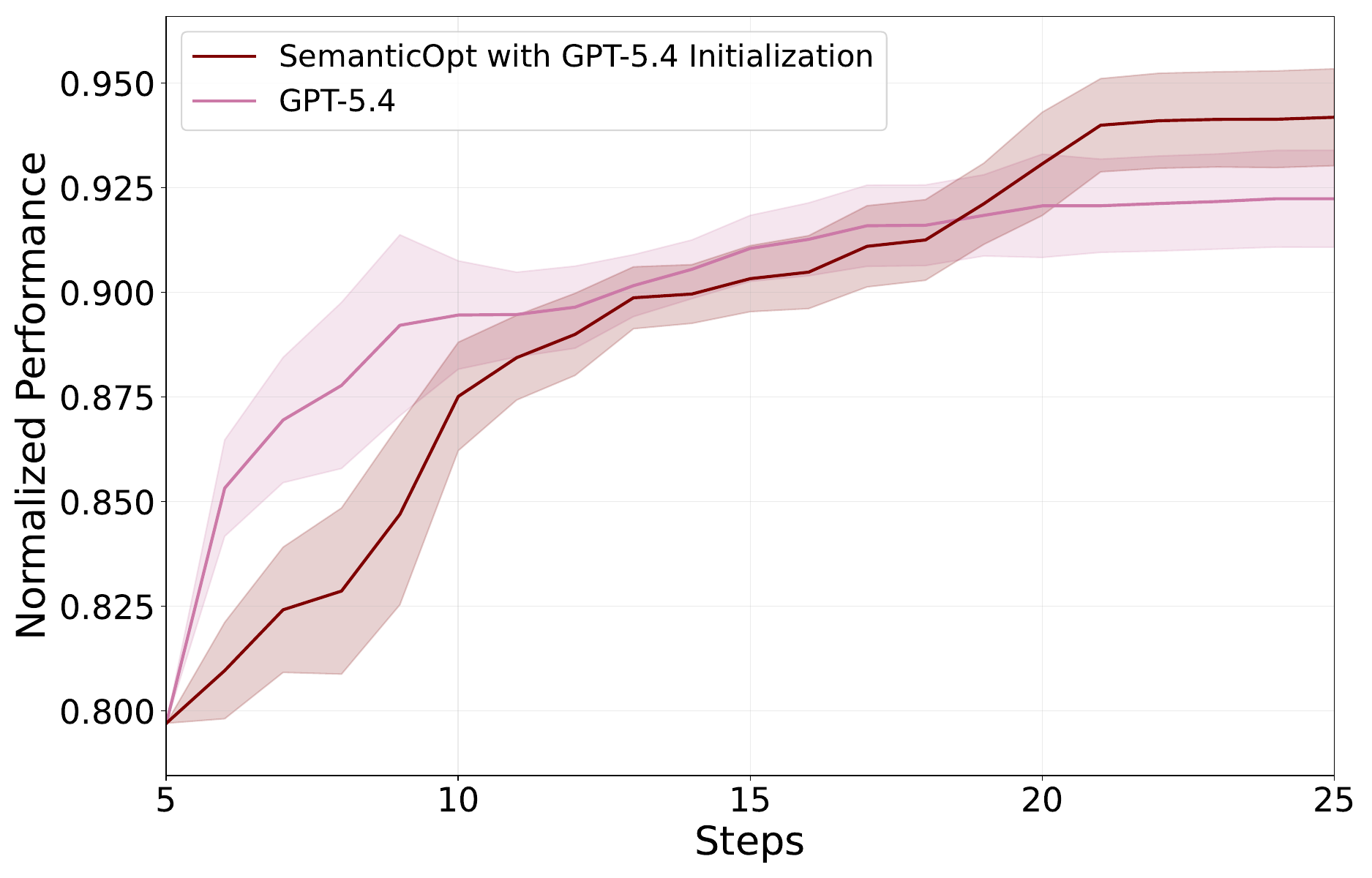}
        \caption{NeqSim Comparison}
        \label{fig:neqsim_comp}
    \end{subfigure}
    \begin{subfigure}[b]{0.32\textwidth}
        \centering
        \includegraphics[width=\linewidth]{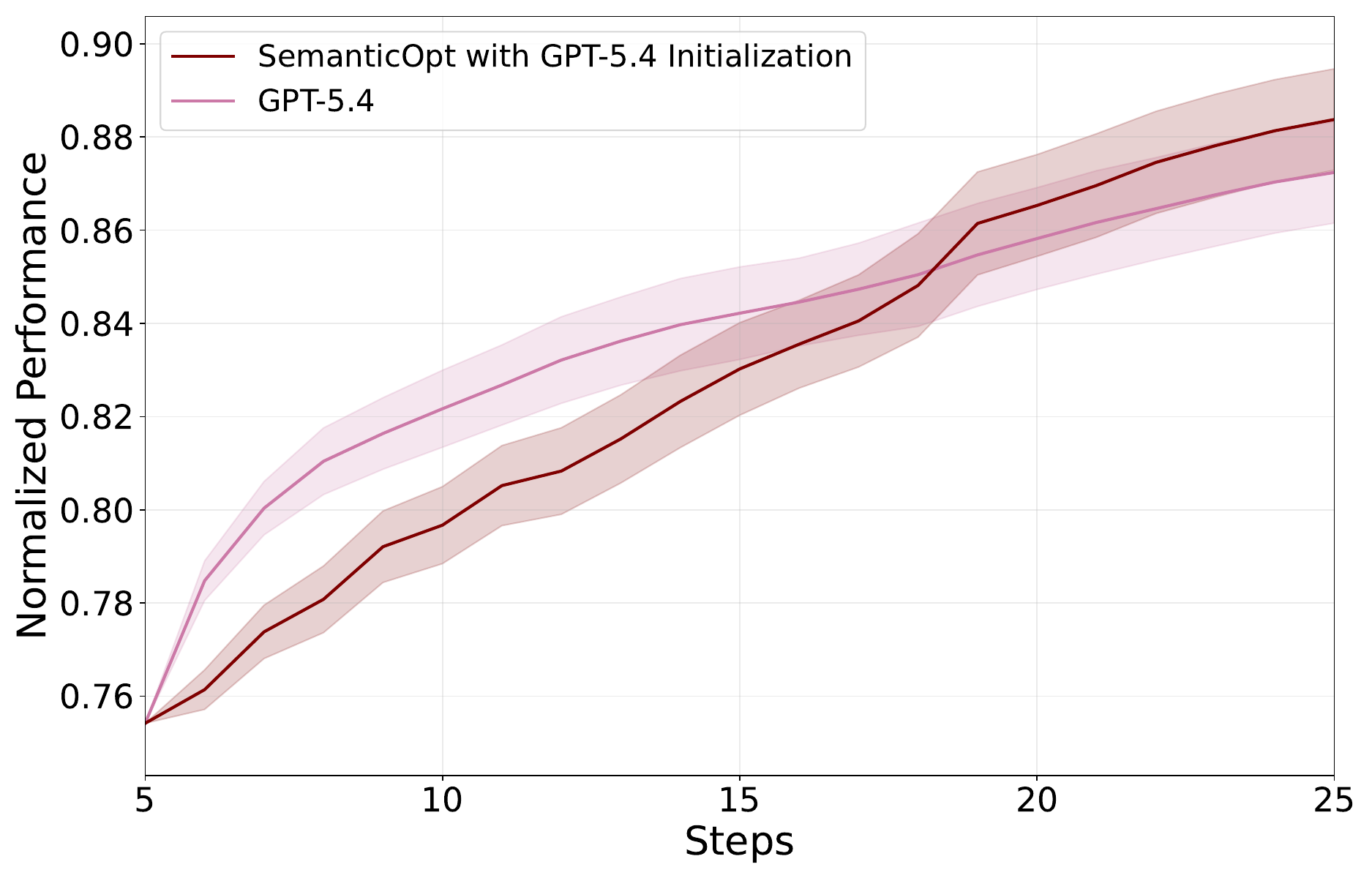}
        \caption{Cookie Recipe Comparison}
        \label{fig:cookie_comp}
    \end{subfigure}
    \begin{subfigure}[b]{0.32\textwidth}
        \centering
        \includegraphics[width=\linewidth]{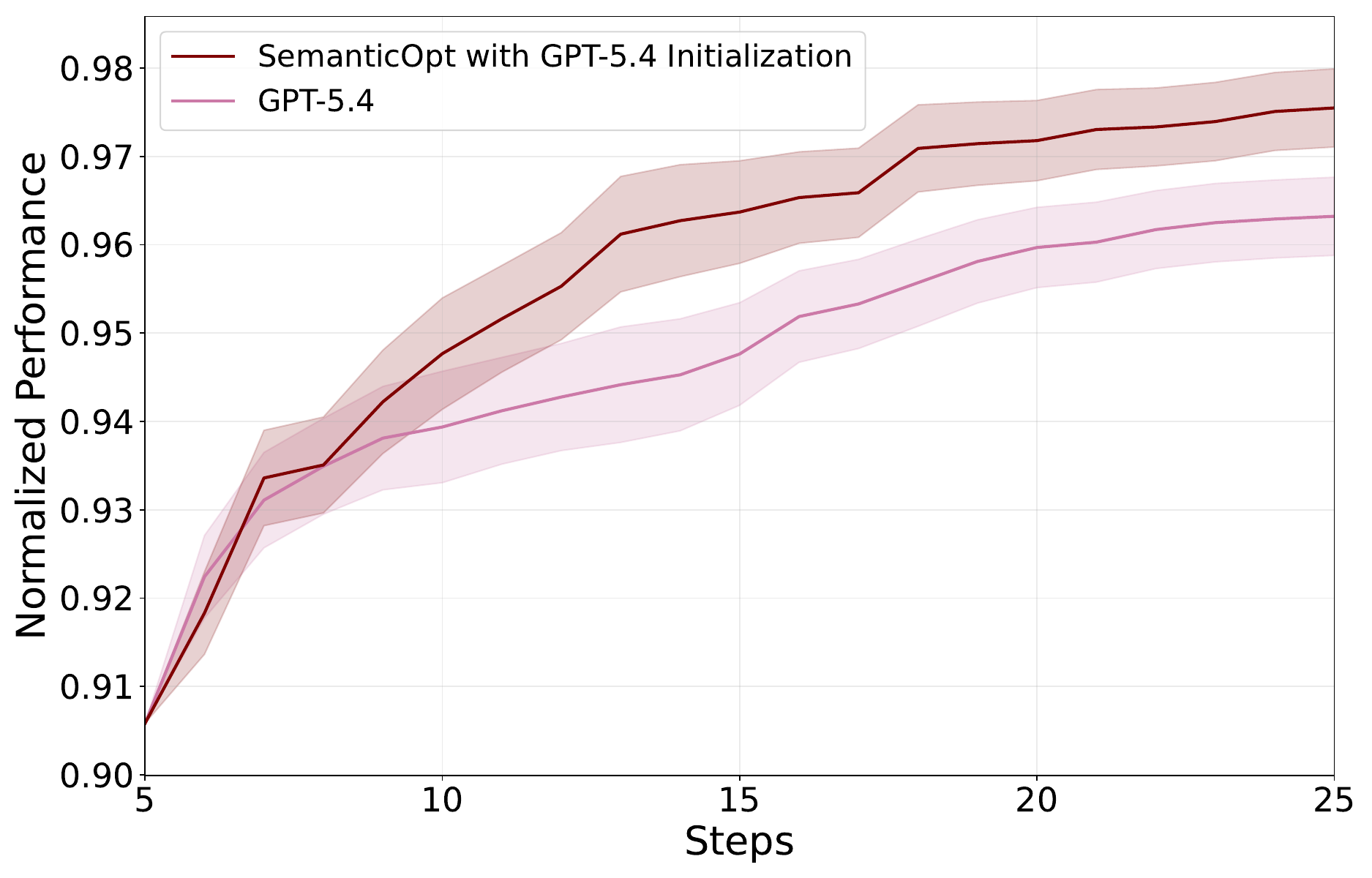}
        \caption{PD1 Comparison}
        \label{fig:pd1_comp}
    \end{subfigure}

    \caption{Mean normalized performance with paired SEM of SemanticOpt with GPT-5.4 initialization. SemanticOpt outperforms GPT-5.4 when given the same initialization, showing strong iterative optimization capabilities, even on benchmarks GPT-5.4 may have memorized.}
    \label{fig:gpt_initialization}
\end{figure*}

To isolate the iterative performance of SemanticOpt compared to GPT-5.4 and reduce the impact of possible memorization, we show the performance of SemanticOpt when given the five initialization steps generated by GPT-5.4 on benchmarks where GPT-5.4 outperformed SemanticOpt in Figure~\ref{fig:gpt_initialization}. SemanticOpt follows a different search pattern than GPT-5.4, initially matching or performing worse but obtaining better results after 25 steps on the benchmarks. This shows that the performance benefit of GPT-5.4 on these benchmarks comes from the initialization, which may use some memorization.

\paragraph{Semantic Data:}

\begin{figure*}[t]
    \centering
    
    \begin{subfigure}[b]{0.32\textwidth}
        \centering
        \includegraphics[width=\linewidth]{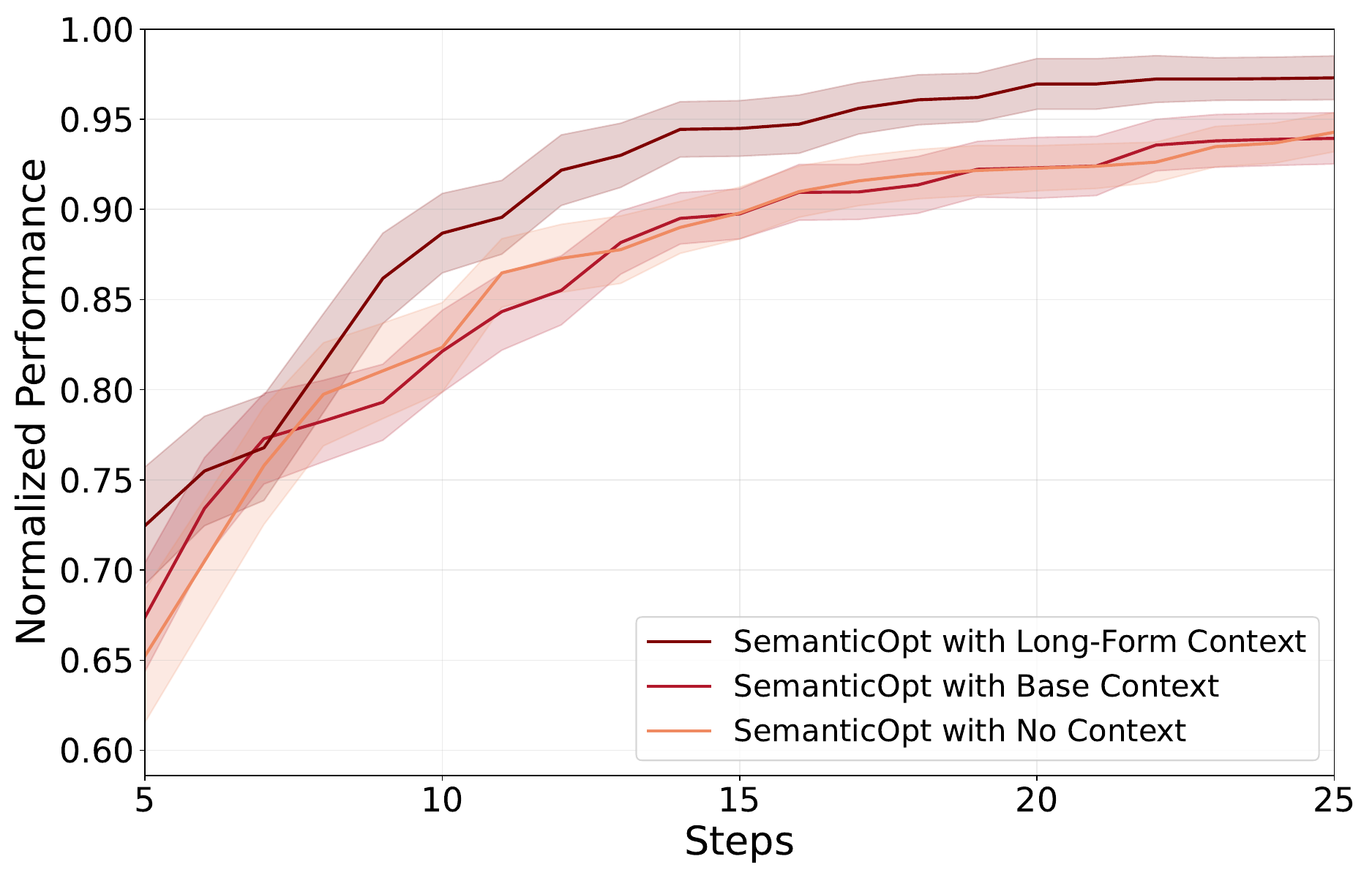}
        \caption{Iron Mind Long Context}
        \label{fig:ironmind_sem}
    \end{subfigure}
    \begin{subfigure}[b]{0.32\textwidth}
        \centering
        \includegraphics[width=\linewidth]{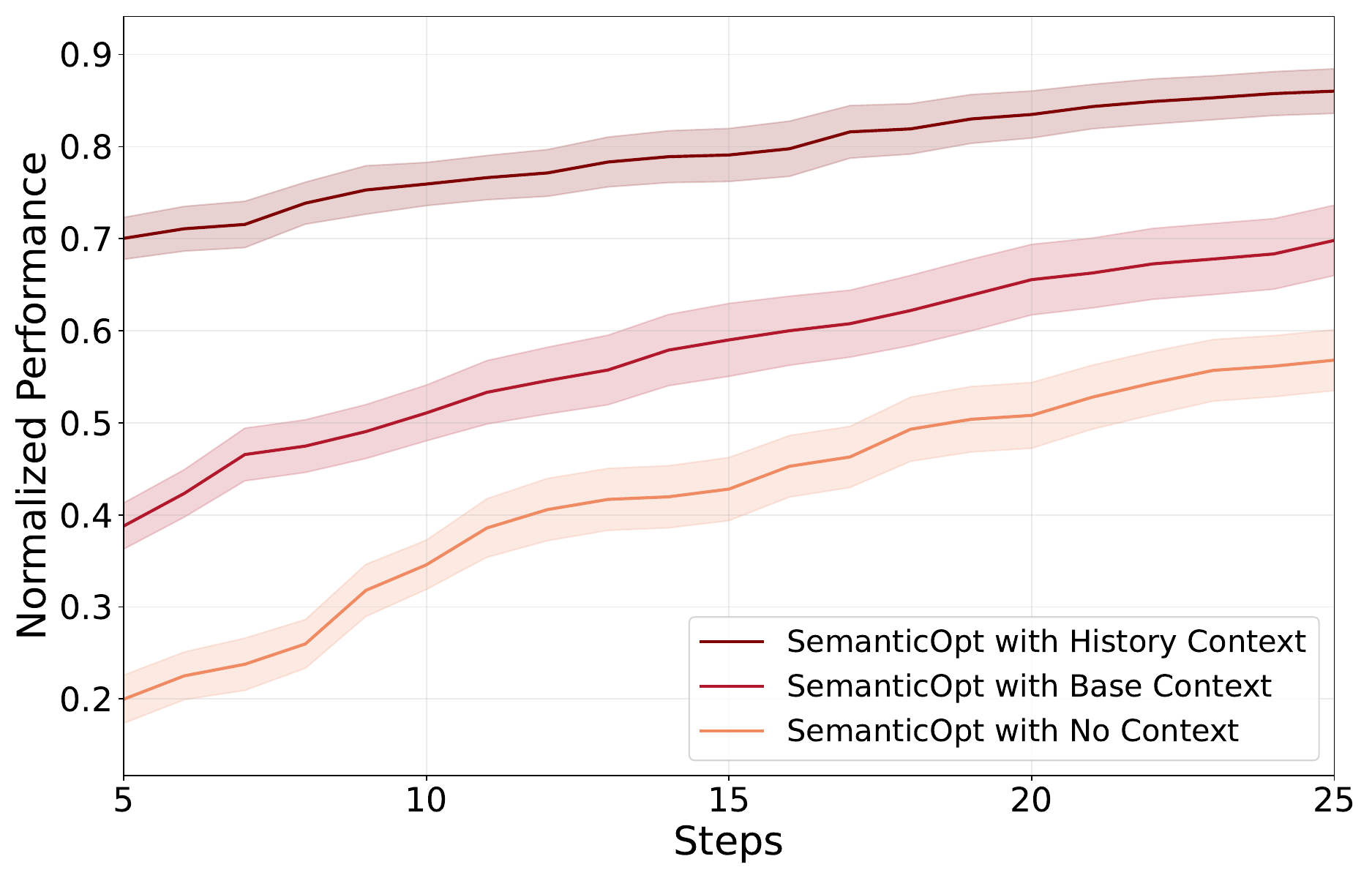}
        \caption{Cookie Recipe History}
        \label{fig:cookie_sem}
    \end{subfigure}
    \begin{subfigure}[b]{0.32\textwidth}
        \centering
        \includegraphics[width=\linewidth]{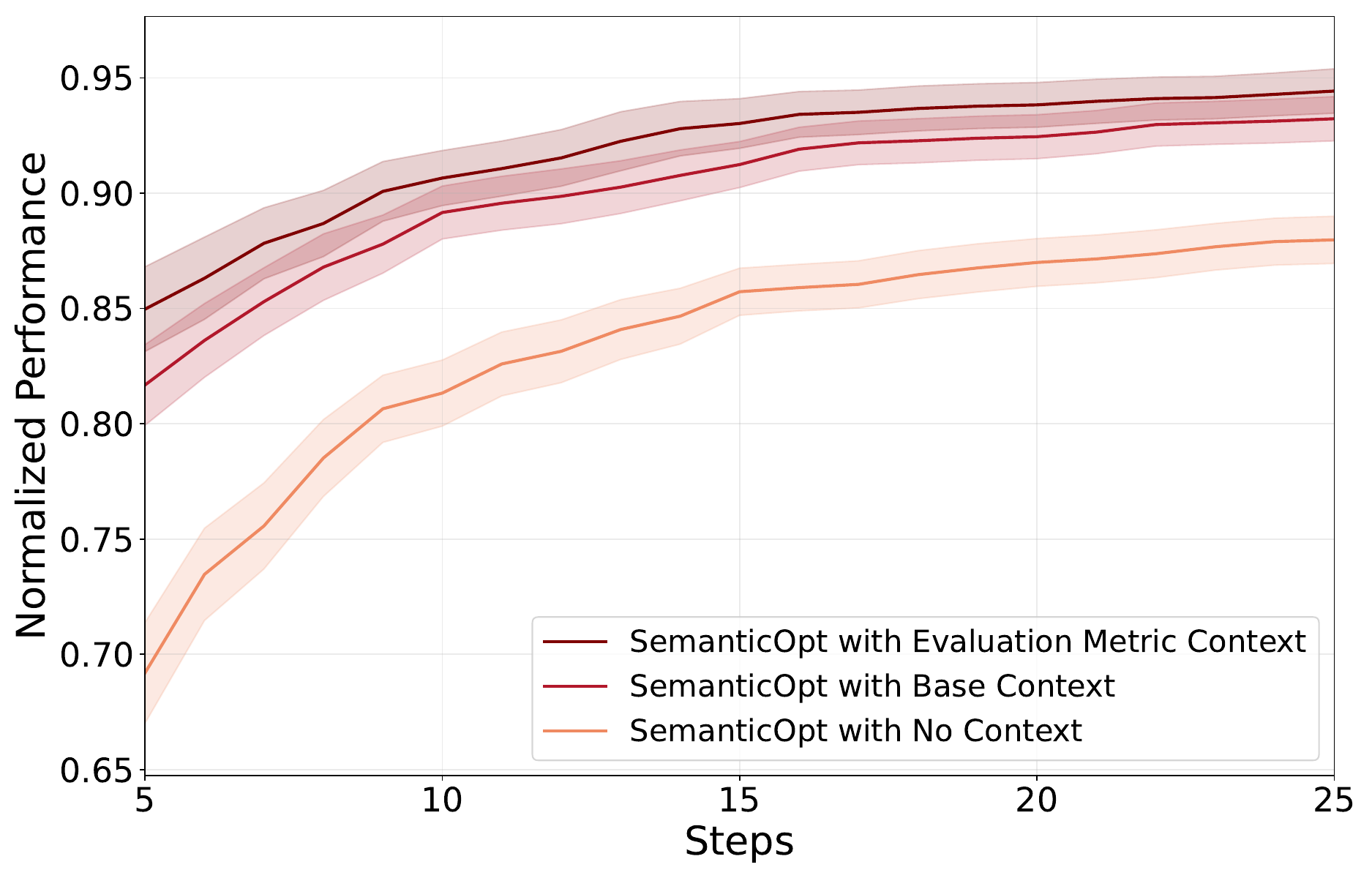}
        \caption{PD1 Evaluation Metrics}
        \label{fig:pd1_sem}
    \end{subfigure}

    \caption{Mean normalized performance with paired SEM for SemanticOpt under different types and levels of semantic data. Adding additional relevant semantic information improves performance.}
    \label{fig:semantic}
\end{figure*}

To understand the effect of the type of semantic data provided to the model, we test SemanticOpt by providing different categories of semantic data. We evaluate this for three different types of information contained in our dataset: long-form context, historical examples, and extra evaluation metrics. We test the Iron Mind benchmark with a long-form summarization of a relevant paper to each chemical reaction optimization problem in Figure~\ref{fig:ironmind_sem}, the cookie recipe benchmark with example cookie recipes gathered online as historical context in Figure~\ref{fig:cookie_sem}, and the PD1 HPO benchmark with additional evaluation time context (train/validation curve information) in Figure~\ref{fig:pd1_sem}. We see that the additional information improves the model performance for each benchmark.  

\paragraph{Model Size:}

\begin{figure*}[h]
    \centering
    
    \begin{subfigure}[b]{0.45\textwidth}
        \centering
        \includegraphics[width=\linewidth]{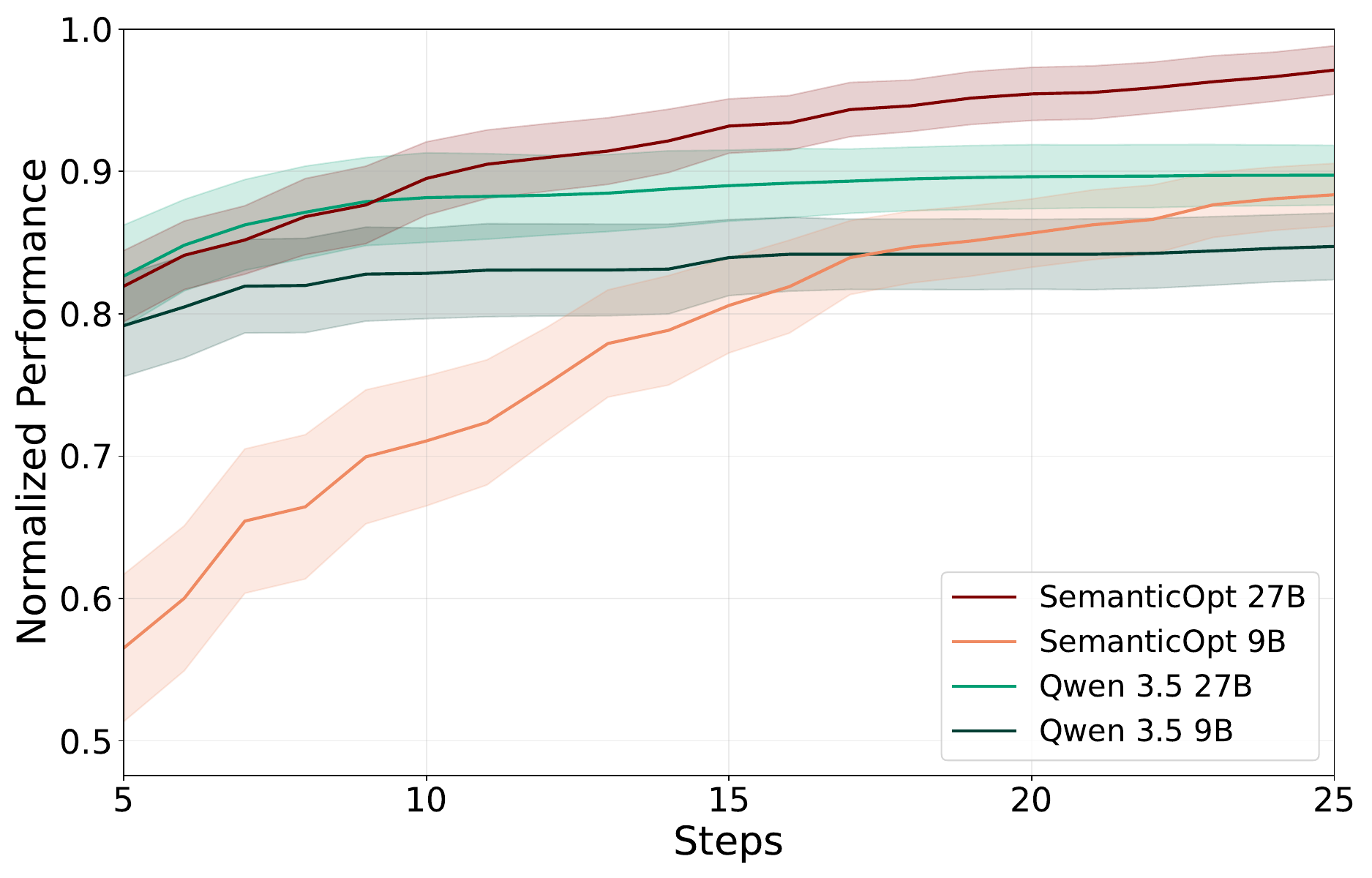}
        \caption{Model Size Ablation}
        \label{fig:model_size}
    \end{subfigure}
    \hfill
    \begin{subfigure}[b]{0.45\textwidth}
        \centering
        \includegraphics[width=\linewidth]{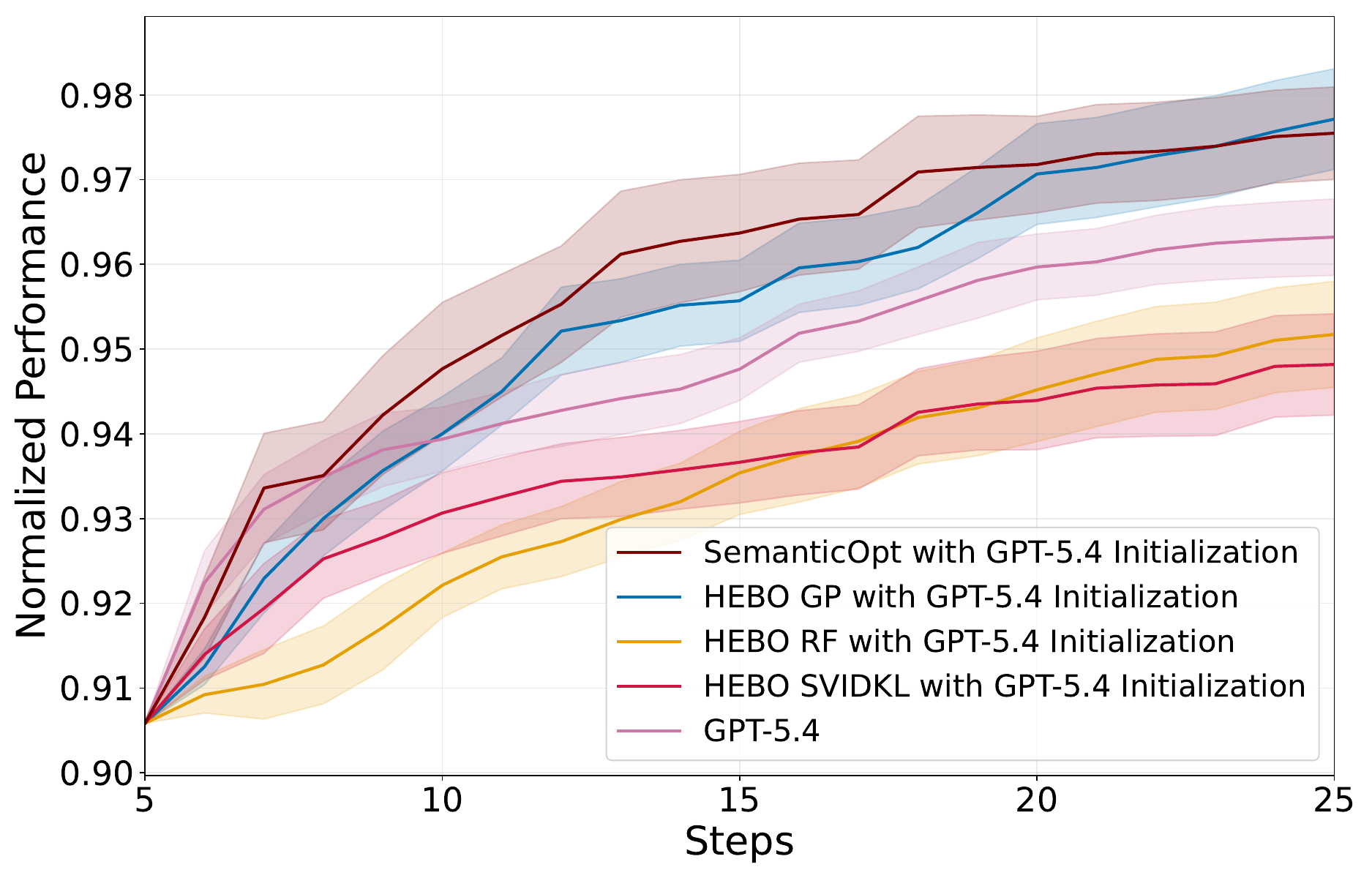}
        \caption{Initialization Ablation}
        \label{fig:initialization_ablation}
    \end{subfigure}

    \caption{Ablation results on model size and initialization benchmark. We see that larger model size is necessary to maintain initialization performance after fine-tuning and that SemanticOpt matches or beats state-of-the-art BO performance when both are given initializations from GPT-5.4.}
    \label{fig:ablation}
\end{figure*}

We evaluate the role of model capacity by fine-tuning Qwen 3.5 models at both 9B and 27B parameter scales. As shown in Figure~\ref{fig:model_size}, the larger model achieves stronger performance and preserves the useful initialization behavior of the base model after fine-tuning on the paint mixing benchmark. However, we see a larger impact of fine-tuning on the initialization capabilities for the 9B model, resulting in significantly lower initialization performance for the fine-tuned smaller model. This suggests capacity-dependent forgetting. As fine-tuning shifts the model, this can overwrite the pretrained priors of the model. The 9B model may have less capacity to retain these priors.

\paragraph{Initialization:}

We further examine whether SemanticOpt's gains arise from improved warm-starting (the choice of initial samples) or from better iterative optimization. To do so, we test BO optimizers and SemanticOpt on GPT-5.4 initializations on the PD1 benchmark. Given the same initializations, we see that SemanticOpt's iterative optimization matches HEBO GP, while outperforming HEBO RF and HEBO SVIDKL, showing that the performance advantage is due to both high-quality initialization and iterative performance, which base LLMs cannot match (see Fig.~\ref{fig:initialization_ablation}).

\section{Limitations and Future Work}

SemanticOpt shows strong performance over a wide range of domains and outperforms both state-of-the-art BO methods and LLMs on average when given relevant context. However, it performs relatively worse on benchmarks with only categorical parameters, such as Nasbench201 and Iron Mind. We hypothesize that this is due to limited representation of purely discrete spaces in the training data and the weaker performance of BO methods in these domains. Additionally, since SemanticOpt is trained on BO trajectories, it may be less effective in domains where BO is poorly calibrated, such as when the objective is highly discontinuous. Finally, SemanticOpt depends on the quality of the provided semantic context and may be sensitive to misleading context, as well as domains where LLMs have little base knowledge. 

Taking further advantage of the capabilities of LLMs by incorporating justifications for decisions is an area of future research. This could enable better user understanding of experimental design processes. Additionally, the flexibility of LLMs could allow for more semantic context to be provided or more flexible function spaces, such as changing parameters in the middle of optimization trajectories or including parameters that are difficult to quantify in traditional optimizers. Lastly, integrating this training approach into more capable models could result in further performance improvements.

\section{Conclusion}

We introduce SemanticOpt, an LLM-based framework for black-box optimization that fine-tunes language models on BO trajectories augmented with surrogate predictions and semantic context. Across diverse benchmarks, SemanticOpt achieves the best average performance among the evaluated BO and LLM-based baselines when relevant semantic information is provided. In particular, SemanticOpt shows improved iterative performance compared to the base Qwen 3.5 27B model. These results demonstrate that LLMs can be trained as sequential optimizers that combine numerical feedback with semantic problem information, enabling end-to-end optimization systems that benefit from using documentation, prior experiments, and auxiliary measurements when proposing new experiments.

\clearpage
\bibliographystyle{unsrtnat}
\bibliography{SemanticOpt}

\newpage
\appendix

\section{Dataset Generation}
\label{app:dataset}

\paragraph{Function Space Generation:}

Because widely available real-world benchmark functions are limited, we construct a large-scale training dataset using realistic and diverse synthetic objectives. We generate two classes of evaluation functions for this purpose: Gaussian process (GP)-based functions and LLM-generated function spaces. Together, these sources provide both scalable synthetic diversity and semantically grounded problem structures. GPs serve as flexible function generators that represent a wide variety of complex functions, enabling large-scale data collection while producing rich optimization landscapes. The motivation is that GP-based Bayesian optimization methods are most widely used as they perform well across many function classes \cite{shahriari2015taking, wang2023recent}. Therefore, by training on GP-generated landscapes, our objective is to learn a policy that can be similarly generalized in diverse problem spaces. To further diversify these tasks, we vary both input and output scales to reflect realistic regimes. For example, objectives may lie in ranges such as $[0,100]$ or $[1,10]$, while parameters may span $[10^{-4}, 10^{-1}]$. We augment sampled GP functions with nonlinear coordinate warping, discontinuities, and constraints, transforming smooth samples into landscapes that better resemble real-world systems. We randomly generate a large number of random scales to allow for generalization. Our GP-based functions are sampled from a Gaussian process prior defined over the input space. Random covariance kernels are generated by combining base kernels (RBF, Matern, Rational Quadratic, Exponential) using addition and multiplication, with up to three kernels per function. Hyperparameters such as lengthscales and variances are drawn from log-uniform distributions. A fixed set of initial points is sampled, and function values are generated from the corresponding multivariate Gaussian. Evaluations are obtained by conditioning on this prior, yielding smooth but diverse optimization landscapes.

We also add further augmentations to the randomly generated GP processes. These augmentations transform smooth GP samples into challenging, realistic landscapes by applying non-linear coordinate warping, introducing discontinuities, and adding complex parameter couplings. The goal of this process is to create a training set that mimics the irregularity of real-world systems (e.g., phase transitions, saturation limits, or discrete configuration spaces). By creating functions with these diverse methods, we provide a base for the model to learn realistic function dynamics across a variety of possible function spaces. Lastly, we augment examples with failure regions, where the function space returns failure cases. These are selected as the worst performing points or randomly selected edge regions.

We also use LLM-generated function spaces to introduce real-world domains into the dataset. To do this, we prompt Qwen 3.5 35B-A3B to produce semantic context of real-world black-box optimization problems, along with 100 evaluated points in the function space. We choose this model for its combination of speed and capabilities. By prompting the model to produce example problems with various types of semantic data included, we aim to create training data that matches the diversity of semantic information available in real-world problems. 

Depending on the context profile, the generated artifact may also include operational background, a full-paper-style domain brief, objective ranges, historical heuristics, analogous successful parameterizations, and additional diagnostic evaluation metrics. These profiles allow us to construct controlled variants that expose different amounts and types of semantic information, matching the heterogeneous context available in real-world optimization workflows. The sampled points are validated against the declared search space and then used to fit a tabular surrogate objective; the benchmark code selects among candidate surrogate families including XGBoost, CatBoost, Gaussian processes, random forests, and extra trees. The fitted surrogate defines the ground-truth function surface used during optimization, while the generated JSON supplies the semantic context given to context-aware samplers. By combining GP-generated and LLM-generated objectives, we create a broader and more representative training distribution for model fine-tuning.

\paragraph{Trajectory Generation:}

Using our generated function spaces, we generate sequences of proposed configurations and evaluated points, which we call trajectories. We use HEBO~\cite{cowen2022hebo}, a state-of-the-art BO framework, to generate training trajectories because we need a high-quality optimizer that includes reliable surrogate modeling capabilities. HEBO is a strong choice because it was designed for practical black-box optimization settings where search spaces are often continuous, discrete, and categorical. It has also demonstrated strong empirical performance, including first place in the NeurIPS 2020 Black-Box Optimization Challenge~\cite{turner2021bayesian}. 

\begin{figure}[h]
    \centering
    \includegraphics[width=0.6\linewidth]{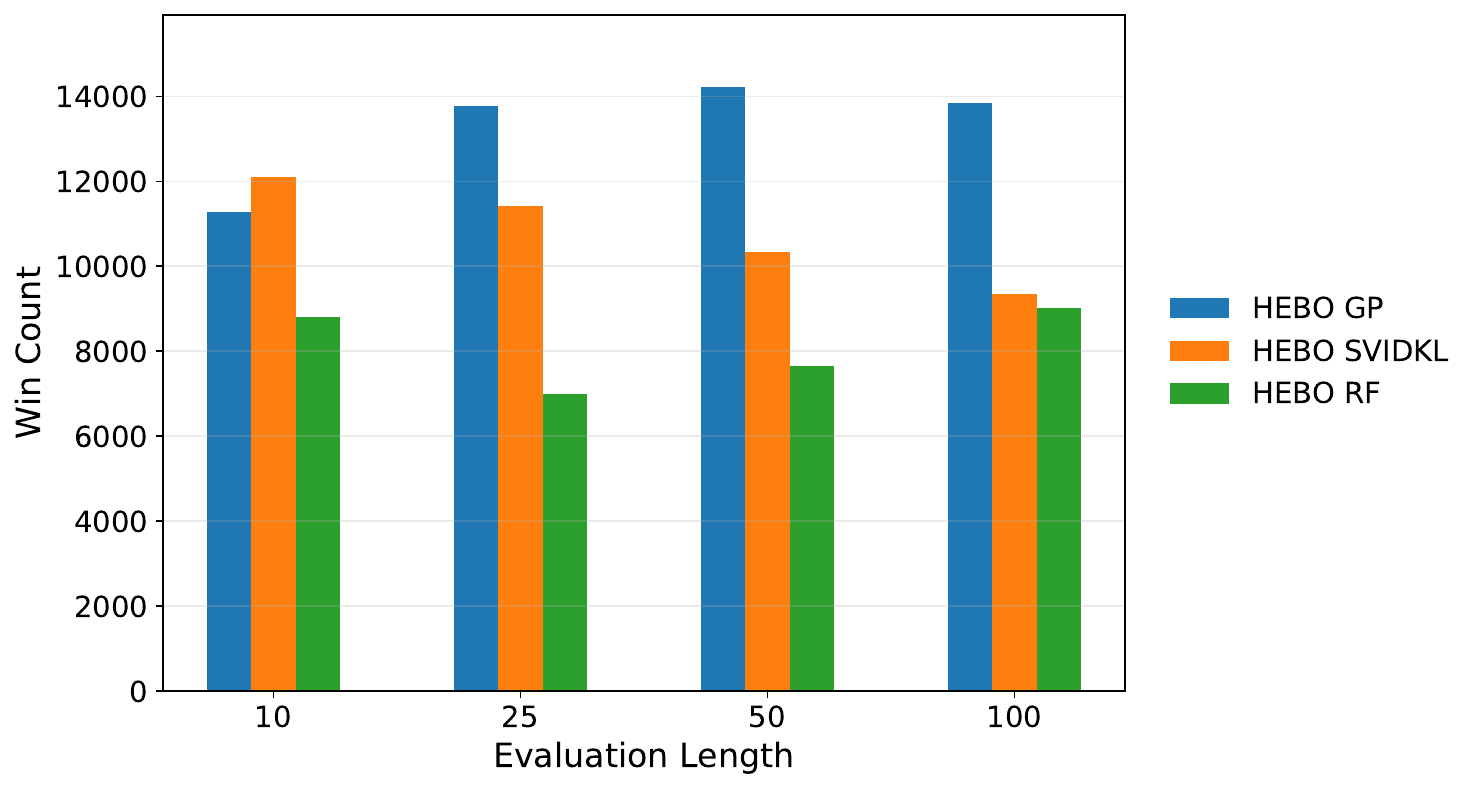}
    \caption{Win count of different HEBO surrogate models. This measures the number of trajectories generated by each HEBO method which are selected as the best trajectory on a function space for each evaluation length. Only the best-performing trajectory for each function space and evaluation length is included in the training data.}
    \label{fig:win_count}
\end{figure}

We generate our training dataset by running HEBO on our generated function spaces with different surrogate models. We begin by initializing our process with 5 initial steps. We use Sobol sampling ~\cite{sobol1967distribution} for the GP-based functions and use LLM-generated initializations for the functions with semantic information. We then proceed with using 3 different surrogate models implemented in HEBO, including Gaussian process (GP), random forest (RF), and Stochastic Variational Inference Deep Kernel Learning (SVIDKL). To compile the dataset, we select the best performing surrogate model after 10, 25, 50, and 100 total steps. The performance of each surrogate model on our training dataset is shown in Figure~\ref{fig:win_count}. These results show that different surrogate models perform better in different function spaces. Additionally, we see that the performance is relatively balanced and the model is learning from a combination of all three surrogate models. Therefore, the model is trained to act like the best BO surrogate in each scenario, with the goal of performing better than any individual method.

We show complete examples of a generated trajectory without context in Figure~\ref{fig:opt_prompt_full_app} and a full initialization with context in Figure~\ref{fig:opt_prompt_full_app2}. The prompt begins with a system instruction defining the agent's role as a black-box optimization assistant and enforcing a strict JSON output format containing the action and predicted mean and standard deviation. We include a failure probability field after the optimizer has observed at least one failed evaluation. The user then provides the problem description and relevant context. We provide all numbers rounded to 4 significant figures and convert the BO steps into a multi-turn user and assistant conversation. The interaction follows in a turn-based manner where the assistant proposes a new configuration, which is the action proposed by BO and the surrogate model's predicted mean and standard deviation, and the user returns the evaluated objective value, repeating until the evaluation budget is exhausted.

\begin{figure*}[h]
\centering
\begin{lstlisting}[
    style=dialogue
]
System:
You are a black-box optimization assistant. Your goal is to iteratively propose configurations to optimize the objective within an evaluation budget.
At each step, return exactly one valid JSON object as your output.
 - Before any evaluations are observed, if the user asks for initialization, the JSON must contain exactly these keys: initial_configuration_1, ..., initial_configuration_5.
 - Before any observed failure, the JSON must contain exactly these keys: next_configuration, mean, std.
 - After at least one observed failure, the JSON must contain exactly these keys: next_configuration, mean, std, failure_probability.
 - next_configuration must include every key in the tunable search space with valid in-range values.
 - mean and std must predict the objective value distribution for the proposed configuration.
Each new configuration should seek to gain new insight about the problem space and achieve the best objective possible within the evaluation budget.
User:
Problem description:
We are optimizing an unknown black-box problem.
Tunable search space:
- param_0: float (log search space) in range [0.002758, 0.1874] | description: unknown float (log search space) parameter | unit: unknown units
- param_1: float in range [0.4848, 3.843] | description: unknown float parameter | unit: unknown units
- param_2: integer in range [3, 24] | description: unknown integer parameter | unit: unknown units
- param_3: integer in range [40, 291] | description: unknown integer parameter | unit: unknown units
Objective:
- Direction: minimize
- Unit: unknown units
- Description: unknown black-box scalar objective
Budget:
- Total evaluations available (including 5 initial evaluations): 100

Initial evaluations:
Evaluation 1:Configuration: {"param_0": 0.168, "param_1": 2.512, "param_2": 19, "param_3": 227}
Objective: -0.3808 unknown units
Evaluation 2: Configuration: {"param_0": 0.1219, "param_1": 2.662, "param_2": 18, "param_3": 204}
Objective: -0.3239 unknown units
Evaluation 3: Configuration: {"param_0": 0.1798, "param_1": 2.53, "param_2": 5, "param_3": 122}
Objective: failure
Evaluation 4: Configuration: {"param_0": 0.1851, "param_1": 1.963, "param_2": 3, "param_3": 62}
Objective: -0.1428 unknown units
Evaluation 5: Configuration: {"param_0": 0.1767, "param_1": 1.251, "param_2": 3, "param_3": 189}
Objective: -0.2731 unknown units

Output the next configuration, including failure_probability.
Assistant:
{"next_configuration": {"param_0": 0.09889, "param_1": 2.057, "param_2": 20, "param_3": 225}, "mean": -0.1429, "std": 0.2828, "failure_probability": 0.178}
User:
Evaluation 6: Objective: -0.6286 unknown units
Output the next configuration, including failure_probability.
Assistant:
{"next_configuration": {"param_0": 0.1022, "param_1": 1.872, "param_2": 23, "param_3": 260}, "mean": -0.5211, "std": 0.2647, "failure_probability": 0.1403}
User:
Evaluation 7: Objective: -0.5112 unknown units
Output the next configuration, including failure_probability.
\end{lstlisting}

\caption{A synthetic example problem from our training data. This example contains failure cases, such that the model outputs failure probability for proposed points.}
\label{fig:opt_prompt_full_app}

\end{figure*}

\begin{figure*}[h]
\centering
\begin{lstlisting}[
    style=dialogue
]
User:
Problem description:
You are optimizing a black-box benchmark: Online Grocery Delivery Window Optimization.
This task involves tuning operational parameters to minimize customer wait times while maximizing driver utilization for same-day grocery deliveries. The system must balance aggressive delivery promises against the risk of missing windows due to traffic and order volume fluctuations. Key levers include time-slot granularity, buffer allocations, and dynamic pricing thresholds that influence demand distribution. The goal is to find a configuration that reduces average lateness without causing excessive driver idle time.
In the online grocery sector, delivery window commitments are a critical competitive differentiator. Customers expect precise time slots, but operational reality is driven by unpredictable variables like traffic, weather, and order clustering. Operators must configure their dispatch algorithms to accept orders within specific constraints that ensure on-time delivery while maintaining cost efficiency. This optimization problem focuses on the interplay between slot definition, buffer management, pricing strategies, and route capacity limits. Poorly tuned parameters can lead to a cascade of missed deliveries, customer churn, and increased operational costs due to inefficient routing or overtime pay.
Tunable search space:
- max_orders_per_route: integer in range [10, 50] | description: Maximum number of orders assigned to a single driver route. | unit: orders
- enable_dynamic_pricing: boolean in set {true, false} | description: Whether to activate surge pricing for peak demand windows. | unit: unitless
- route_reoptimization_interval: category in set {hourly, every_30min, none} | description: How often the routing plan is recalculated during the day. | unit: frequency
- pricing_elasticity: float (log search space) in range [0.2, 1.5] | description: Sensitivity of demand to price changes; higher values shift demand to off-peak times. | unit: ratio
- cancellation_threshold_minutes: integer in range [15, 120] | description: Time before the window start when an order is automatically cancelled if not confirmed. | unit: minutes
- buffer_minutes: integer in range [0, 30] | description: Extra time added to the promised window to absorb minor delays. | unit: minutes
- slot_duration_minutes: integer in range [30, 120] | description: The duration of each available delivery time slot. | unit: minutes
- priority_weight_factor: float in range [0.5, 3] | description: Weight applied to VIP orders in the routing algorithm. | unit: multiplier
Objective:
- Direction: minimize
- Name: weighted_service_cost
- Unit: cost_units
- Description: A composite metric combining the total minutes of lateness weighted by order value and the cost of driver idle time. Lower values indicate better service reliability and operational efficiency.
Budget:
- Total evaluations available (including 5 initial evaluations): 100

Provide 5 initial configurations that explore promising regions in the search space.
\end{lstlisting}

\caption{An example initialization with contextual information from our training data.}
\label{fig:opt_prompt_full_app2}
\end{figure*}

Table~\ref{tab:included-trajectories} summarizes the trajectory blocks used to construct the training data. For the real-world surrogate data, we include examples of function spaces with and without regions with failure cases, giving the factor of two in the real-world total. These failure regions are returned as unknown constraint failures to allow the model to learn function spaces where failures may occur. Each configuration is sampled across nine input dimensions, $d \in \{2,\ldots,10\}$, with the per-dimension counts shown in the table. The real-world blocks therefore contribute $2 \cdot 9 \cdot 800 = 14{,}400$ function spaces. The synthetic data uses four dataset families, with $500$ functions per dimension, contributing $4 \cdot 9 \cdot 500 = 18{,}000$ function spaces. In total, this yields $32{,}400$ function spaces. Since we select the best trajectory in the function space from four trajectory lengths, $T \in \{10,25,50,100\}$, the final dataset contains $32{,}400 \cdot 4 = 129{,}600$ training sequences.

\begin{table}[h]
\centering
\small
\caption{Function space variants used for training-data generation. All functions are generated for dimensions $d \in \{2,\ldots,10\}$ and trajectories are selected for trajectory lengths $T \in \{10,25,50,100\}$.}
\label{tab:included-trajectories}
\begin{tabular}{@{}llr@{}}
\toprule
\textbf{Dataset Family} & \textbf{Function Space Variants} & \textbf{Functions / Dimension} \\
\midrule
\multirow{8}{*}{Real-world surrogate}
& Base context & 200 \\
& History & 200 \\
& Long-form context & 100 \\
& Extra metrics & 50 \\
& History + long-form & 100 \\
& History + extra metrics & 50 \\
& Long-form + extra metrics & 50 \\
& All context & 50 \\
\addlinespace[0.3em]
\midrule
\multirow{4}{*}{Synthetic}
& Base & 500 \\
& Augmented & 500 \\
& Contains Failure Regions & 500 \\
& Augmented + Contains Failure Regions & 500 \\
\midrule
\multicolumn{2}{@{}r}{Real-world Function Spaces}
& $2 \cdot 9 \cdot 800 = 14{,}400$ \\
\multicolumn{2}{@{}r}{Synthetic Function Spaces}
& $4 \cdot 9 \cdot 500 = 18{,}000$ \\
\multicolumn{2}{@{}r}{\textbf{Total Function Spaces}}
& $\mathbf{32{,}400}$ \\
\multicolumn{2}{@{}r}{\textbf{Total Training Trajectories}}
& $\mathbf{129{,}600}$ \\
\bottomrule
\end{tabular}
\end{table}

\section{Model Training and Usage}

\subsection{Training}
\label{app:train}

We use the Qwen 3.5 family of models as our base model. Specifically, we fine-tune the Qwen 3.5 27B model, which provides strong instruction-following and numerical reasoning capabilities while remaining feasible to adapt with parameter-efficient fine-tuning. We fine-tune the model using low-rank adaptation (LoRA)~\cite{hu2022lora} with Unsloth~\cite{unsloth}. This provides an efficient framework for adapting large language models within our computational budget. We use our generated dataset of conversational optimization trajectories, where each example contains the problem context, previous evaluations, surrogate-model predictions, and the next configuration selected by the optimizer to train the model. We use the resulting model for all SemanticOpt experiments.

We use the hyperparameters shown in Table~\ref{tab:hyper} to fine-tune the model. The trajectories are formatted as multi-turn text conversations, as shown in Figure~\ref{fig:opt_prompt_full_app}, and the model is trained with supervised next-token prediction on the assistant responses only. Training is performed with BF16 precision using LoRA adapters, which enables fine-tuning of the 27B-parameter base model within our available GPU memory. The same hyperparameters were used for training the 9B parameter model. Further hyperparameter tuning and longer training runs were limited by computational budget. We found that higher learning rates improved training loss but resulted in worse initialization performance. Therefore, we achieved a good balance by using a lower learning rate.

\begin{table}[h]
  \centering
  \caption{Hyperparameters for model training.}
  \label{tab:hyper}
  \begin{tabular}{ll}
    \textbf{Hyperparameter} & \textbf{Value} \\
    \hline
    Base Model & Qwen 3.5 27B \\
    Fine-tuning Method & LoRA \\
    Learning Rate & $1 \times 10^{-5}$ with cosine scaling \\
    Batch Size & 64 trajectories \\
    LoRA Rank & 16 \\
    LoRA Alpha & 16 \\
    Precision & BF16 \\
    Epochs & 1 \\
    Maximum Sequence Length & 24,576 tokens \\
    \hline
  \end{tabular}
\end{table}

\subsection{Inference}
\label{app:inference}

At inference time, SemanticOpt uses the same multi-turn structure as in training. For problems with semantic context, the model first receives the problem description and relevant contextual information, then proposes 5 initial configurations. After these initial evaluations are returned, SemanticOpt iteratively proposes new configurations until the evaluation budget is reached.

For each optimization step, we sample two candidate configurations from the fine-tuned model. Each candidate includes a proposed configuration, the model's predicted objective mean and standard deviation, and a predicted probability that the evaluation will succeed. We then compute expected improvement using the self-predicted objective distribution and weight it by the predicted success probability:
\[
\mathbf{x}_{t+1} = \mathbf{x}^{(j^*)}, \qquad
j^* = \arg\max_j p_{\mathrm{succ}}^{(j)}
\, \mathrm{EI}(\mu^{(j)}, \sigma^{(j)}, y_t^*).
\]
Equivalently, this downweights candidates that may have high predicted objective improvement but are likely to fail during evaluation. This provides a lightweight exploration-exploitation mechanism that also accounts for feasibility, while keeping inference simple and fully contained within the LLM-based optimizer.

\begin{figure*}[h]
    \centering
    
    \begin{subfigure}[b]{0.32\textwidth}
        \centering
        \includegraphics[width=\linewidth]{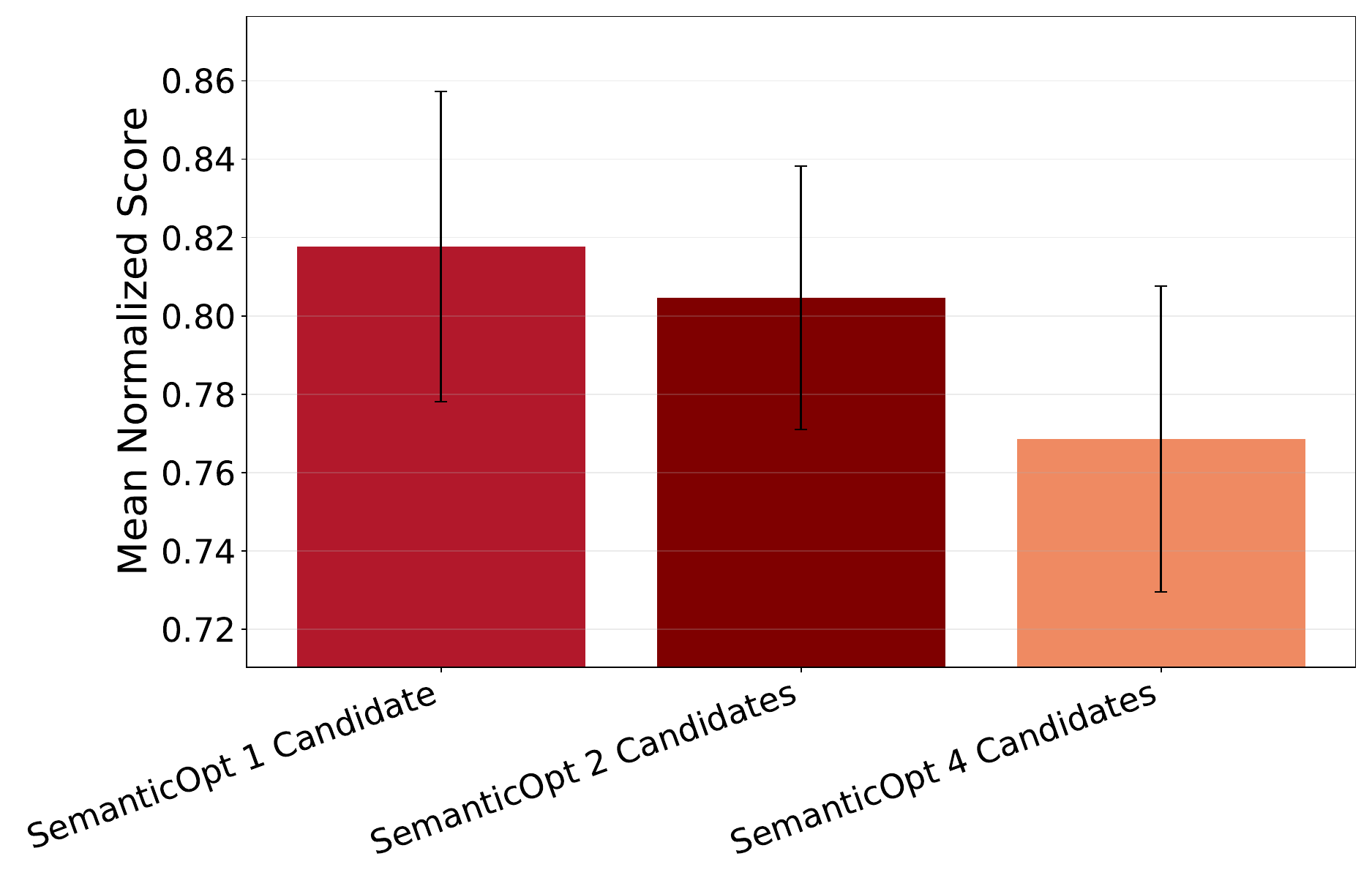}
        \caption{Airfoil Results}
        \label{fig:airfoil_cand}
    \end{subfigure}
    \begin{subfigure}[b]{0.32\textwidth}
        \centering
        \includegraphics[width=\linewidth]{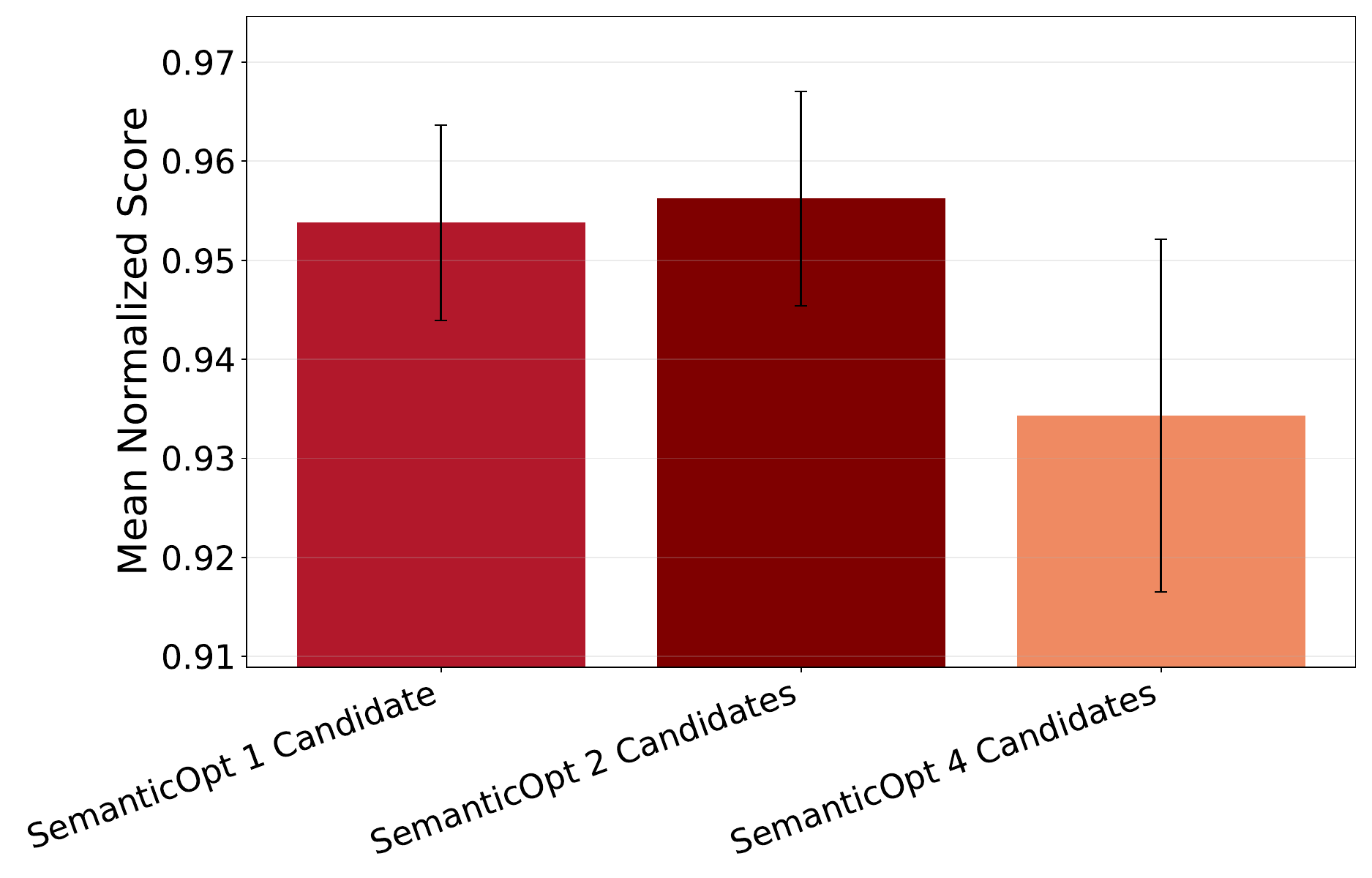}
        \caption{PD1 Results}
        \label{fig:pd1_cand}
    \end{subfigure}
    \begin{subfigure}[b]{0.32\textwidth}
        \centering
        \includegraphics[width=\linewidth]{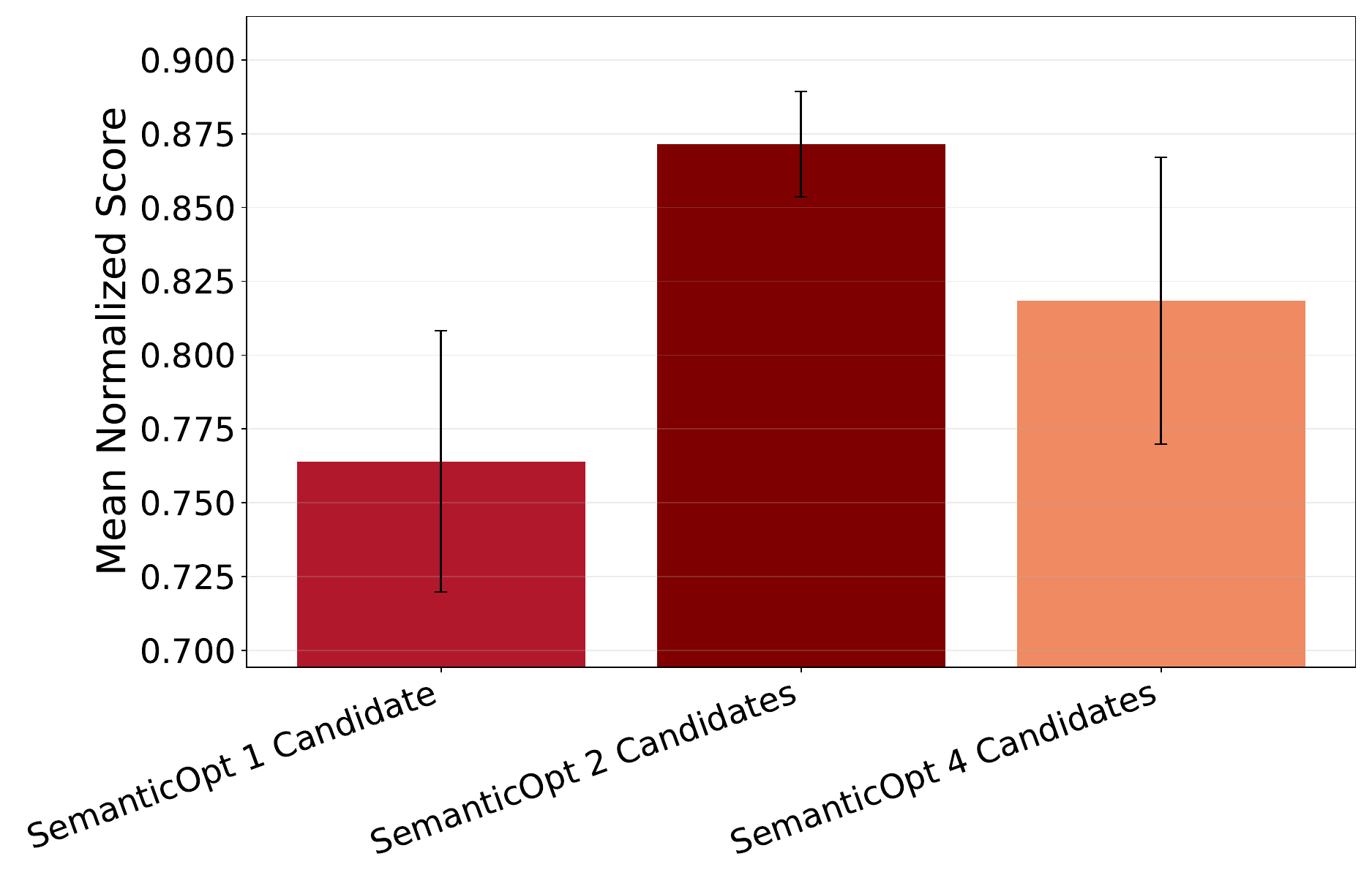}
        \caption{NeqSim Results}
        \label{fig:neqsim_cand}
    \end{subfigure}

    \caption{Mean normalized performance with paired SEM for SemanticOpt with different candidate counts. We see that using two candidates provides a good middle ground solution.}
    \label{fig:cand}
\end{figure*}

We select two candidates based on the ablation in Figure~\ref{fig:cand}. We test proposing one, two, and four candidates at each iterative step with SemanticOpt. We find that using one or two candidates gives similar results on Airfoil and PD1, while using four candidates leads to a drop in performance. On NeqSim, however, using two candidates provides a substantial improvement, likely because NeqSim contains evaluation failures. In this setting, sampling multiple candidates allows SemanticOpt to trade off predicted improvement against predicted success probability, avoiding configurations that appear promising but are likely to fail.

Testing multiple proposed points on the base Qwen 3.5 27B results in adverse results. We find that conditioning on selected points results in the model consistently selecting duplicate or out-of-bounds values that make the model difficult to evaluate. Therefore, we use one candidate selection for the base model.

Inference is practical on a single GPU. On an NVIDIA L40S, SemanticOpt takes roughly one minute to generate a complete 25-step optimization trajectory with our fast surrogate models for evaluation.

\section{Baselines and Benchmarks}
\label{app:benchmarks}

\subsection{Baselines}
\label{app:baselines}

\paragraph{Bayesian Optimization Baselines:}
\begin{description}
    \item[HEBO-GP~\cite{cowen2022hebo}:]
    A Bayesian optimization baseline from the HEBO framework using a Gaussian process surrogate model. The GP surrogate provides a probabilistic model of the objective and selects new configurations using an acquisition function that balances exploitation of high-performing regions with exploration of uncertain regions. This is a strong classical BO baseline for continuous and mixed search spaces.

    \item[HEBO-RF~\cite{cowen2022hebo}:]
    A HEBO variant using a random forest surrogate model. Random forest surrogates can be more robust on irregular, discontinuous, or mixed categorical-continuous search spaces where a smooth GP assumption may be less appropriate. This baseline helps test whether SemanticOpt improves over a tree-based BO method that can handle non-smooth response surfaces.

    \item[HEBO-SVIDKL~\cite{cowen2022hebo}:]
    A HEBO variant using stochastic variational inference deep kernel learning as the surrogate model. This combines neural feature learning with kernel-based uncertainty estimation, making it a more flexible surrogate than a standard GP while retaining an uncertainty-aware BO acquisition strategy. We include this baseline to compare SemanticOpt against a stronger model-based optimizer that can capture more complex objective structure.

    \item[HEBO Baseline Setup:]
    For all HEBO variants, we use the OptunaHub implementations with their default settings. These methods receive the same numerical search space and evaluated objective values as SemanticOpt, but they do not use natural-language semantic context such as parameter descriptions, papers, documentation, historical examples, or auxiliary textual information. The comparison therefore separates the benefit of semantic information and learned LLM-based optimization from strong classical BO performance.
\end{description}

\paragraph{Transformer-Based and Learned Optimization:}
\begin{description}
    \item[LLAMBO (Large Language Model Bayesian Optimization)~\cite{liu2024large}:]
    An LLM-assisted Bayesian optimization method that uses a pre-trained language model to help propose and model candidate configurations. The core method involves prompting an LLM many times to produce a surrogate model and using an acquisition function to optimize this model. We evaluate the OptunaHub implementation with an adaptation to support categorical values natively within the model, which uses GPT-4o mini~\cite{hurst2024gpt} by default. Using a higher-quality model or a reasoning model would be prohibitively expensive due the nature of the many API calls required by the method. However, we did see that LLAMBO frequently failed to generate suggestions for categorical parameters, including suggesting categories that were not presented as options. The methodology does not provide a clear structure for data input, which could result in improved performance. 

    \item[PFNs4BO (Prior-Data Fitted Networks for BO)~\cite{muller2023pfns4bo}:]
    A transformer-based Bayesian optimization method based on prior-data fitted networks. PFNs4BO is pre-trained on synthetic function data and performs surrogate-style BO inference with a single forward pass, avoiding repeated Gaussian process fitting. We use the pre-trained HEBO+ model implemented in Optunahub.
\end{description}

\paragraph{LLM Baselines:}
\begin{description}
    \item[Qwen 3.5 27B~\cite{team2026qwen3} \& OpenAI GPT-5.4~\cite{singh2025openai}:]
    We evaluate general-purpose LLMs as zero-shot optimization agents using the same prompt structure and semantic information as SemanticOpt. Qwen 3.5 27B is the base model used for SemanticOpt, so this comparison isolates the effect of fine-tuning on structured optimization trajectories. We test the Qwen model in no-reasoning mode, in the same fashion as our fine-tuned model. Both models are quantized to allow for easily inference capabilities. However, tests with reasoning enabled did not improve results. GPT-5.4 provides a frontier-model baseline for testing whether stronger general-purpose reasoning alone is sufficient for semantic black-box optimization. We use medium reasoning level for GPT-5.4.

    These baselines can often provide strong initial configurations, especially when the benchmark resembles common public optimization tasks. However, they typically struggle to improve consistently over longer trajectories. This comparison is intended to show that semantic context and prompting alone are not sufficient for reliable iterative optimization, and that fine-tuning on BO-style trajectories is needed to obtain stronger long-horizon behavior.
\end{description}

\section{Benchmarks}
\label{app:benchmark_descriptions}

\paragraph{Airfoil CFD~\cite{jasak2009openfoam}:}
The Airfoil CFD benchmark evaluates black-box optimization in aerodynamic shape design. Each trial defines a reduced PARSEC-style airfoil parameterization, including geometric quantities such as leading-edge radius, thickness, camber, crest location, trailing-edge properties, and angle of attack. Candidate designs are evaluated a surrogate trained on OpenFOAM CFD simulation samples. The objective is to minimize drag while satisfying a seed-dependent target lift coefficient, with penalties for lift mismatch, invalid geometries, and nonphysical aerodynamic behavior. This benchmark tests whether an optimizer can use semantic information about airfoil geometry and aerodynamic objectives to navigate a mixed continuous, integer, and categorical design space.

\paragraph{Convex Decomposition:}
The Convex Decomposition benchmark models the tuning of heuristic parameters for decomposing concave polygons into convex components. The optimizer does not choose the shapes themselves; instead, it selects weights and preferences used by a deterministic splitting policy, such as penalties for long diagonals, unbalanced splits, low-compactness fragments, remaining reflex vertices, and tiny sliver components. Each configuration is evaluated on a fixed corpus of procedurally generated polygons, and the default objective aggregates part count, cut length, compactness loss, imbalance, and runtime-related penalties. This provides a lightweight but structured geometry-processing benchmark where semantic descriptions of the heuristic parameters are directly relevant to performance.

\paragraph{Cookie Recipe~\cite{solnik2017bayesian}:}
The Cookie Recipe benchmark represents a food formulation problem based on chocolate-chip cookie optimization. The search space contains ten recipe and preparation variables, including ingredient amounts, sugar balance, butter and egg quantities, leavening and salt, vanilla, chocolate-chip types, and baking temperature. Candidate recipes are scored against hidden high-quality recipe anchors using a weighted similarity function that accounts for numeric deviations, categorical chip-type mismatch, and recipe-balance terms such as sugar-to-butter ratio. The objective is to maximize an estimated cookie-quality score, making this benchmark a semantically rich mixed-space optimization task where ingredient meanings and historical recipe examples can guide search.

\paragraph{Iron Mind~\cite{macknight2025pre}:}
The Iron Mind benchmark suite covers categorical reaction optimization tasks derived from experimental chemistry datasets. It includes reaction families such as Buchwald-Hartwig coupling, Suzuki reactions, reductive amination, alkylation/deprotection, and Chan-Lam chemistry. Each optimization problem selects among discrete experimental choices, such as reactants, ligands, bases, solvents, or other categorical reaction conditions, and evaluates the resulting outcome using dataset-backed objective values. The benchmark is especially useful for studying semantic optimization because the available context can include reaction descriptions, parameter metadata, objective definitions, and related paper text, all of which may encode chemically meaningful priors.

\paragraph{MuJoCo Control~\cite{todorov2012mujoco}:}
The MuJoCo Control benchmark evaluates compact continuous-control optimization problems based on deterministic rigid-body simulations. Rather than optimizing high-dimensional policy-network weights, the tasks expose interpretable controller or hyperparameter variables for environments such as cartpole swing-up, double-cartpole balancing, and planar reaching. Each candidate controller is rolled out in simulation, or evaluated through a tabular surrogate, and receives a scalar loss reflecting task performance over a finite horizon. The benchmark combines nonlinear dynamics, seeded environment perturbations, actuator constraints, and mixed parameter types, making it a useful test of whether semantic descriptions of environments and controller parameters improve black-box search.

\paragraph{NASBench201~\cite{dong2020bench}:}
The NASBench201 benchmark evaluates neural architecture search over a discrete architecture space. Each trial specifies categorical architectural choices, typically corresponding to operation selections on edges of a small cell topology, and the benchmark returns tabulated performance for standard image-classification datasets. Because the expensive training runs have already been evaluated and stored, the benchmark provides reproducible architecture-search objectives while avoiding the cost of retraining networks during optimization. Its semantic context consists of parameter descriptions and test-performance objectives, allowing optimizers to exploit knowledge about neural architecture design while operating in a purely categorical search space.

\paragraph{NeqSim~\cite{neqsim2026software}:}
The NeqSim benchmark models process-simulation optimization for gas-processing systems. Candidate configurations control operating variables in separator and compressor processes, including continuous and integer process settings, with evaluation modes ranging from dry-gas and wet-gas compression to multi-stage export, offshore network compression, and CO$_2$-management scenarios. Each trial is evaluated by direct NeqSim simulation or a surrogate, with objectives such as process cost or operating efficiency and explicit feasibility constraints for simulation failures or process violations. The benchmark tests optimization in an engineering domain where process context, objective definitions, and parameter roles can provide useful guidance.

\paragraph{Paint Mix:}
The Paint Mix benchmark is a controlled color-formulation task. For each seeded instance, three available paints are selected from a palette, and the optimizer chooses continuous mixing ratios for those paints. The resulting RGB color is compared against a named target color, and the objective is a normalized mismatch score from 0 to 10, where lower values indicate a closer match. Because the benchmark exposes the selected paint names, target-paint description, and ratio-variable roles, it provides a simple but interpretable setting for testing whether semantic color information can help guide continuous optimization.

\paragraph{PD1~\cite{JMLR:v25:23-0269}:}
The PD1 benchmark wraps tabular neural-network training experiments for hyperparameter optimization. Each task corresponds to a dataset and model family from the PD1 neural network tuning corpus, with objectives predicted by surrogate models trained on recorded runs and feasibility represented through a learned failure constraint. The search spaces vary by task and may include continuous, integer, and categorical hyperparameters such as learning rate, optimizer, regularization, and model-configuration choices. In addition to the primary objective, the benchmark can expose auxiliary training metrics and contextual descriptions of the task, model, parameters, and expected heuristics, making it a semantically rich benchmark for neural-network hyperparameter tuning.

\section{Further Results}

\subsection{Step Curves:}
\label{app:step}

\begin{figure*}[h]
    \centering
    
    \begin{subfigure}[b]{0.45\textwidth}
        \centering
        \includegraphics[width=\linewidth]{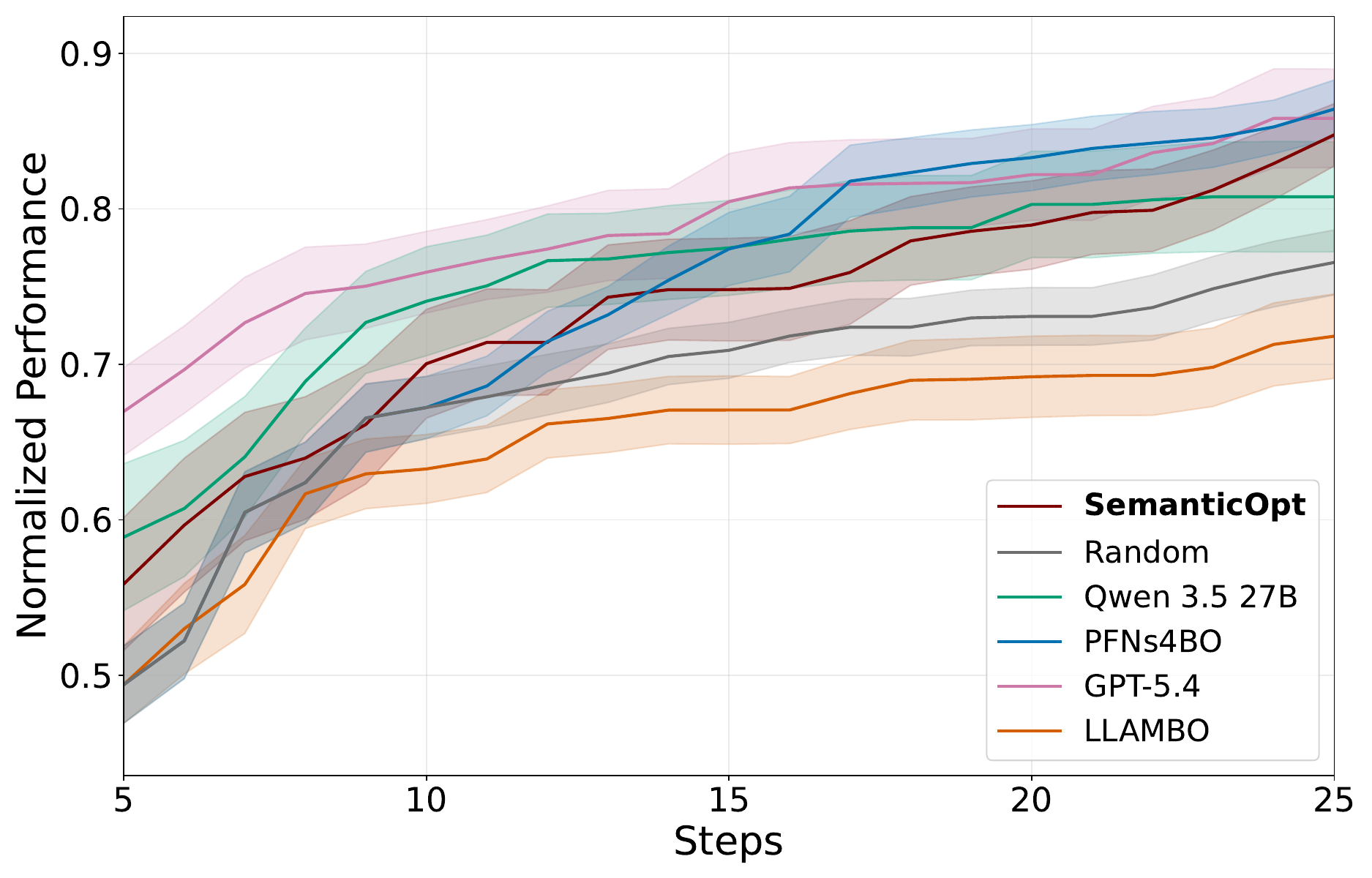}
        \caption{Convex Decomposition Results}
        \label{fig:convex}
    \end{subfigure}
    \hfill
    \begin{subfigure}[b]{0.45\textwidth}
        \centering
        \includegraphics[width=\linewidth]{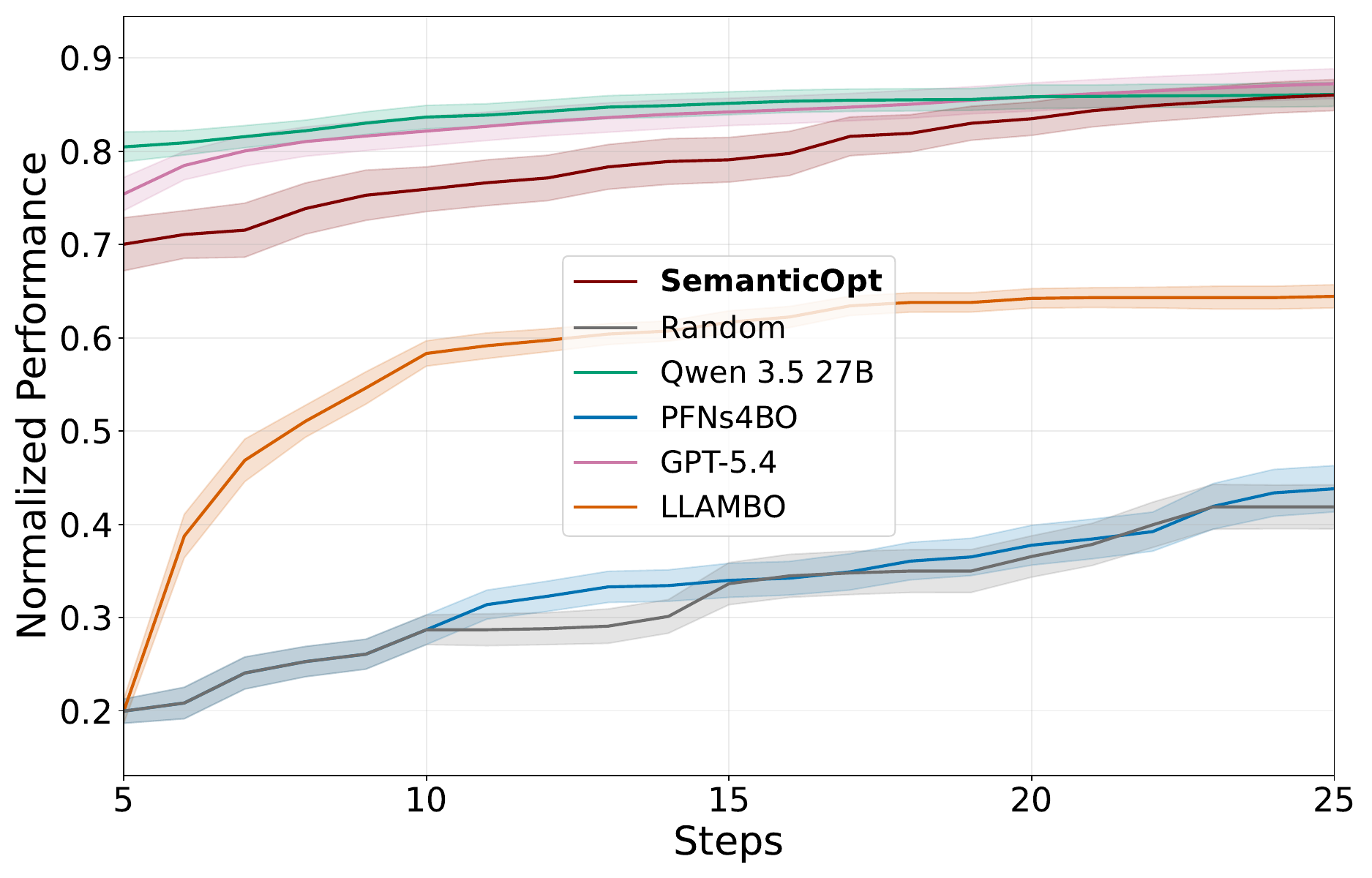}
        \caption{Cookie Recipe Results}
        \label{fig:cookie}
    \end{subfigure}

    \begin{subfigure}[b]{0.45\textwidth}
        \centering
        \includegraphics[width=\linewidth]{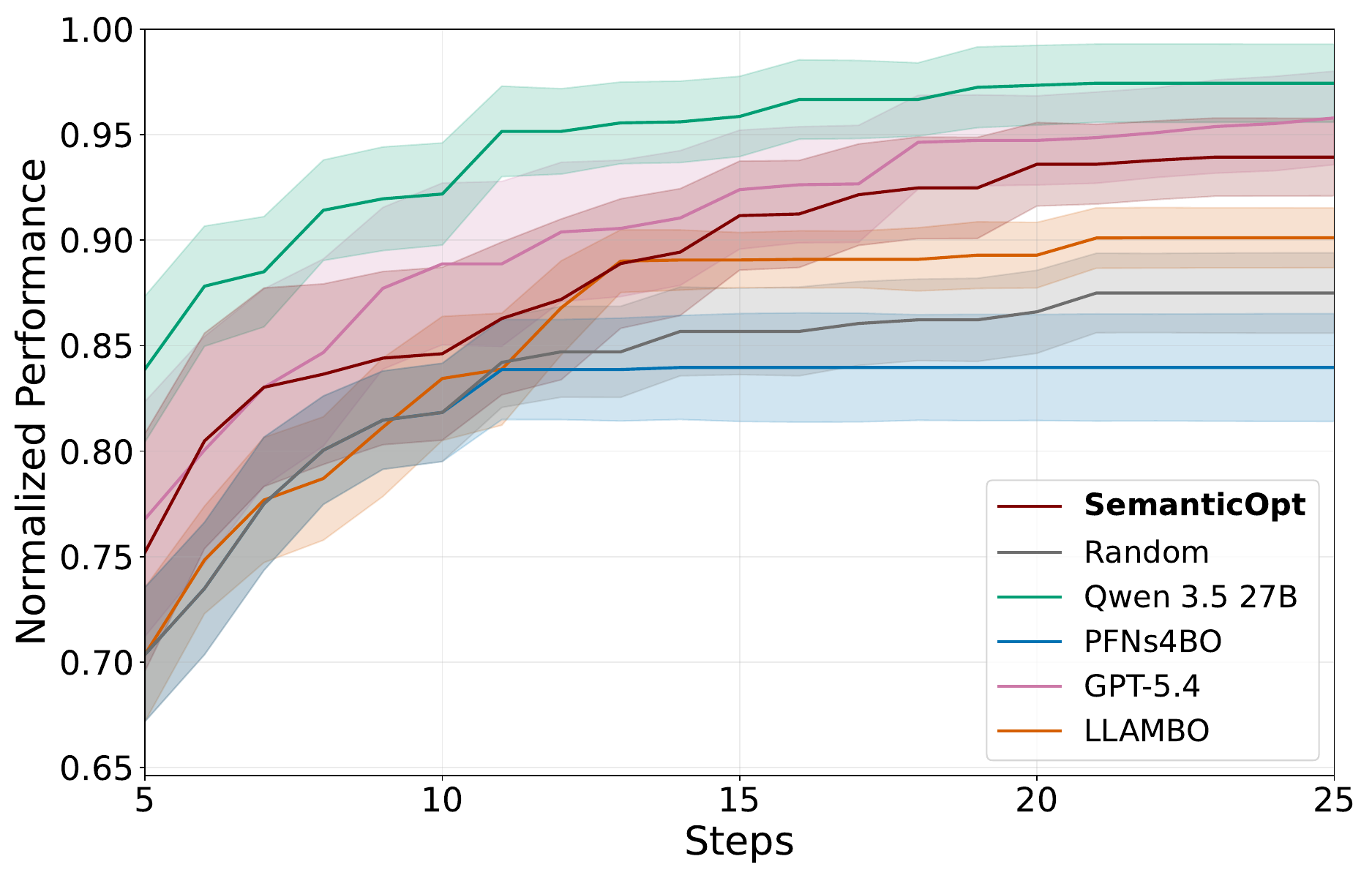}
        \caption{Iron Mind Results}
        \label{fig:iron_mind}
    \end{subfigure}
    \hfill
    \begin{subfigure}[b]{0.45\textwidth}
        \centering
        \includegraphics[width=\linewidth]{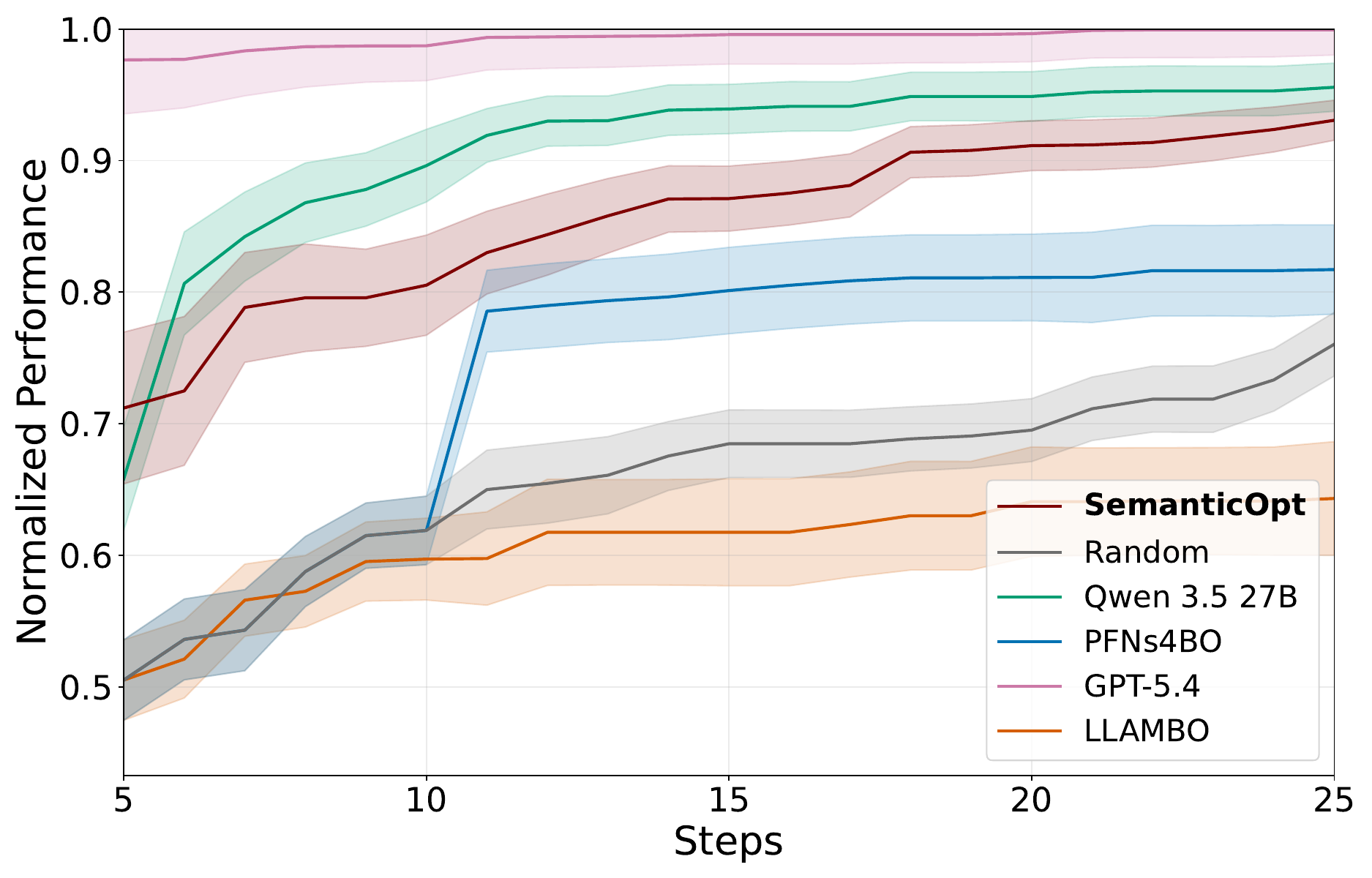}
        \caption{Nasbench201 Results}
        \label{fig:nas}
    \end{subfigure}

    \begin{subfigure}[b]{0.45\textwidth}
        \centering
        \includegraphics[width=\linewidth]{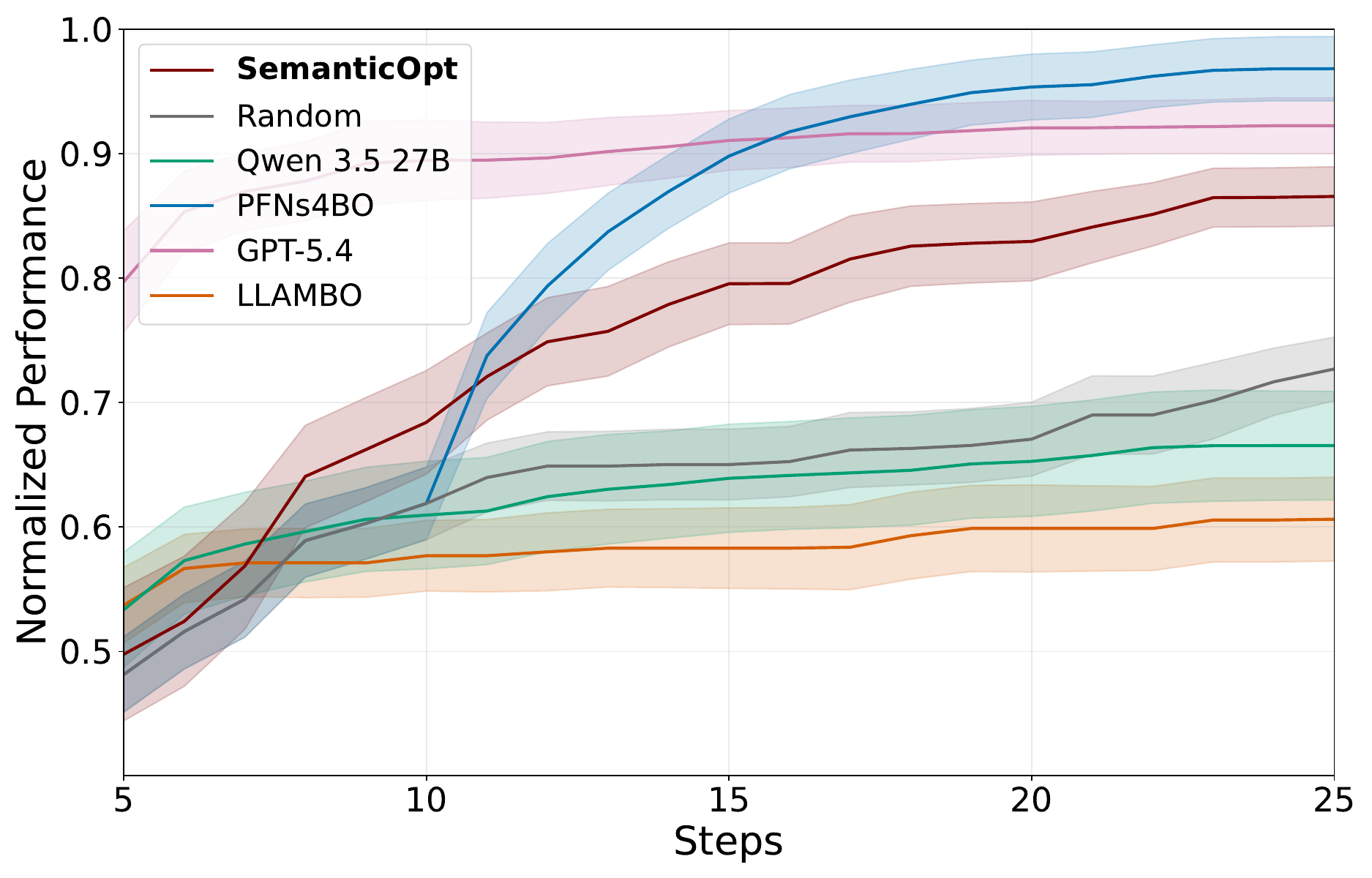}
        \caption{NeqSim Results}
        \label{fig:neqsim}
    \end{subfigure}

    \caption{Mean normalized performance over steps with paired standard error of the mean (SEM) of SemanticOpt compared to other LLM and transformer-based methods over various benchmarks with semantic data. Our method is SemanticOpt, which is fine-tuned on the Qwen 3.5 27B model.}
    \label{fig:step_results2}
\end{figure*}

We show the remaining step curves not shown in the main paper in Figure~\ref{fig:step_results2}.

\subsection{Win Rates:} 
\label{app:win_rate}

\begin{figure*}[t]
    \centering

    \begin{subfigure}[b]{0.32\textwidth}
        \centering
        \includegraphics[width=\linewidth]{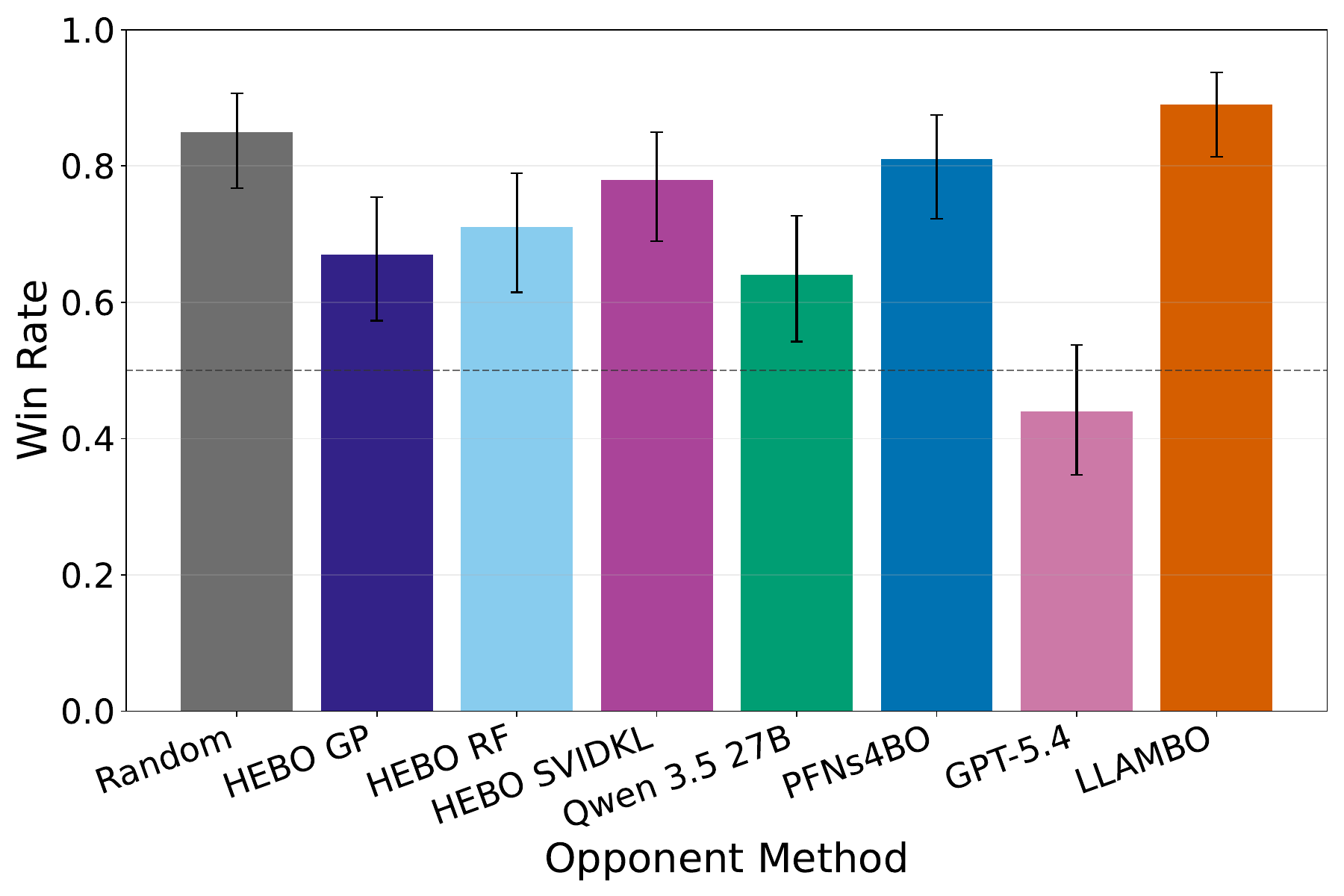}
        \caption{PD1}
        \label{fig:pd1_win}
    \end{subfigure}
    \hfill
    \begin{subfigure}[b]{0.32\textwidth}
        \centering
        \includegraphics[width=\linewidth]{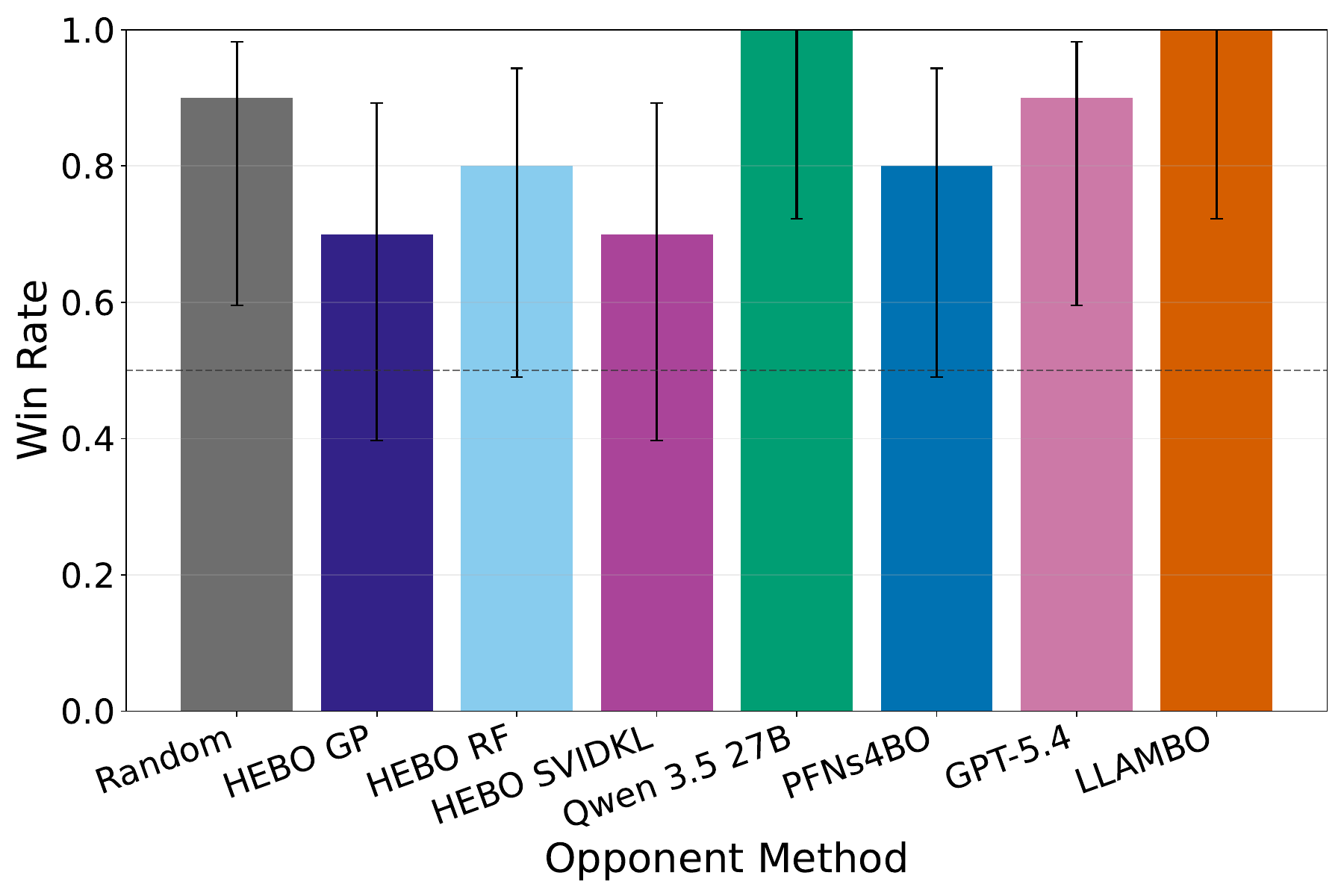}
        \caption{MuJoCo Control}
        \label{fig:mujoco_win}
    \end{subfigure}
    \hfill
    \begin{subfigure}[b]{0.32\textwidth}
        \centering
        \includegraphics[width=\linewidth]{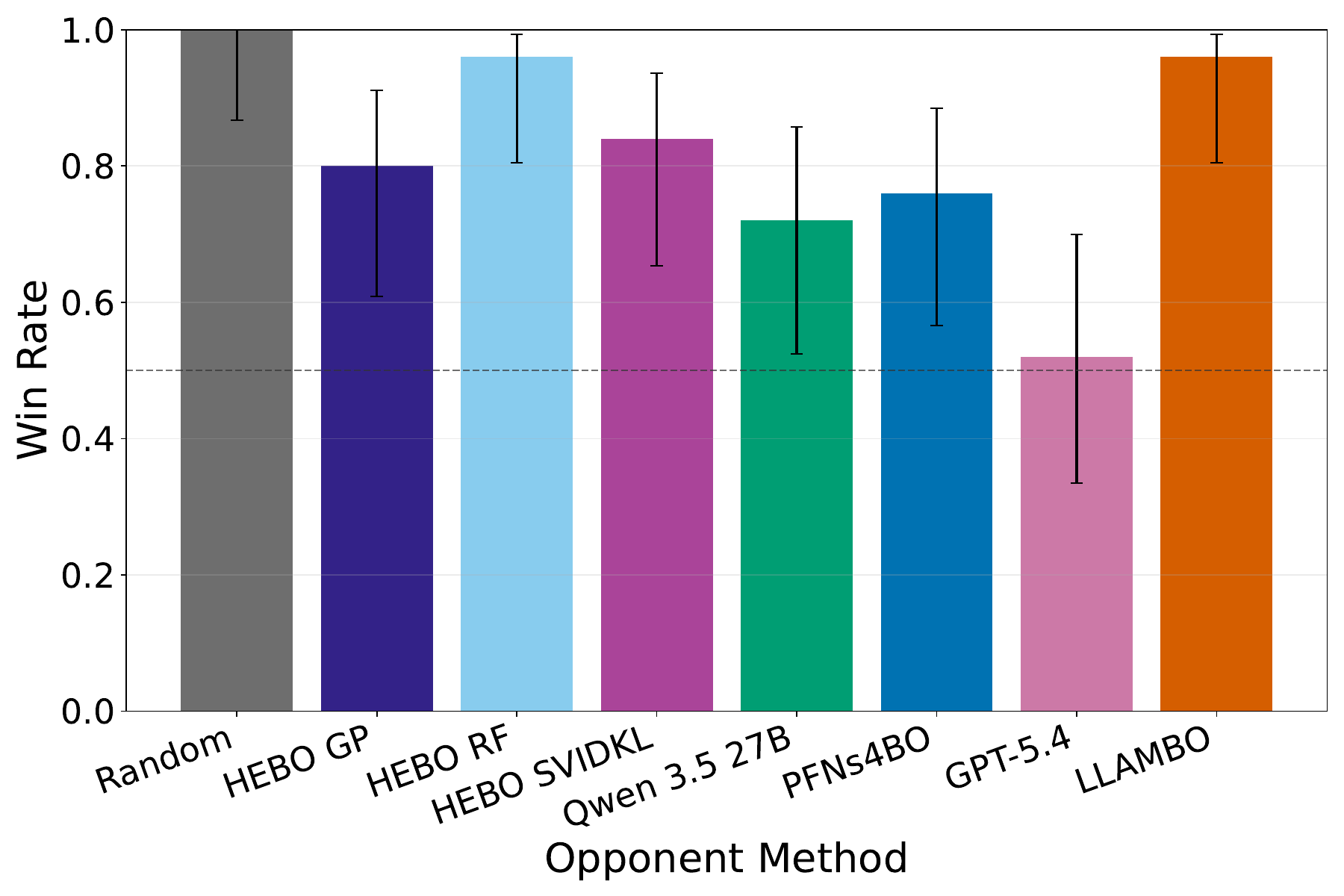}
        \caption{Paint Mixing}
        \label{fig:paint_mix_win}
    \end{subfigure}

    \vspace{0.5em}

    \begin{subfigure}[b]{0.32\textwidth}
        \centering
        \includegraphics[width=\linewidth]{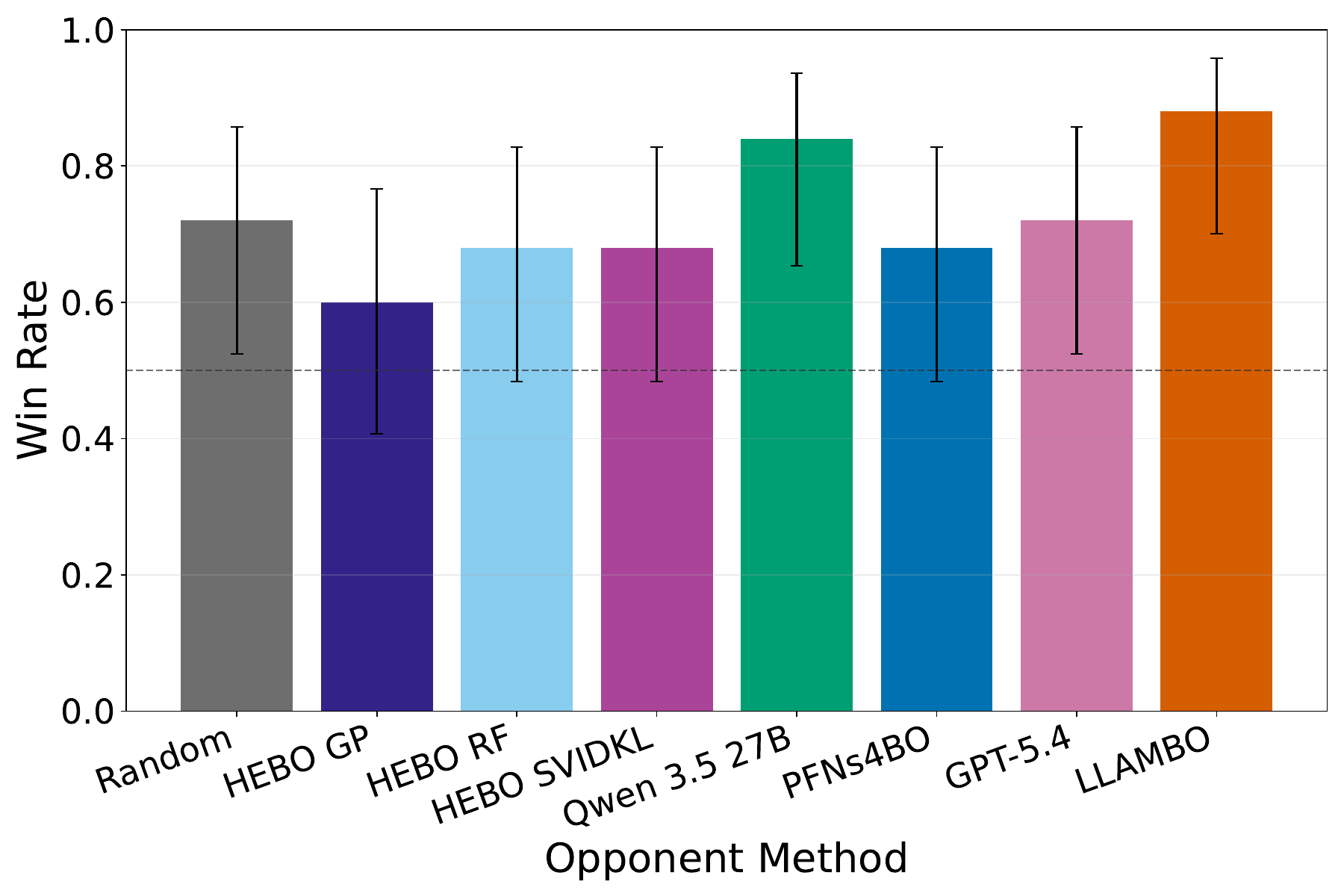}
        \caption{Airfoil CFD}
        \label{fig:airfoil_win}
    \end{subfigure}
    \hfill
    \begin{subfigure}[b]{0.32\textwidth}
        \centering
        \includegraphics[width=\linewidth]{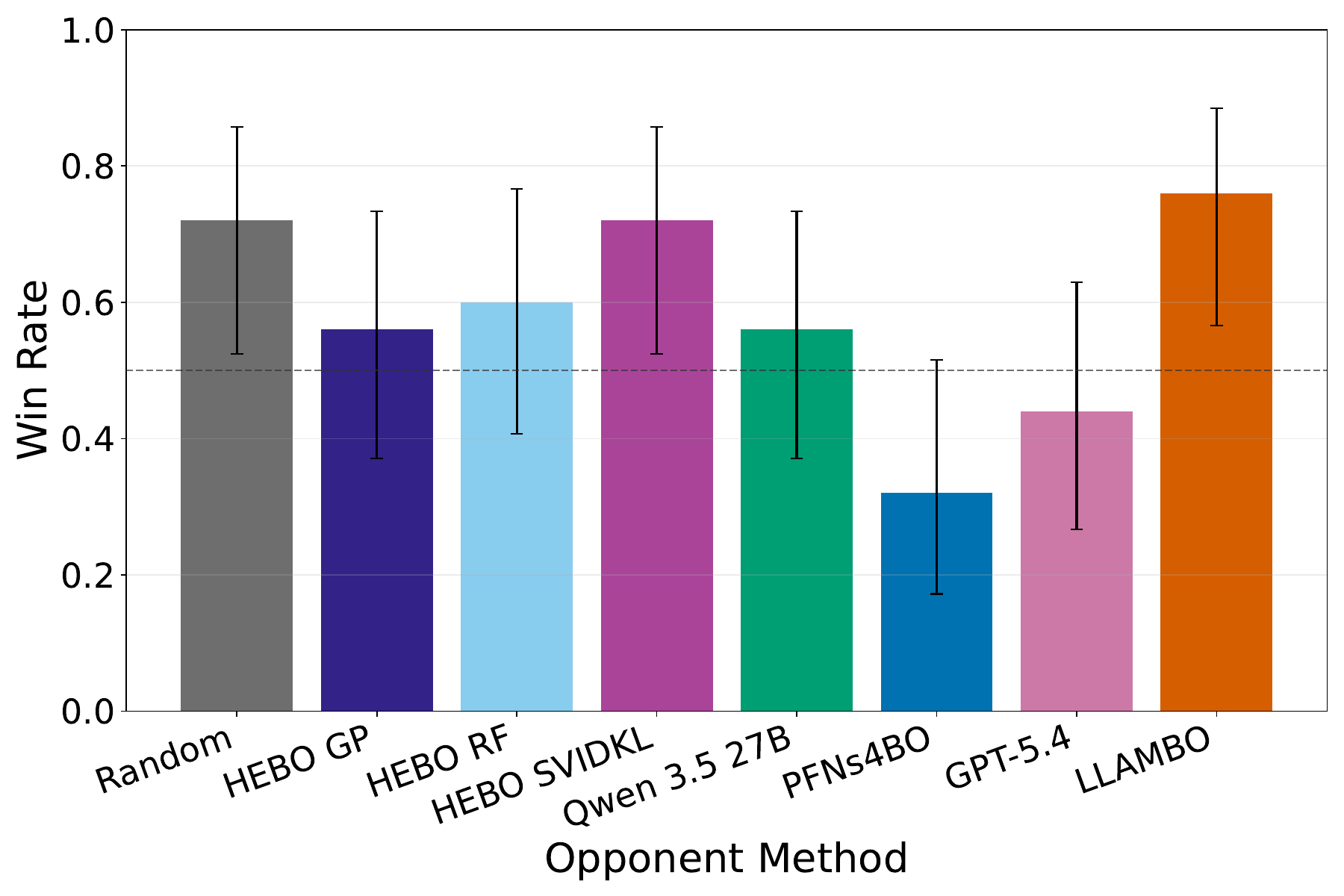}
        \caption{Convex Decomposition}
        \label{fig:convex_win}
    \end{subfigure}
    \hfill
    \begin{subfigure}[b]{0.32\textwidth}
        \centering
        \includegraphics[width=\linewidth]{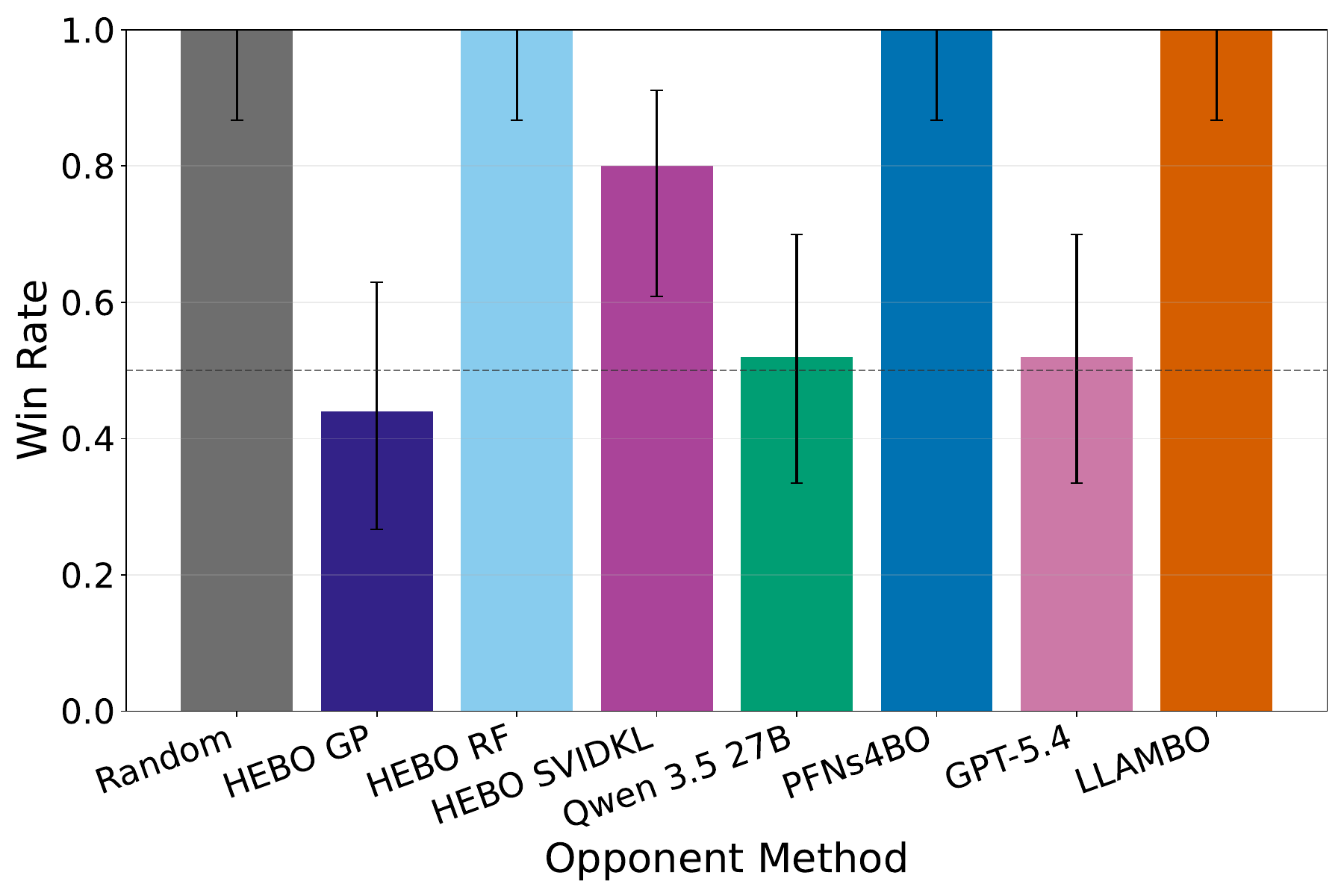}
        \caption{Cookie Recipe}
        \label{fig:cookie_win}
    \end{subfigure}

    \vspace{0.5em}

    \begin{subfigure}[b]{0.32\textwidth}
        \centering
        \includegraphics[width=\linewidth]{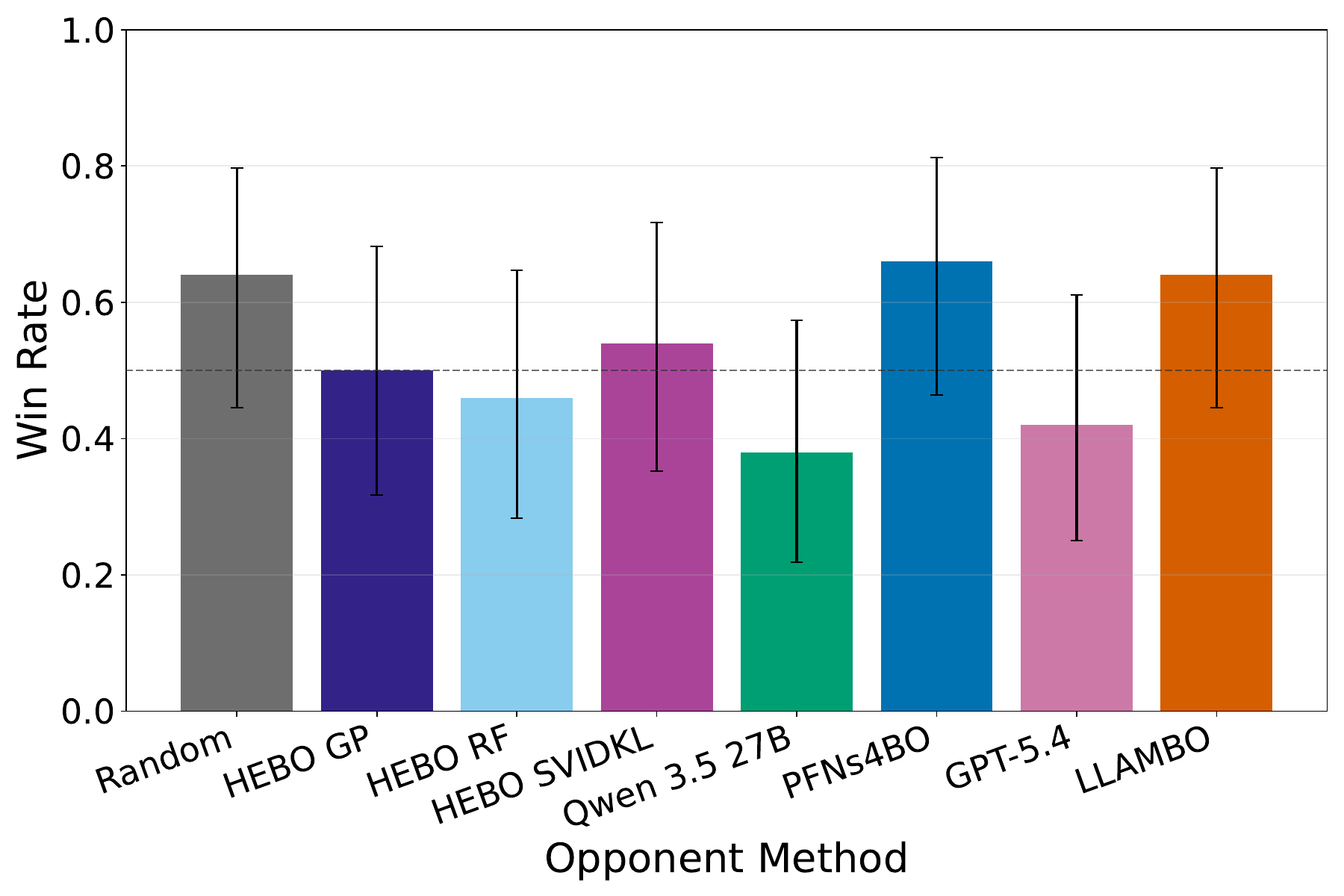}
        \caption{Iron Mind}
        \label{fig:ironmind_win}
    \end{subfigure}
    \hfill
    \begin{subfigure}[b]{0.32\textwidth}
        \centering
        \includegraphics[width=\linewidth]{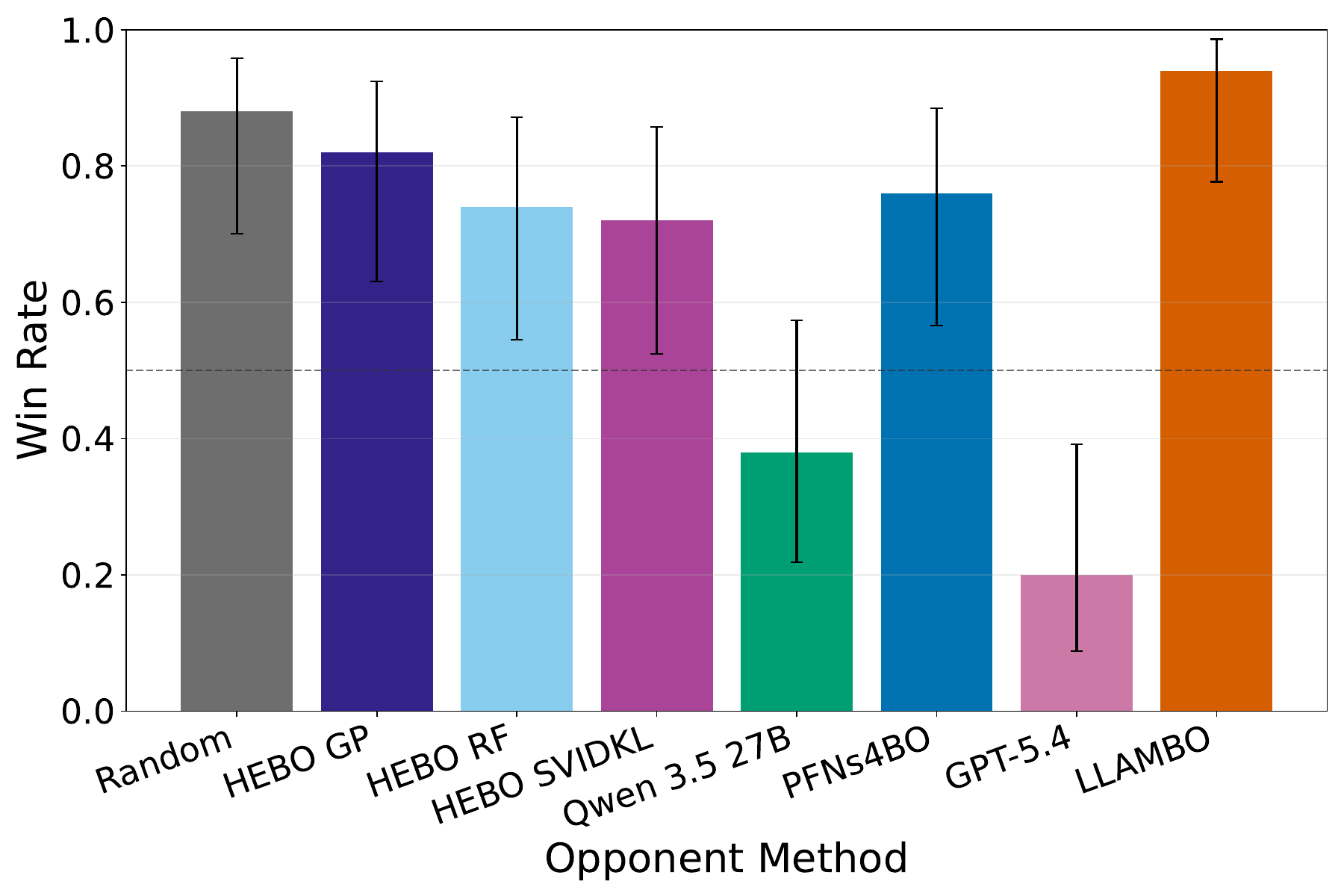}
        \caption{Nasbench201}
        \label{fig:nasbench_win}
    \end{subfigure}
    \hfill
    \begin{subfigure}[b]{0.32\textwidth}
        \centering
        \includegraphics[width=\linewidth]{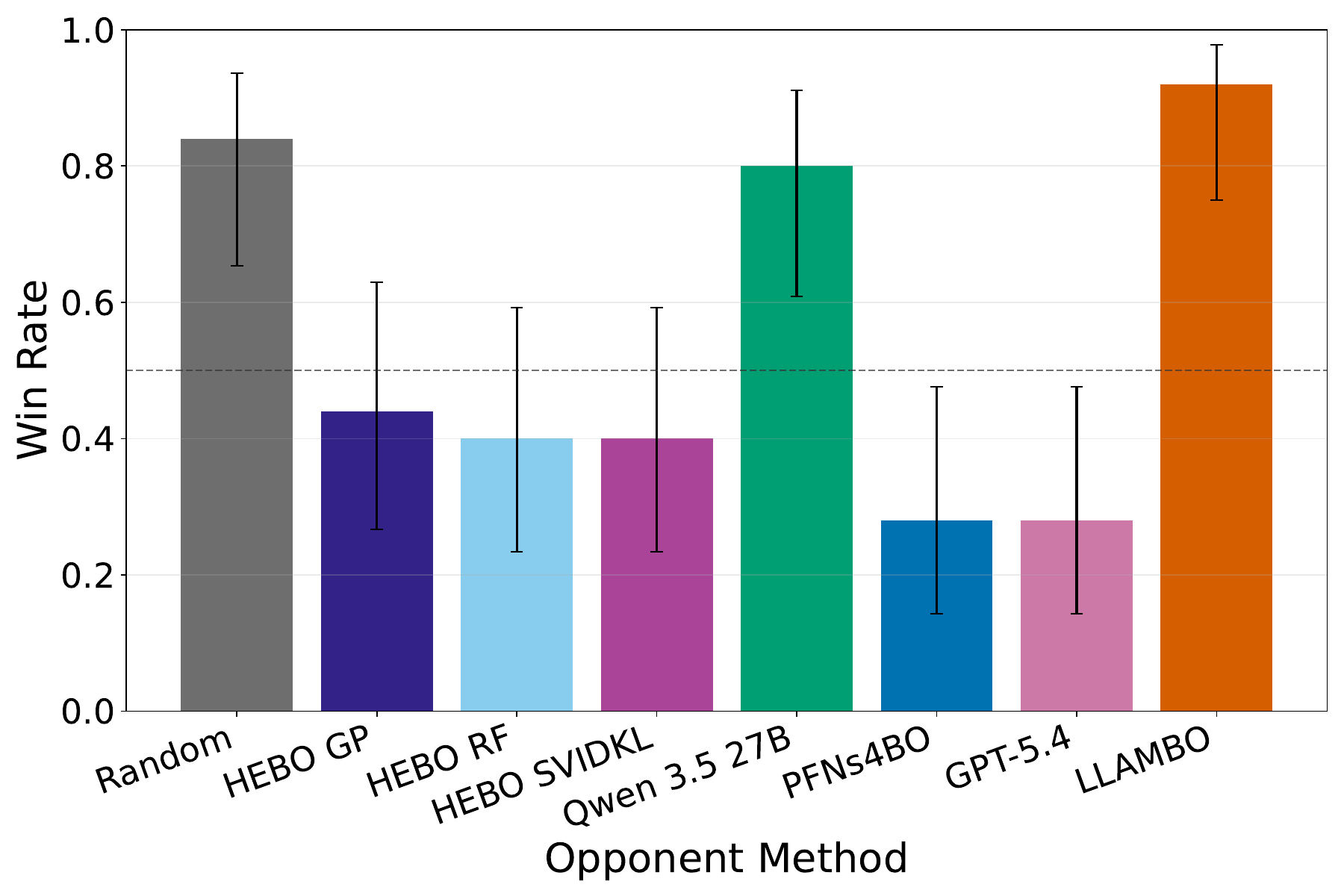}
        \caption{NeqSim}
        \label{fig:neqsim_win}
    \end{subfigure}

    \caption{Win rates of SemanticOpt compared to other LLM- and transformer-based methods across benchmark tasks with semantic data.}
    \label{fig:win_rates}
\end{figure*}

We show the win rates after 25 total evaluations of SemanticOpt against baseline optimizers in Figure~\ref{fig:win_rates}. A higher win rate means SemanticOpt beat the opponent more times throughout the benchmark seeds. The results are consistent with the mean normalized performance tested in previous results. 

\subsection{Trajectory Lengths:}
\label{app:length}

\begin{figure*}[h]
    \centering
    
    \begin{subfigure}[b]{0.45\textwidth}
        \centering
        \includegraphics[width=\linewidth]{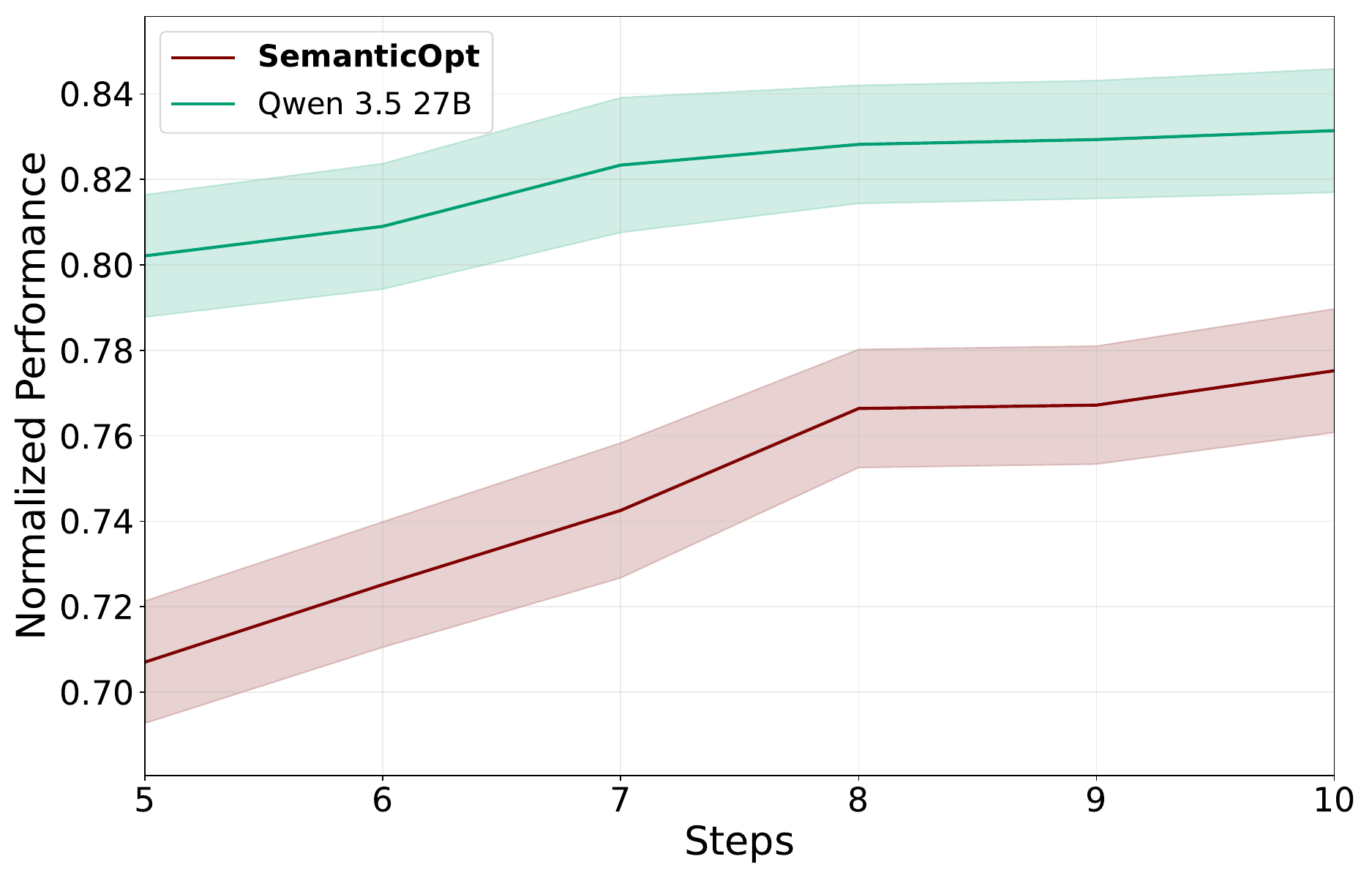}
        \caption{10 Total Evaluations}
        \label{fig:step_10}
    \end{subfigure}
    \hfill
    \begin{subfigure}[b]{0.45\textwidth}
        \centering
        \includegraphics[width=\linewidth]{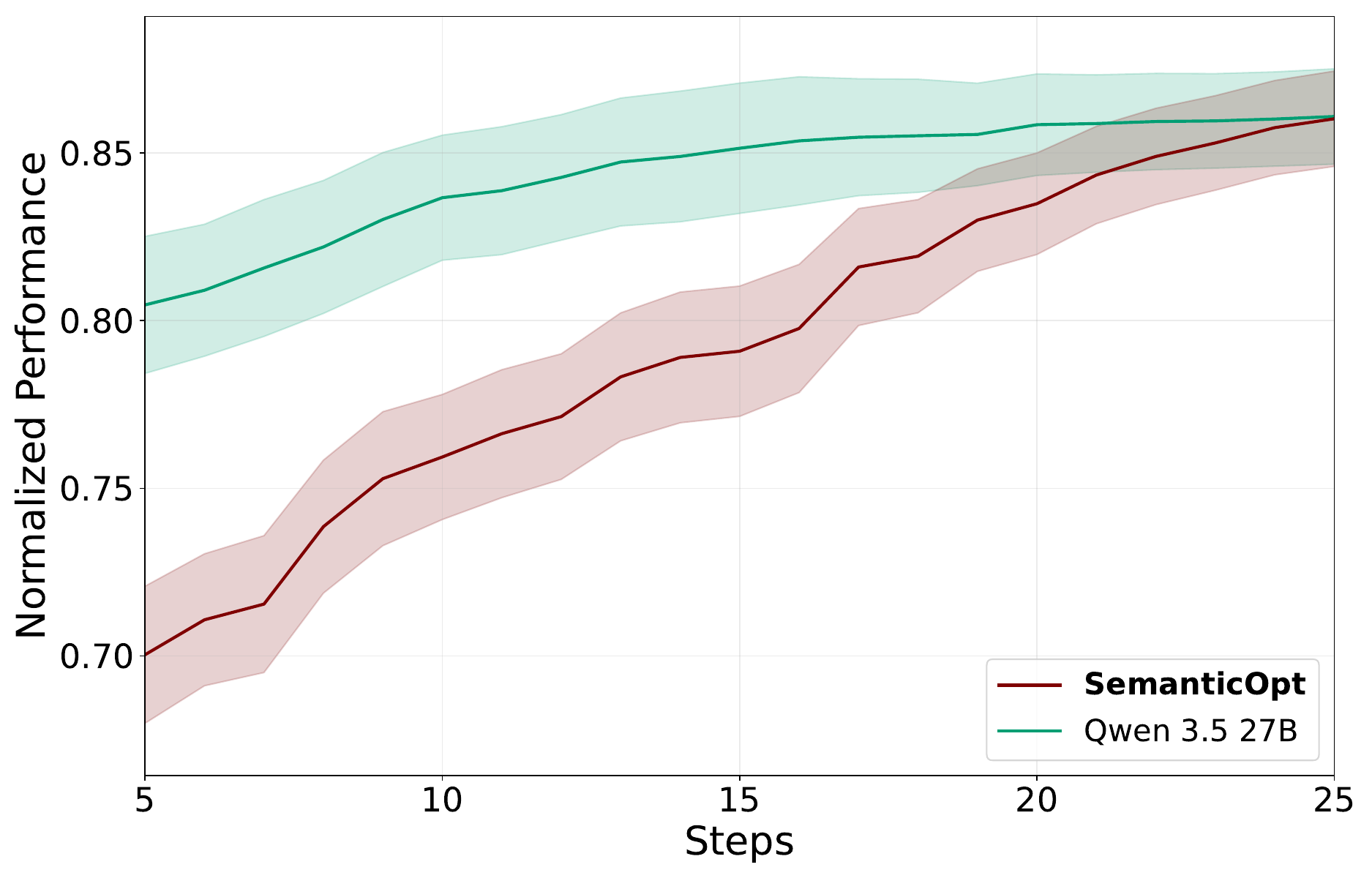}
        \caption{25 Total Evaluations}
        \label{fig:step_25}
    \end{subfigure}

    \begin{subfigure}[b]{0.45\textwidth}
        \centering
        \includegraphics[width=\linewidth]{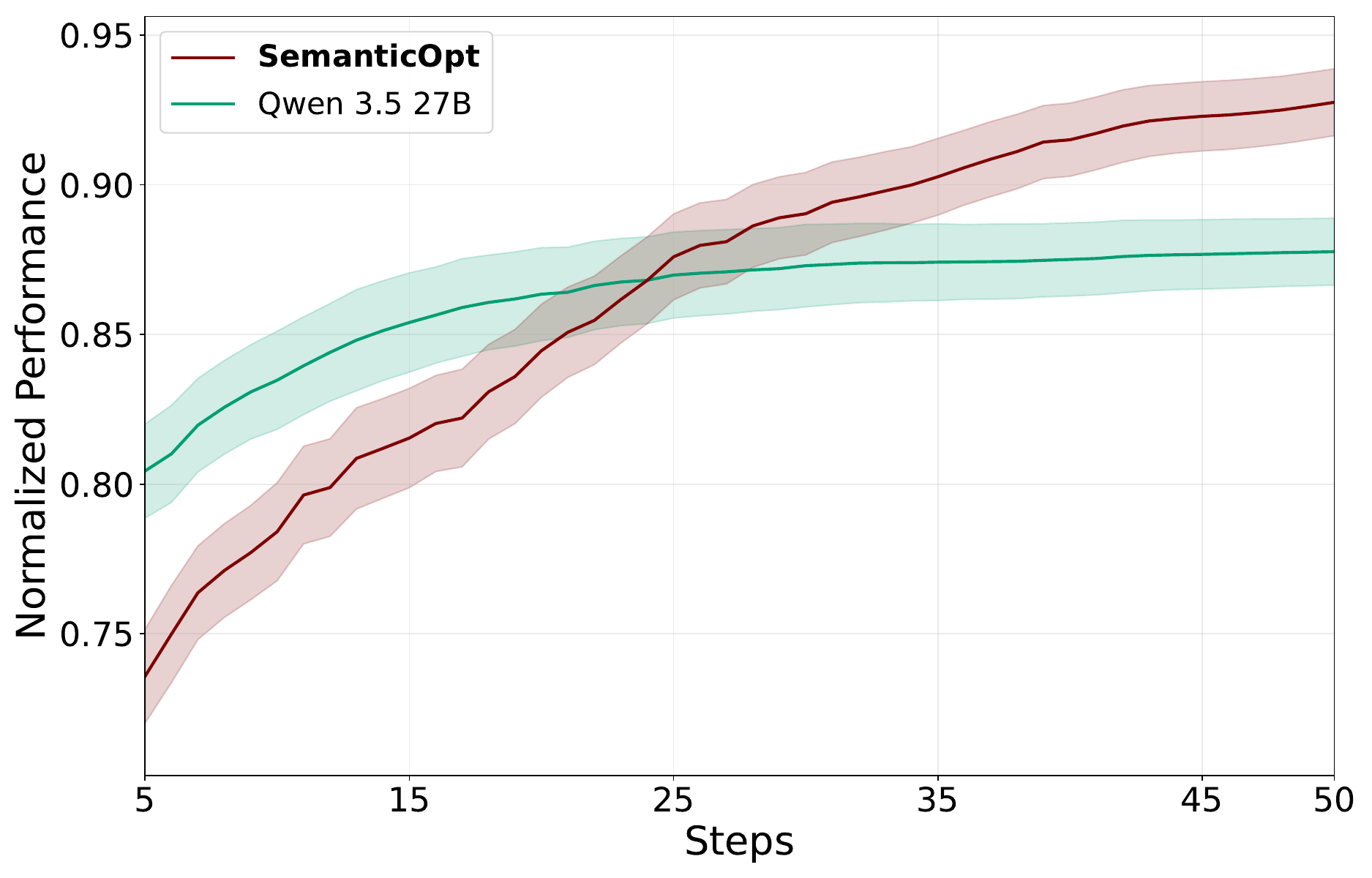}
        \caption{50 Total Evaluations}
        \label{fig:step_50}
    \end{subfigure}
    \hfill
    \begin{subfigure}[b]{0.45\textwidth}
        \centering
        \includegraphics[width=\linewidth]{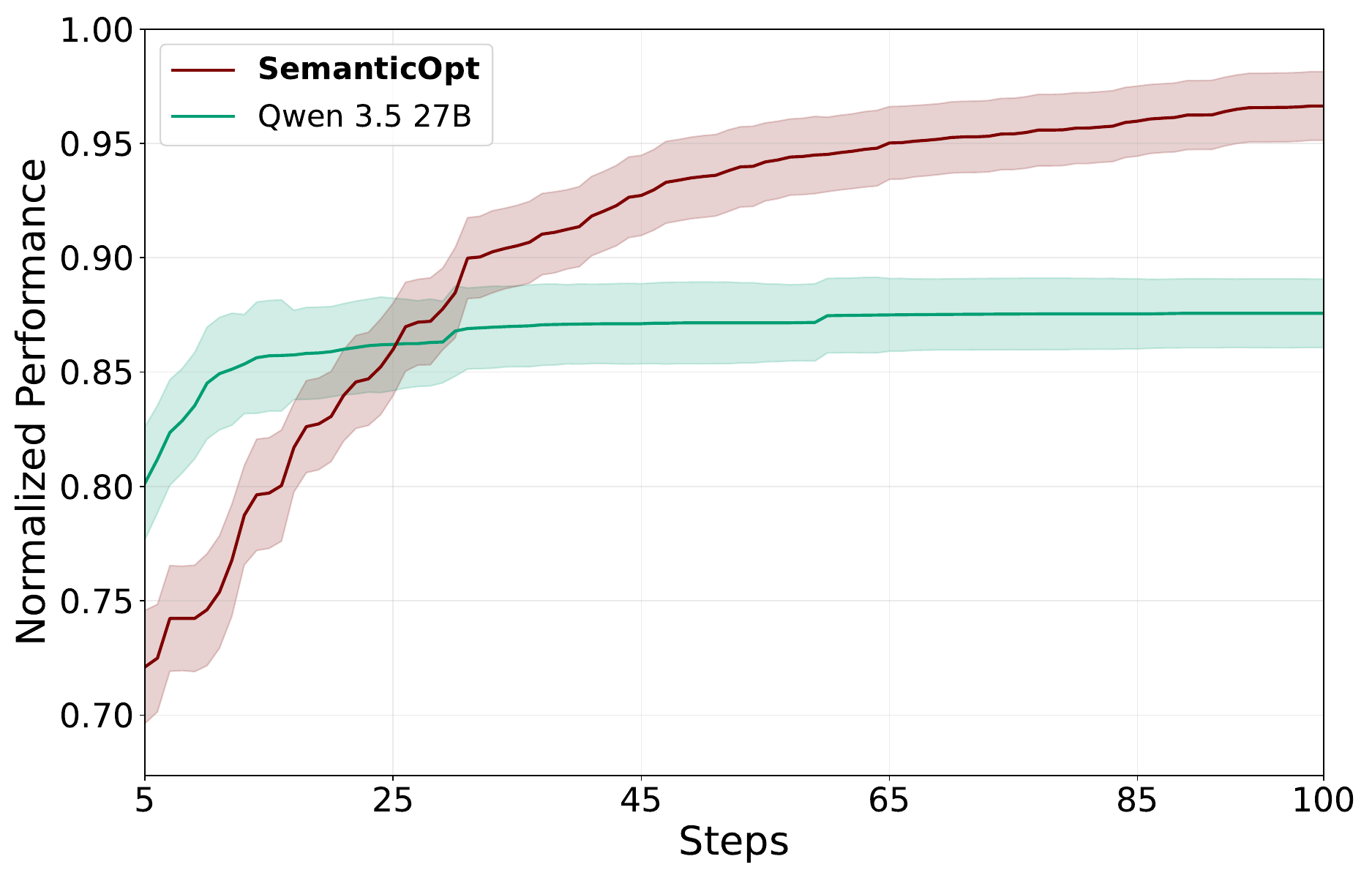}
        \caption{100 Total Evaluations}
        \label{fig:step_100}
    \end{subfigure}

    \caption{Mean normalized performance with paired SEM of SemanticOpt against Qwen 3.5 27B over different trajectory lengths on the Cookie Recipe benchmark.}
    \label{fig:lengths}
\end{figure*}

\begin{figure*}[h]
    \centering
    
    \begin{subfigure}[b]{0.45\textwidth}
        \centering
        \includegraphics[width=\linewidth]{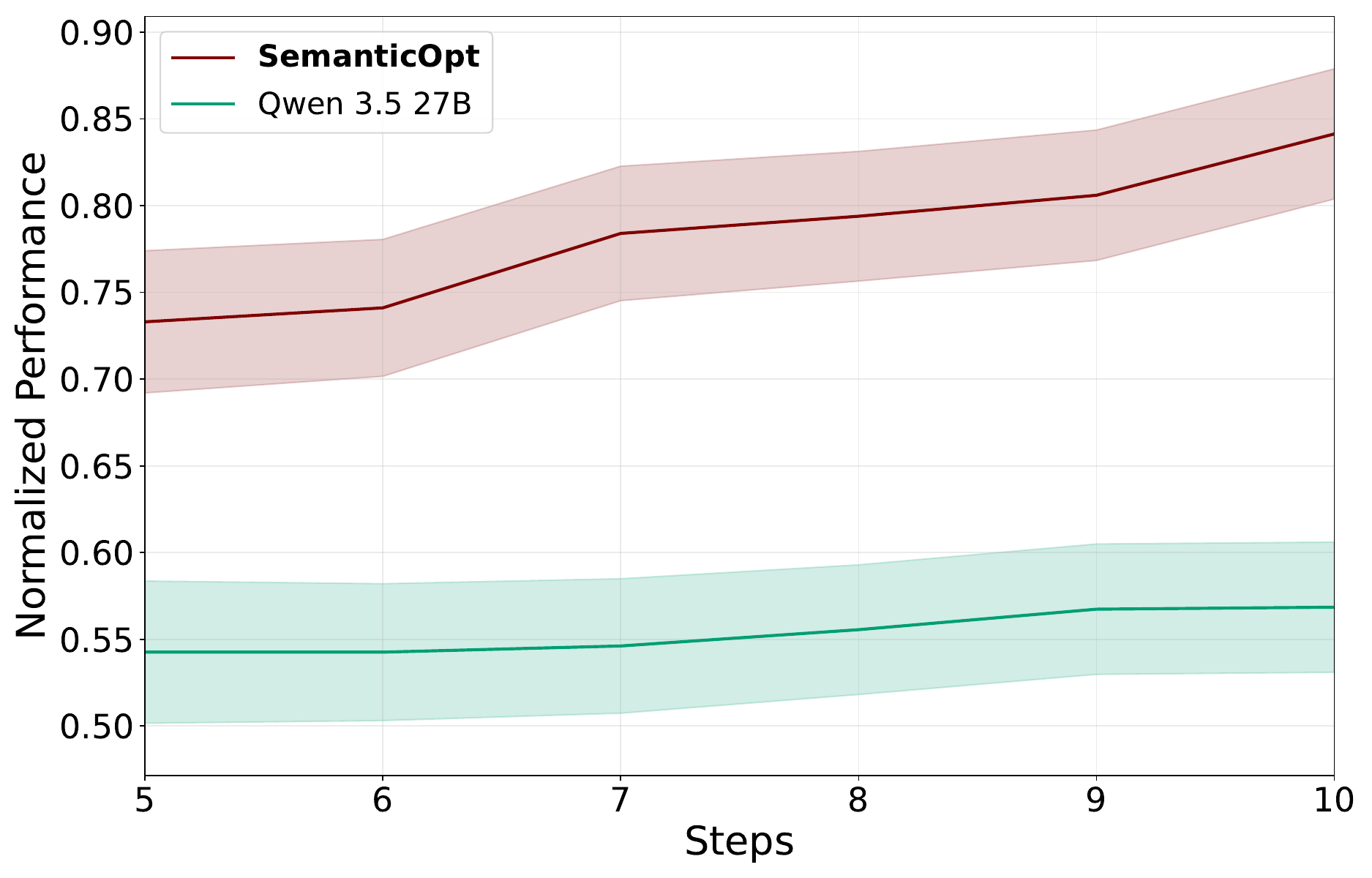}
        \caption{10 Total Evaluations}
        \label{fig:step2_10}
    \end{subfigure}
    \hfill
    \begin{subfigure}[b]{0.45\textwidth}
        \centering
        \includegraphics[width=\linewidth]{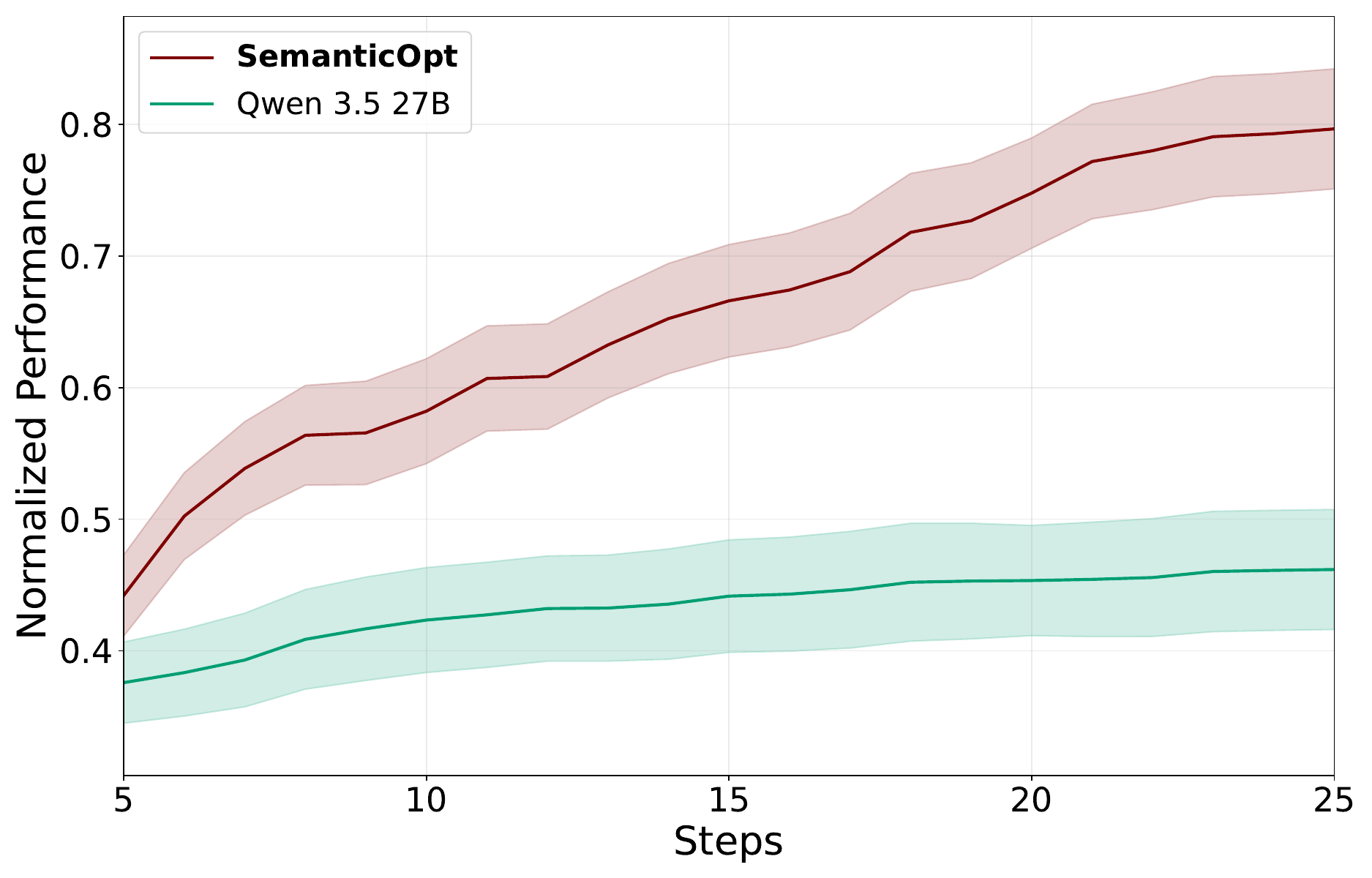}
        \caption{25 Total Evaluations}
        \label{fig:step2_25}
    \end{subfigure}

    \begin{subfigure}[b]{0.45\textwidth}
        \centering
        \includegraphics[width=\linewidth]{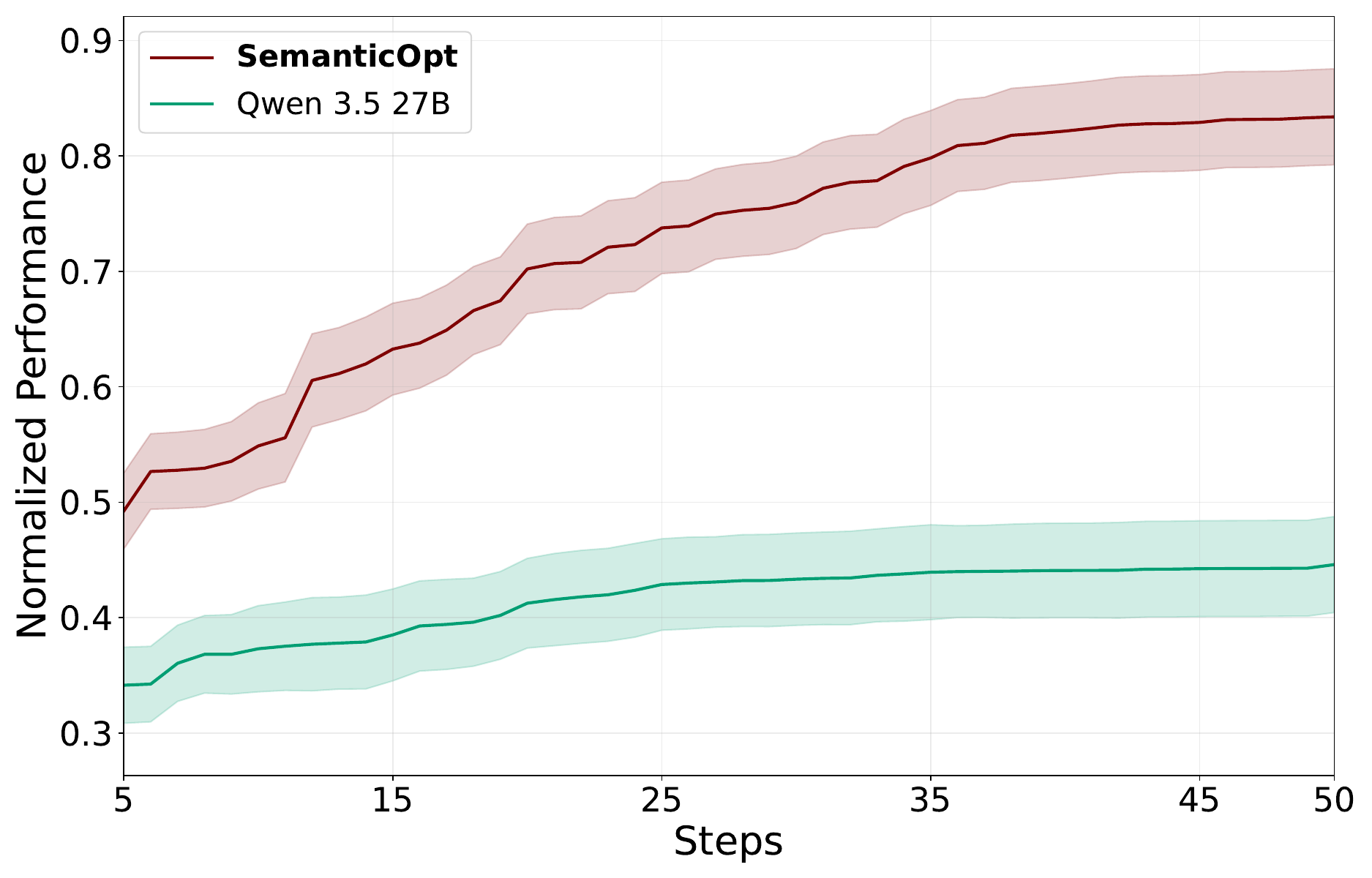}
        \caption{50 Total Evaluations}
        \label{fig:step2_50}
    \end{subfigure}
    \hfill
    \begin{subfigure}[b]{0.45\textwidth}
        \centering
        \includegraphics[width=\linewidth]{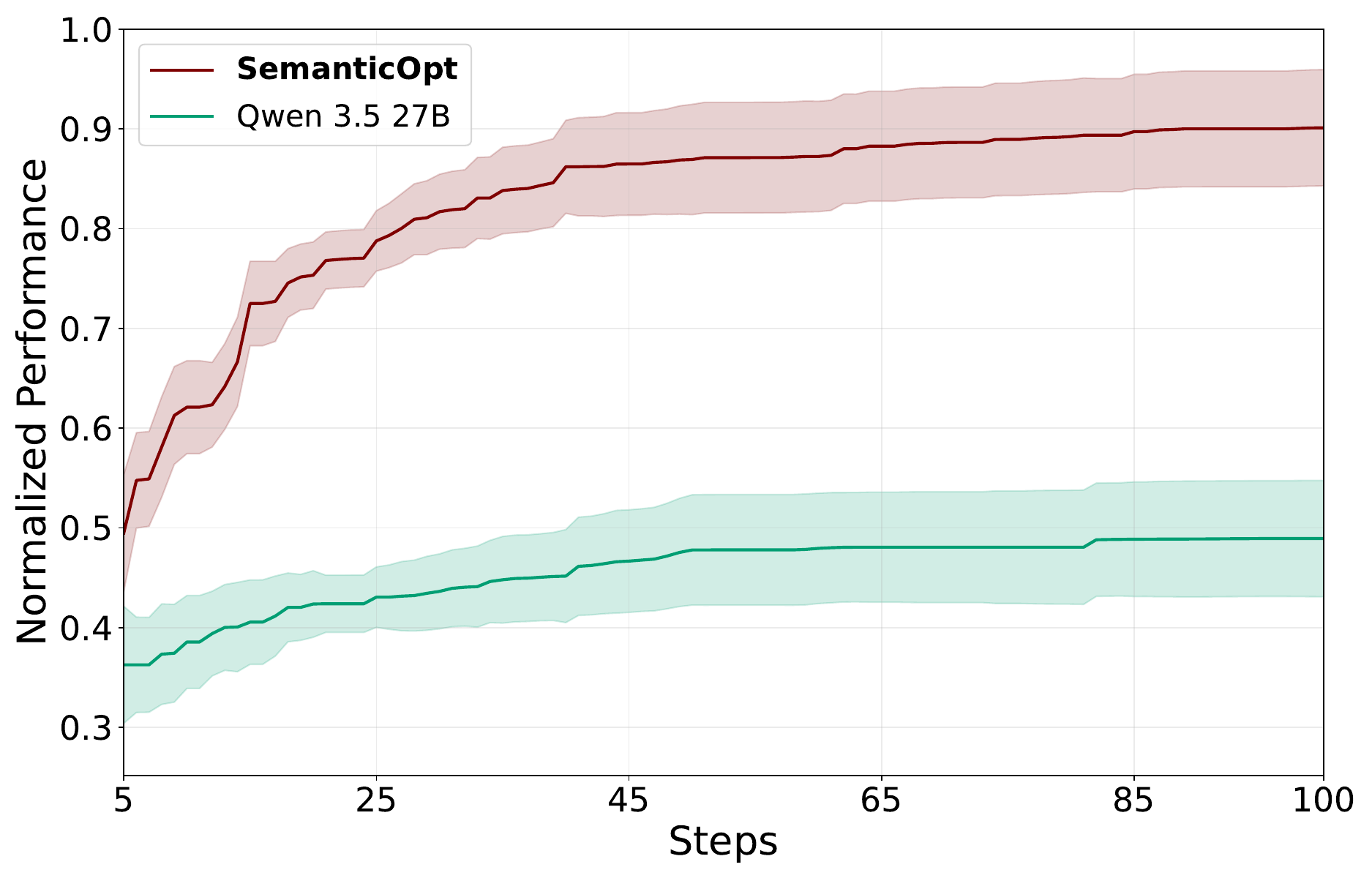}
        \caption{100 Total Evaluations}
        \label{fig:step2_100}
    \end{subfigure}

    \caption{Mean normalized performance with paired SEM of SemanticOpt against Qwen 3.5 27B over different trajectory lengths on the Airfoil CFD benchmark.}
    \label{fig:lengths2}
\end{figure*}

We provide comparisons of SemanticOpt performance over a range of trajectory lengths on the Cookie Recipe benchmark in Figure~\ref{fig:lengths}. We also show results on the Airfoil CFD benchmark in Figure~\ref{fig:lengths2}. The 10, 25, and 50 step tests are conducted with 25 seeds and the 100 step test is conducted with 10 seeds. These benchmarks were selected as there was still a considerable gap towards the global optimum after 25 steps. Overall, we see that SemanticOpt performance continues to improve over the range of evaluations, whereas Qwen 3.5 27B performance plateaus.

\subsection{Surrogate Model Performance:}
\label{app:surrogate}

\begin{figure*}[h]
    \centering
    
    \includegraphics[width=0.6\linewidth]{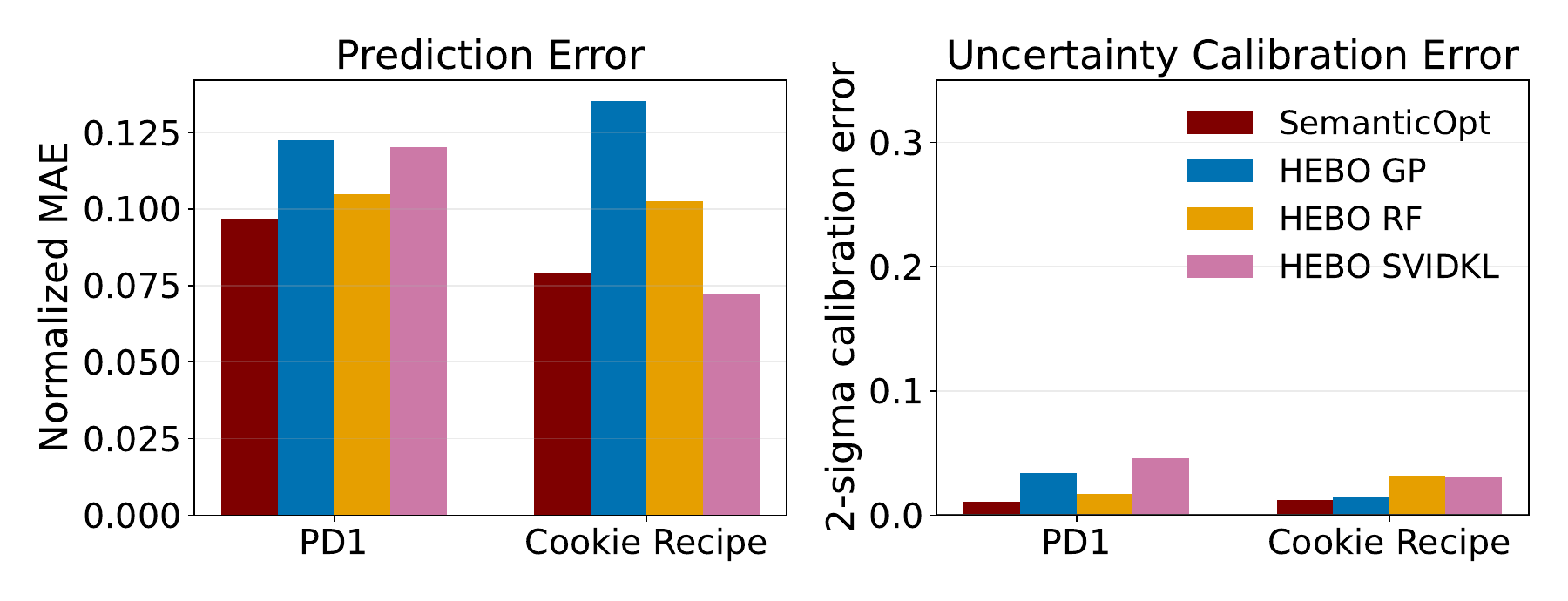}

    \caption{Surrogate model performance of SemanticOpt. The fine-tuned SemanticOpt model matches or beats the performance of baseline surrogate models (GP, RF, and SVIDKL). This result is tested by fitting the surrogate model to the same evaluation history as SemanticOpt and evaluating the performance of the surrogate model at SemanticOpt's selected point.}
    \label{fig:surrogate}
\end{figure*}

To evaluate the quality of the surrogate model outputs, we show the improvement of SemanticOpt by comparing the model's distribution predictions along their trajectories to HEBO GP, RF, and SVIDKL in Figure~\ref{fig:surrogate}. We fit the HEBO surrogate models to the previously selected points of SemanticOpt and evaluate their predicted distributions at the selected point to the ground truth. We see that SemanticOpt has either best or second best mean absolute error performance compared to the baselines. We also report 2\(\sigma\) calibration error, defined as the gap between empirical coverage of \([\mu - 2\sigma, \mu + 2\sigma]\) and the nominal \(95\%\) Gaussian coverage. SemanticOpt's uncertainty calibration is better than the surrogate baselines. As a whole, the fine-tuned model beats or matches the surrogate model performance of our baselines on our test benchmarks.

\end{document}